\documentclass[12pt]{article}
\usepackage[dvips]{graphicx}
\usepackage{makeidx}
\usepackage{amsmath}
\usepackage{amsfonts}

\usepackage{color}
\usepackage{setspace}

\newcommand{\Var}[1]{{\lambda_{#1}}}

\newcommand {\col}{{\mathrm{col}}}
\newcommand {\ess}{{\mathrm{ess}}}

\newcommand {\const}{{\mathrm{const}}}

\newcommand {\lambdavec}{\boldsymbol{\lambda}}

\newcommand {\thetavec}{\mbox{\boldmath $\theta$}}

\newcommand{\norms}[1]{\norm{#1}_{\mathcal{A}}}

\newcounter{defnc}
\newcounter{thmc}

\newcounter{lemc}
\newcounter{propc}
\newcounter{problemc}
\newcounter{examplec}
\newcounter{figurec}
\newcounter{assumec}

\newtheorem{thm}[thmc]{Theorem}

\newtheorem{lem}[lemc]{Lemma }

\newtheorem{problem}[problemc]{Problem}

\newtheorem{defn}[defnc]{Definition}

\newcommand{\eps}{\varepsilon}

\newcommand{\bfx}{\mathbf{x}}

\newcommand{\bff}{\mathbf{f}}
\newcommand{\bfz}{\mathbf{z}}

\newcommand{\bfu}{\mathbf{u}}
\newcommand{\bfg }{\mathbf{g}}
\newcommand{\bfq }{\mathbf{q}}

\newcommand{\bfy }{\mathbf{y}}

\newcommand{\pd}{{\partial}}

\newcommand{\Real}{\mathbb{R}}
\newcommand{\Natural}{\mathbb{N}}

\newcommand{\norm}[1]{\left\Vert#1\right\Vert}

\newcommand{\red}[1]{\textcolor{red}{#1}}

\makeindex

\begin{document}

\renewcommand{\thedefnc}{\arabic{defnc}}
\renewcommand{\theexamplec}{\arabic{examplec}}
\renewcommand{\thepropc}{\arabic{propc}}
\renewcommand{\thethmc}{\arabic{thmc}}
\renewcommand{\theproblemc}{\arabic{problemc}}
\renewcommand{\thefigurec}{\arabic{figurec}}
\renewcommand{\theassumec}{\arabic{assumec}}

\title{\bf Invariant template matching in systems with spatiotemporal coding: a vote for instability}
\author{Ivan Tyukin\thanks{Laboratory for Perceptual Dynamics, RIKEN (Institute for Physical and Chemical Research)
                           Brain Science Institute, 2-1, Hirosawa, Wako-shi, Saitama, 351-0198, Japan, e-mail:
                           \{tyukinivan\}@brain.riken.jp}, Tatiana
                           Tyukina\thanks{Laboratory for Perceptual Dynamics, RIKEN (Institute for Physical and Chemical Research)
                           Brain Science Institute, 2-1, Hirosawa, Wako-shi, Saitama, 351-0198, Japan, e-mail:
                           \{tatianat\}@brain.riken.jp}, Cees
                            van Leeuwen\thanks{{\bf Corresponding author.} Laboratory for Perceptual Dynamics, RIKEN (Institute for Physical and Chemical Research)
                           Brain Science Institute, 2-1, Hirosawa, Wako-shi, Saitama, 351-0198, Japan, e-mail:
                           \{ceesvl\}@brain.riken.jp}
}
\date{\today}
\maketitle{}

\begin{abstract}

We consider the design of a pattern recognition  that matches
templates to images, both of which are spatially sampled and
encoded as temporal sequences.
The image is subject to a combination of various perturbations.
These include ones that can be modeled as parameterized
uncertainties such as image blur, luminance, translation, and
rotation as well as unmodeled ones.  Biological and neural systems
require that these perturbations be processed through a minimal
number of channels by simple adaptation mechanisms. We found that
the most suitable mathematical framework to meet this requirement
is that of weakly attracting sets. This framework provides us with
a normative and unifying solution to the pattern recognition
problem. We analyze the consequences of its explicit
implementation in neural systems. Several properties inherent to
the systems designed in accordance with our normative mathematical
argument coincide with known empirical facts. This is illustrated
in mental rotation, visual search and blur/intensity adaptation.
We demonstrate how our results can be applied to a range of
practical problems in template matching and pattern recognition.
\end{abstract}

\baselineskip 7mm

\section{Notational preliminaries}


We define an image as a mapping $S_0(x,y)$  from a class of
locally bounded mappings $\mathcal{S}\subseteq
L_\infty(\Omega_x\times\Omega_y)$, where $\Omega_x\subseteq\Real$,
$\Omega_y\subseteq\Real$, and ${L}_\infty(\Omega_x\times\Omega_y)$
is the space of all functions $f:\Omega_x \times
\Omega_y\rightarrow\Real$ such that $\|f\|_{\infty}=\ess
\sup\{\|f(x,y)\|, x\in \Omega_x, \ y\in\Omega_y\}<\infty$. Symbols
$x$, $y$ denote variables on different spatial axes. The value of
$S_0(x,y)$ depends on the characteristic value of interest (e.g.
 brightness, contrast, color, etc.). We assume that an image
can be described within the system as a set of a-priory specified
templates, $S_i(x,y)\in\mathcal{S}$,
$i\in\mathcal{I}\subset\Natural$, where $\mathcal{I}$ is the set
of indices of all templates associated with the image
$S_0(x,y)\in\mathcal{S}$. Symbol $\mathcal{I}^+$ is reserved for
$\mathcal{I}^+=\mathcal{I}\cup 0$.

The solution of a system of differential equations
$\dot{\bfx}=\bff(\bfx,t,\thetavec,\bfu(t))$, $\bfu:\Real_{\geq
0}\rightarrow\Real^m$,  $\thetavec\in\Real^d$ passing through
point $\bfx_0$ at $t=t_0$  will be denoted for $t\geq t_0$ as
$\bfx(t,\bfx_0,t_0,\thetavec,\bfu)$, or simply as $\bfx(t)$ if it
is clear from the context what  the values of $\bfx_0,\thetavec$
are and how the function $\bfu(t)$ is defined.

By ${L}^n_\infty[t_0,T]$, $t_0\geq 0$, $T\geq t_0$ we denote the
space of all functions $\bff:\Real_{\geq 0}\rightarrow\Real^n$
such that $\|\bff\|_{\infty,[t_0,T]}=\ess \sup\{\|\bff(t)\|,t \in
[t_0,T]\}<\infty$; $\|\bff\|_{\infty,[t_0,T]}$ stands for the
${L}^n_\infty[t_0,T]$ norm of $\bff(t)$.

Let $\mathcal{A}$ be a set in ${\Real^n}$ and $\|\cdot\|$ be the
usual Euclidean norm in $\Real^n$. By the symbol $\norms{\cdot}$
we denote the following induced norm:
\[
\norms{\bfx}=\inf_{\bfq\in\mathcal{A}}\{\|\bfx-\bfq\|\}
\]
In case $x$ is a scalar and $\Delta\in\Real_{>0}$, notation
$\|x\|_\Delta$ stands for the following
\[
\|x\|_{\Delta}=\left\{
                    \begin{array}{ll}
                    |x|-\Delta, & |x|> \Delta\\
                    0, & |x|\leq \Delta
                    \end{array}
                    \right.
\]

\section{Introduction}

Template matching is the oldest and most common method for
detecting an object in an image. According to this method the
image is searched for items that match a template. A template
consists of one or more local arrays of values representing the
object, e.g. intensity, color, or texture.
Between these templates and certain domains of the image, a
similarity value is calculated\footnote{Traditionally a
correlation measure is commonly used for this purpose
\cite{IEEE_TPAMI:Jain:2000}.}, and a domain is associated with a
template once their similarity exceeds a given threshold.

Despite the simple and straightforward character of this method,
its implementation requires us to consider two fundamental
problems. The first relates to {\it what} features should be
compared between the image $S_0(x,y)$ and the template $S_i(x,y)$,
$i\in\mathcal{I}$. The second problem is {\it how} this comparison
should be done.

The normative answer to the question of {\it what} features should
be compared invokes solving the issue of optimal image
representation, ensuring most effective utilization of available
resources and, at the same time, minimal vulnerability to
uncertainties. Principled solutions to this problem are well-known
from the literature and can be characterized as {\it spatial
sampling}. For example, when the resource is  frequency bandwidth
of a single measurement measurement mechanism, the optimality of
spatially sampled representations is proven in Gabor's seminal
work \cite{Gabor:1946}\footnote{Consider, for instance, a system
which measures image $S_i(x,y)$ using a set of sensors
$\{m_1,\dots,m_n\}$. Each sensor $m_i$ is capable of  measuring
signals within the given frequency band $\Delta_{i}$ at the
location $x_i$ in corresponding spatial dimension $x$. Then
according to \cite{Gabor:1946}, sensor $m_i$ can measure both the
frequency content of a signal and its spatial location with
minimal uncertainty only if the signal has a Gaussian envelope in
$x$: $S_i(x,y)\sim e^{\sigma_i ^{-2}(x-x_i)^2}$. In other words,
the signal should be practically spatially bounded. This implies
that the image must be spatially sampled.}. In classification
problems, the advantage of spatially sampled image representations
is demonstrated in \cite{Ullman:2002}. In general, these
representations are obtained naturally when balancing the system
resources and uncertainties in the measured signal. A simple
argument supporting this claim is provided in \ref{appendix:1}.

\begin{figure}
\begin{center}
\includegraphics[width=450pt]{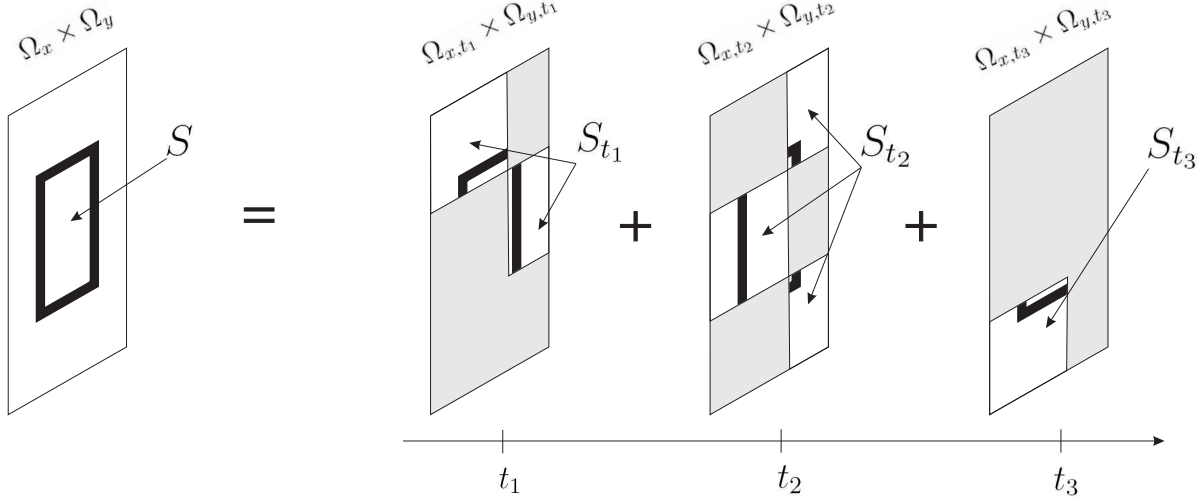}
\caption{Spatial sampling of image
$S(x,y):\Omega_x\times\Omega_y\rightarrow\Real_+$ according to the
factorization of $\Omega_x\times\Omega_y$ into  subsets
$\Omega_{x,t_1}\times\Omega_{y,t_1}$,
$\Omega_{x,t_2}\times\Omega_{y,t_2}$,
$\Omega_{x,t_3}\times\Omega_{y,t_3}$}\label{fig:chap:5:square_in_time}
\end{center}
\end{figure}

The simplest form of spatial sampling can be achieved by
factorizing both the domain $\Omega_x\times\Omega_y$ of the image
$S_0$ and the templates $S_i$, $i\in\mathcal{I}$ into subsets:
\begin{equation}\label{eq:chap:5:image:factorization}
\Omega_x\times\Omega_y=\bigcup_t \Omega_{x,t}\times\Omega_{y,t}, \
t\in \Omega_t, \ \Omega_{x,t}\subseteq \Omega_x, \
\Omega_{y,t}\subseteq \Omega_y.
\end{equation}
Factorization (\ref{eq:chap:5:image:factorization}) induces
sequences $\{S_{i,t}\}$, where $S_{i,t}$ are the restrictions of
mappings $S_i$ to the domains $\Omega_{x,t}\times\Omega_{y,t}$.
These sequences constitute sampled representations of  $S_i$,
$i\in\mathcal{I}^+$ (see Figure \ref{fig:chap:5:square_in_time}).
Notice  that the sampled image and template representations
$\{S_{i,t}\}$ are, strictly speaking, sequences of functions. In
order to compare them, scalar values $f(S_{i,t})$ are normally
assigned to each $S_{i,t}$. Examples include various functional
norms, correlation functions, spectral characterizations (average
frequency or phase), or simply weighted sums of the values of
$S_{i,t}$ over the entire domain $\Omega_{x,t}\times\Omega_{y,t}$.
Formally, $f$ could be defined as a functional, which maps
restrictions $S_{i,t}$ into the field of real numbers:
\begin{equation}\label{eq:spatially_sampled}
\begin{split}
f:& \ L_\infty(\Omega_{x,t}\times\Omega_{y,t})\rightarrow \Real
\end{split}
\end{equation}
This formulation allows a simple representation of images and
templates as sequences of scalar values $\{f(S_{i,t})\}$,
$i\in\mathcal{I}^+$, $t\in\Omega_t$. We will therefore adopt this
method here.

The answer to the second question, that of {\it how}  the
comparison is done, involves finding the best possible and most
simple way to utilize information provided by a given image
representation, at the same time ensuring invariance to basic
distortions. Despite the fact that considerable attention has been
given to this problem, a principled and unified solution is not
yet available. The primary goal of our current contribution is to
present a unified framework to solve this problem for a class of
systems of sufficiently broad theoretical and practical relevance.

\begin{figure}
\begin{center}
\includegraphics[width=470pt]{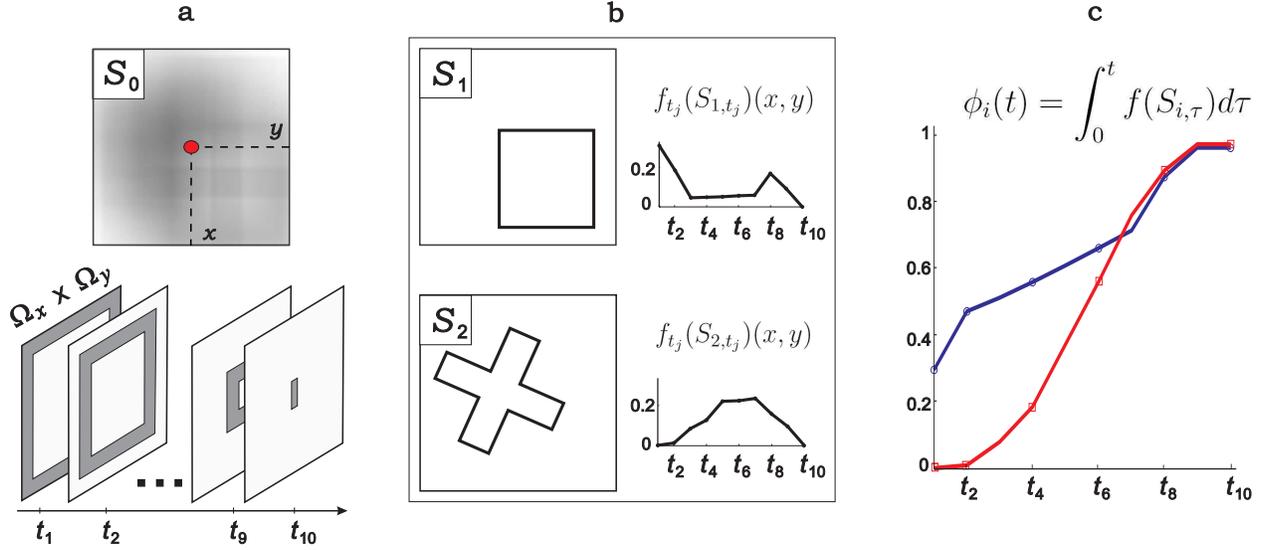}
\caption{ Spatiotemporal image representation via spatial sampling
and temporal integration. Panel $a$ contains original object,
$S_0$; $(x,y)$ marks a point on the image with respect to which
the correlation is calculated; factorization of the domain
$\Omega_x\times\Omega_y$ into ten nonintersecting subsets
$\Omega_x\times\Omega_y=\cup_{j=1}^{10}
\Omega_{x,t_j}\times\Omega_{y,t_{j}}$. Panel $b$ -- templates
$S_1$, $S_2$ and plots of $f_{t_j}(S_{1,t_j})(x,y)$,
$f_{t_j}(S_{2,t_j})(x,y)$ -- the values of the normalized
correlation between
$S_{i,t_j}=S_i(\Omega_{x,t_j}\times\Omega_{y,t_j})$ and
$S_0(\Omega_{x,t_j}\times\Omega_{y,t_j})$. Panel $c$ -- plots of
the values of (\ref{eq:temporally_integral}) as a function of
parameter  $t$ for templates $S_1$ (blue line) and $S_2$ (red
line).}\label{fig:chap:5:space_zone_correlation}
\end{center}
\end{figure}

We consider the class of systems in which spatially sampled image
representations are encoded as temporal sequences. In other words,
parameter $t$ in the notation $f(S_{i,t})$ is the time variable.
This type of representation is frequently encountered in neuronal
networks \cite{Gutig_2006} (see also references therein), and so
such systems have a claim to biological plausibility.  In
addition, they enable a simple solution to a well-known dilemma.
The dilemma is about whether comparison between templates and
image domains should be made on a large, global, or on a small,
local scale. The solution to this dilemma consists in temporal
integration. Let, for instance, $\Omega_t=[0,T]$,
$T\in\Real_{>0}$. Then an example of a temporally-integral, yet
spatially sampled, representation is:
\begin{equation}\label{eq:temporally_integral}
f(S_{i,t})\mapsto \phi_i(t)=\int_{0}^t f(S_{i,\tau})d\tau, \
t\in[0,T], \ i\in\mathcal{I}^+
\end{equation}
The temporal integral $\phi_i(t)$ contains both spatially local
and global image characterizations. Whereas its time-derivative at
$t$ equals to $f(S_{i,t})$ and corresponds to spatially sampled,
local representation $S_{i,t}$, the global representation
$\phi_i(T)$ equals to the integral, cumulative characterization of
$S_i$. An example illustrating these properties is provided in
Figure \ref{fig:chap:5:space_zone_correlation}.
A further advantage of spatiotemporal representations $\phi_i(t)$
is that they offer powerful mechanisms for comparison, processing
and matching of $\phi_i(t)$, $i\in\mathcal{I}$. These mechanisms
can generally be characterized in terms of dynamic oscillator
networks which synchronize when their inputs are converging to the
same function.

Despite advantages such as optimality, simplicity and biological
plausibility, there are theoretical issues which have prevented
wide application of spatiotemporal representations to template
matching. The most important issues, from the authors' viewpoint,
are, first, how to achieve effective recognition in the presence
of modeled disturbances, of which the most common ones are blur,
luminance, and rotational and translational distortion. Second,
how to take into account inevitable unmodeled perturbations.

The first class of problems amounts to finding an
identification/adaptation algorithm capable of reconstructing
parameters of generally nonlinear perturbations. Currently
available approaches either are restricted to linear
parametrization of disturbances, involve overparametrization, or
use domination feedback.
However, linear parametrization is too restricted to be plausible,
overparametrization is expensive in terms of the number of
adjustable units, and domination lacks adequate sensitivity. For
these reasons these methods remain unsatisfactory. The second
class of problems calls for procedures for recognizing an image
from its perturbed temporal representation $\phi_i(t)$. At this
level the system is facing contradictory requirements of ensuring
robust performance while being highly sensitive to minor changes
in the stimulation.

Both these problems  are traditionally dealt with within the
concept of Lyapunov-stable attractors. By allowing the system to
converge on an attractor, it is possible to eliminate modeled and
unmodeled distortions and thus, for instance, complete an
incomplete pattern in the input \cite{Amit85,
Fuchs88,Herz89,Hopfield82, Ritter89}. The advantage of these
methods resides in the robustness inherent in uniform asymptotic
Lyapunov stability. This advantage, however, comes at a cost: such
systems are generally lacking in flexibility. Each stable
attractor represents one pattern; but often an image contains more
than one pattern. When the system is steered to one template, the
other is lost from the representation. It would, therefore, be
preferable to have a system that allows flexible switching between
alternative patterns. Yet, the very notion of stable convergence
to an attractor prevents switching and
exploration across patterns. 
Furthermore, as we will show, for a class of the images with
multiple representations and various symmetries globally stable
solutions to the problem of invariant template matching may not
even exist.

We propose a unifying framework capable of combining robustness
and flexibility. In contrast to common intuitions, which aim at
achieving desired robustness by means of stable attractors, we
advocate instability as an advantageous substitute. We demonstrate
that a specific type of instability, the concept of weakly
attracting sets, provides both the necessary invariance and
flexibility.

\begin{figure}
\centering
\includegraphics[width=370pt]{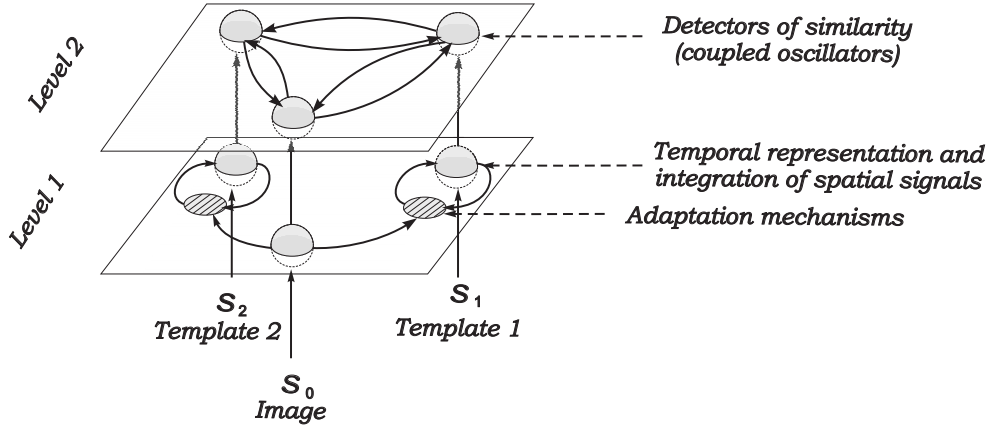}
\caption{General scheme of a system for adaptive template matching
using temporal codes. Level $1$ contains the adaptive
compartments. Its functional role is to ensure invariance to
modeled uncertainties. Level $2$ corresponds to the comparison
compartments and consists of coupled nonlinear oscillators. Solid
arrows represent the information flow in the
system.}\label{fig:chap:5:visual_system}
\end{figure}

To illustrate these principles we designed a recognition system
consisting of two major subcomponents (see Figure
\ref{fig:chap:5:visual_system}).
The first is an adaptive component in which information is
processed by a class of spatiotemporal filters. These filters
represent internal models of distortions. The models of most
common distortions, including rotation, translation, and blur, are
often nonlinearly parameterized. Until recently adequate
compensatory mechanisms  for nonlinear parameterized uncertainties
where unexplored territory. In recent work
\cite{Conf:IFAC_CONGRESS_2005:2} we have shown that the problem of
non-dominating adaptation could, in principle, be solved within
the concept of Milnor or weak, unstable attractors. Here we
provide a solution to this problem that will enable systems to
deal with specific nonlinearly parameterized models of distortions
that are typical for a variety of optical and geometrical
perturbations.

The second component consists of a network of coupled nonlinear
oscillators. These operate as coincidence detectors. Each
oscillator in our system represents a Hindmarsh-Rose model neuron.
These model neurons are generally believed to provide a good
qualitative approximation to biological neuron behavior. At the
same time they are computationally effective in simulations
\cite{IEEE_TNN:Izhikevich:2004}. For networks of these oscillators
we prove, first of all, boundedness of the state of the perturbed
solutions. In addition, we specify the parameter values which lead
to emergence of globally stable invariant manifolds in the system
state space. Although we do not provide explicit criteria for
meta-stability in this class of networks, the conditions presented
allow us to narrow substantially the domain of relevant parameter
values in which this behavior is to be found.



There is an interesting consequence to the unstable character of
the compensation for modeled perturbations. When the system
negotiates multiple classes of uncertainties simultaneously (e.g.
focal/contrast and intensity/luminance), different types of
compensatory adjustments occur at different time scales.
Adaptation at different time scales is a well-known phenomenon in
biological visual systems, in particular when light/dark
adaptation is combined with optical/neuronal blur \cite{Hofer2002,
Matter2006, Mon-Williams98, Rodieck98}; experiments have shown
that combined adaptation processes take place simultaneously but
at different time-scales \cite{Baccus2002, Demontis2002, Sharpe99,
Smirnakis97}. Our analytical study suggests that this difference
in time-scales emerges naturally  as a sufficient condition for
the proper operation of our system.

This paper is organized as follows. In Section
\ref{sec:problem_formulation} we provide a formal description of
the class of images and templates, and formally state the problems
of our study. In Section \ref{sec:main_results} we provide the
main results of our present contribution. In Section
\ref{sec:discussion} we discuss the theoretical results, relate
them to relevant observations in the empirical literature on
visual perception and adaptation, and provide an application of
our approach to  a realistic pattern recognition problem: the
detection of morphological changes in dendritic spines based on
measurements obtained from multiphoton scanning microscope.

\section{Preliminaries and problem
formulation}\label{sec:problem_formulation}


We assume that the values $S_0(x,y)$ of the original image are not
available explicitly to the system; the system is able to measure
only perturbed values of $S_0(x,y)$. Perturbation is defined as a
mapping $\mathcal{F}$:
\[
\mathcal{F}[S_0,\thetavec]: \
L_{\infty}(\Omega_x\times\Omega_y)\times\Real^d\rightarrow
L_{\infty}(\Omega_x\times\Omega_y),
\]
where $\thetavec$ is the vector of parameters of the perturbation.
The values of $\thetavec$ are assumed to be unknown a-priori,
whereas the mapping $\mathcal{F}$ is known.

In systems for processing spatial information, mappings
$\mathcal{F}$ often belong to a specific class that can be defined
as follows:
\begin{equation}\label{eq:perturbation}
\begin{split}
\mathcal{F}[S_0,\thetavec]&=\theta_1 \cdot
\bar{\mathcal{F}}[S_0,\theta_2], \ \theta_1\in\Real, \
\theta_2\in\Real\\
\bar{\mathcal{F}}[S_0,\theta_2]:& \
L_{\infty}(\Omega_x\times\Omega_y)\times\Real\rightarrow
L_{\infty}(\Omega_x\times\Omega_y),\\
\thetavec&=(\theta_1,\theta_2)
\end{split}
\end{equation}
Parameter
$\theta_1\in[\theta_{1,\min},\theta_{1,\max}]\subset\Real$ in
(\ref{eq:perturbation}) models {\it linear} perturbations, for
instance variations of overall brightness or intensity of the
original image $S_0$. It can be interpreted also as an a-priori
unknown gain in the measurement channel of a sensor. Mapping
$\bar{\mathcal{F}}(S_0,\theta_2)$ in (\ref{eq:perturbation}),
parameterized by
$\theta_2\in[\theta_{2,\min},\theta_{2,\max}]\subset\Real$,
corresponds to typical {\it nonlinear} perturbations of image
$S_0$. Table \ref{table:examples_nonlinear_perturbations} provides
examples of these perturbations, their mathematical models and the
physical meaning of parameter $\theta_2$.
\begin{table}[t]
  \centering
  \caption{Examples of typical nonlinear perturbations of $S_0$. Parameter $\Delta_\theta$ in the right column is a positive
  constant}\label{table:examples_nonlinear_perturbations}
\vskip 5mm {\small
  \begin{tabular*}{\textwidth}{@{\extracolsep{\fill}}|c|c|c|}
    \hline
     & & \\
     Physical meaning &  Mathematical model & Domain of \\
     & of $\bar{\mathcal{F}}[S_0,\theta_2]$ & physical relevance\\
    \hline
     & &  \\
     Translation (in $x$ dimension)& $\bar{\mathcal{F}}[S_0,\theta_2]=S_0(x+\theta_2,y)$ & $-\Delta_\theta \leq \theta_2\leq \Delta_\theta $\\
$\theta_2$ -- shift   &  &   \\
  \hline
   & & \\
     Scaling (in $x$ dimension) & $\bar{\mathcal{F}}[S_0,\theta_2]=S_0(\theta_2 \cdot x, y)$ &  $0<\theta_2\leq \Delta_\theta$ \\
$\theta_2$ -- scaling factor &   &  \\
   \hline
 & &  \\
 Rotation & $\bar{\mathcal{F}}[S_0,\theta_2]=S_0(x_r(x,y,\theta_2),y_r(x,y,\theta_2))$ & $0\leq \theta_2\leq 2\pi$ \\
 around the origin  & &  \\
  & $x_r(x,y,\theta_2)=\cos(\theta_2)x-\sin(\theta_2)y$ &  \\
$\theta_2$ -- angle of rotation    & $y_r(x,y,\theta_2)=\sin(\theta_2)x+\cos(\theta_2)y$ &  \\
    &  &  \\
 \hline
  & &  \\
 Image blur \cite{Banham_1997}& $\bar{\mathcal{F}}[S_0,\theta_2]=\int_{\Omega_x\times\Omega_y} h \cdot S_2(\xi,\gamma) d\xi d\gamma$ & $0<\theta_2\leq\Delta_\theta$ \\
 (not normalized)  &  &  \\
$\theta_2$ -- blur parameter &
\begin{minipage}[l]{0.4\linewidth}
 1) Gaussian:
 \[
 h=\exp^{-\frac{1}{\theta_2}((x-\xi)^2+(y-\gamma)^2)}
 \]
  \end{minipage}&  \\
   &  &  \\
 & \begin{minipage}[l]{0.4\linewidth}
 2) Out-of-focus:
 \[
 h=\left\{ \begin{array}{ll}
           \frac{1}{\pi \theta_2^2}, &
           \sqrt{(x-\xi)^2+(y-\gamma)^2}\leq  \theta_2\\
           0, & \mathrm{else}
          \end{array}
 \right.
 \]
   \end{minipage}
 &  \\
   &  &  \\
 \hline
  \end{tabular*}
}
\end{table}
Throughout the paper we assume that mappings
$\bar{\mathcal{F}}[S_0,\theta_2]$ are Lipschitz in $\theta_2$:
\begin{equation}\label{eq:lipshcitz_constraint}
\begin{split}
\exists \ D\in\Real_{>0}: \ \
\left|\bar{\mathcal{F}}[S_0,\theta_2'](x,y)-\bar{\mathcal{F}}[S_0,\theta_2''](x,y)\right|&\leq
D |\theta_2'-\theta_2''|, \\
 \forall \ (x,y)\in\Omega_x\times\Omega_y, & \
\theta_2',\theta_2''\in\Real
\end{split}
\end{equation}
Notice that, strictly speaking, several typical transformations
such as translation, scaling, and rotation, are not always
Lipschitz. This is because image $S_0$ can, for instance, have
sharp edges which corresponds to discontinuities in $x,y$. In
practice, however, prior application of a blurring linear filter
will render sharp edges in an image smooth, thus assuring that
condition (\ref{eq:lipshcitz_constraint}) applies\footnote{In
biological vision discontinuity of $S_0$ in $x,y$ corresponds to
images with abrupt local changes in brightness along spatial
dimensions $x,y$. Although this is a rather common situation in
nature, in visual systems actual images $S_0$ rarely reach a
sensor in their spatially discontinuous form. In fact, prior to
reaching the sensory part, they are subject to linear filtering
induced by optics. Therefore the images that reach the sensor are
always smooth. Hence condition (\ref{eq:lipshcitz_constraint})
will generally be satisfied.}.

The image $\mathcal{F}[S_0,\thetavec]$ is assumed to be spatially
sampled according to factorization
(\ref{eq:chap:5:image:factorization}):
\begin{equation}\label{eq:sampling_def}
\mathcal{F}_t [S_0,\thetavec] (x,y)= \left\{\begin{array}{ll}
                                  \mathcal{F} [S_0,\thetavec] (x,y),&
                                  (x,y)\in\Omega_{x,t}\times\Omega_{y,t},\\
                                  0,  & \mathrm{else}
                                  \end{array}
                                  \right. \  \ t\in\Omega_t
\end{equation}
Because index $t$ in (\ref{eq:sampling_def}) is assumed to be a
time variable we let $\Omega_t=[0,\infty)$. To each $\mathcal{F}_t
[S_0,\thetavec]$ a value $f(\mathcal{F}_t
[S_0,\thetavec])\in\Real$ is assigned. Formally this procedure can
be defined by a functional which maps mappings $\mathcal{F}_t
[S_0,\thetavec]$ into the real values:
\begin{equation}\label{eq:sampling_functional_def}
f: L_\infty(\Omega_x\times\Omega_y)\rightarrow\Real.
\end{equation}
In the singular case, when $\Omega_{x,t}\times\Omega_{y,t}$ is a
point $(x_t,y_t)$, the mapping $\mathcal{F}_t [S_0,\thetavec]
(x,y)$ and functional $f$ will be defined as  $f=\mathcal{F}_t
[S_0,\thetavec] (x_t,y_t)= \mathcal{F} [S_0,\thetavec] (x_t,y_t)$.

We concentrated our efforts on obtaining a principled solution to
the problem of invariant template matching  in systems with
spatiotemporal processing of information. For this reason we
prefer not to provide a specific description of functionals $f$.
We do, however, restrict our consideration to {\it linear} and
Lipschitz functionals, e.g. the functionals satisfying the
following constraints:
\begin{equation}\label{eq:sampling_functional}
f(\kappa  \mathcal{F})=\kappa f(\mathcal{F}), \ \forall \
\kappa\in\Real, \ \  \left|f(\mathcal{F}
)-f(\mathcal{F}')\right|\leq D_2 \|\mathcal{F}
-\mathcal{F}'\|_\infty, \  \ D_2\in\Real_{>0}
\end{equation}
Examples of functionals $f$ satisfying conditions
(\ref{eq:sampling_functional}) and their physical interpretations
are provided in Table \ref{table:examples_sampling}.
\begin{table}[t]
  \centering
  \caption{Examples of spatially-sampled representations of $S_0$}\label{table:examples_sampling}
\vskip 5mm {\small
  \begin{tabular*}{\textwidth}{@{\extracolsep{\fill}}|c|c|}
    \hline
     &  \\
     Physical meaning &  Mathematical model of $f$ \\
     & \\
    \hline
     & \\
     Spectral power within & \\
  the given frequency bands:& $f=\int_{\omega_{a}}^{\omega_{b}}\int_{\omega_{c}}^{\omega_{d}}\left\|\int_{\Omega_x\times\Omega_y}\mathcal{F}_t[S_0,\thetavec](x,y)
e ^{-j(\omega_x x+\omega_y y)}dx dy\right\| d\omega_x d\omega_y$ \\
$\omega_x\in[\omega_{a},\omega_{b}]$, $\omega_{y}\in[\omega_{c},\omega_{d}]$  & \\
 &   \\
  \hline
   & \\
   Weighted sum & $f=\int_{\Omega_x\times\Omega_y}\mathcal{F}_t[S_0,\thetavec](x,y)e^{-|x-x_0|-|y-y_0|}dx dy$  \\
 (for instance, convolution &   \\
with exponential kernel)& $(x_0,y_0)$ is the reference, ``attention'' point \\
 &  \\
   \hline
 &  \\
 Scanning the image & $\Omega_{x,t}\times\Omega_{y,t}=(\xi(t),\gamma(t))$ \\
 along a given trajectory & \\
 $(x(t),y(t))=(\xi(t),\gamma(t))$  & $f=\mathcal{F}[S_0,\thetavec](\xi(t),\gamma(t))$  \\
    &   \\
 \hline
   \end{tabular*}
}
\end{table}

Taking into account (\ref{eq:perturbation}),
(\ref{eq:sampling_def}) and the fact that $f$ is linear, the
following equality holds
\begin{equation}\label{eq:sampling_functional_final_def}
f(\mathcal{F}_t [S_0,\thetavec])=\theta_1
f(\bar{\mathcal{F}}_t[S_0,\theta_2]), \ \
\bar{\mathcal{F}}_t[S_0,\theta_2]=\left\{
                                      \begin{array}{ll}
                                \bar{\mathcal{F}} [S_0,\theta_2] (x,y),&
                                  (x,y)\in\Omega_{x,t}\times\Omega_{y,t},\\
                                  0,  & \mathrm{else}
                                      \end{array}
                                      \right.
\end{equation}
For the sake of compactness, in what follows  we replace
$f(\bar{\mathcal{F}}_t[S_0,\theta_2])$ in the definition of
$f(\mathcal{F}_t [S_0,\thetavec])$ in
(\ref{eq:sampling_functional_final_def}) with the following
notation
\begin{equation}\label{eq:sampling_notation}
f(\bar{\mathcal{F}}_t[S_0,\theta_2])=f_0(t,\theta_2), \
f_0:\Omega_t\times\Real\rightarrow \Real
\end{equation}
Notation $f_0(t,\theta_2)$ in (\ref{eq:sampling_notation}) allows
us to emphasize the dependence of $f$ on unknown $\theta_2$, time
variable $t$, and original image $S_0$. Subscript ``$0$'' in
(\ref{eq:sampling_notation}) indicates that $f_0(t,\theta_2)$
corresponds to the sampled and perturbed $S_0$ (equations
(\ref{eq:perturbation}), (\ref{eq:sampling_functional_def}),
(\ref{eq:sampling_functional})), and  argument $\theta_2$ is the
nonlinear parameter of the perturbation applied to the image.
Adhering to this logic, we introduce  the notation
\[
f(\mathcal{F}_t [S_i,\thetavec])=\theta_1
f(\bar{\mathcal{F}}_t[S_i,\theta_2])=\theta_1 f_i(t,\theta_2),
\]
where subscript ``$i$'' indicates that $f_i(t,\theta_2)$
corresponds to the perturbed and sampled template $S_i$, and
$\theta_2$ is the nonlinear parameter of the perturbation applied
to the template.

Let us now specify the class of schemes realizing temporal
integration of spatially sampled image representations. Explicit
realization of temporal integration (\ref{eq:temporally_integral})
is not optimal because it might lead to unbounded outputs for a
wide class of relevant signals, for instance signals that are
constant or periodic with a nonzero average. The behavior of a
temporal integrator (\ref{eq:temporally_integral}), however, can
be fairly well approximated by a first-order linear filter. For
the sampled image and template representations $\theta_1
f_i(t,\theta_2)$, these filters can be defined as follows:
\begin{equation}\label{eq:integral_filter}
\begin{split}
\dot{\phi}_0&=-\frac{1}{\tau}\phi_0 + k\cdot\theta_1
f_0(t,\theta_2)\\
\dot{\phi}_i&=-\frac{1}{\tau}\phi_i  + k\cdot \theta_1
f_i(t,\theta_2), \ k,\tau\in\Real_{>0}, \ i\in\mathcal{I}
\end{split}
\end{equation}
In  contrast to (\ref{eq:temporally_integral}),  for filters
(\ref{eq:integral_filter}) it is ensured that their state remains
bounded for bounded inputs. In addition, on a first approximation,
equations (\ref{eq:integral_filter}) present a simple model of
neural sensors, collecting and encoding spatially-distributed
information in the form of a function of time\footnote{In
principle, equation (\ref{eq:integral_filter}) can be replaced
with a more plausible model of temporal integration such as
integrate-and-fire, Fitzhugh-Nagumo, or Hodgkin-Huxley model
neurons. These extensions, however, are not immediately relevant
for the purpose of our current study. Therefore for the sake of
clarity we decided to keep the mathematical description of the
system as simple as possible, keeping in mind the possibility of
extension to a wider class of temporal integrators
(\ref{eq:integral_filter}).}. With respect to the physical
realizability of (\ref{eq:integral_filter}), in addition to
requirements (\ref{eq:lipshcitz_constraint}),
(\ref{eq:sampling_functional}) we shall only assume that spatially
sampled representations $\theta_1 f_i(t,\theta_2)$,
$i\in\mathcal{I}^{+}$ of $S_i$ ensure the existence of solutions
for system (\ref{eq:integral_filter}).

Consider the dynamics of variables $\phi_0(t)$ and $\phi_i(t)$,
$i\in\mathcal{I}$ defined by (\ref{eq:integral_filter}). We say
that the $i$-th template matches the image iff for some given
$\varepsilon\in\Real_{\geq 0}$ the following condition holds
\begin{equation}\label{eq:invariance_goals}
\limsup_{t\rightarrow\infty} |\phi_0(t)-\phi_i(t)|\leq
\varepsilon.
\end{equation}
The problem, however, is that parameters $\theta_1$, $\theta_2$ in
(\ref{eq:integral_filter}) are unknown a-priori. While
perturbations affect the image directly, they do not necessarily
influence the templates. Rather to the contrary, for consistent
recognition the templates are better kept isolated from external
perturbations --  at least within the time frame of pattern
recognition, although they may, in principle, be affected by
adaptive learning on a larger time scale. Having fixed, unmodified
templates in comparison with perturbed image representations
implies that even in the cases when objects corresponding to the
templates are present in the image, temporal image representation
$\phi_0(t)$ will likely be different from any of the templates,
$\phi_i(t)$. This will render the chances  that condition
(\ref{eq:invariance_goals}) is satisfied very small, so a template
would almost never be detected in an image.

We propose that the proper way for the system to meet requirement
(\ref{eq:invariance_goals}) is to mimic the effect of disturbances
in the template. In order to achieve this template matching system
should be able to track the unknown values of parameters
${\theta}_1$, $\theta_2$. Hence the original equations for
temporal integration (\ref{eq:integral_filter}) will be replaced
with the following
\begin{equation}\label{eq:temporal_coding}
\begin{split}
\dot{\phi}_0&=-\frac{1}{\tau}\phi_0 + k\cdot\theta_1
f_0(t,\theta_2)\\
\dot{\phi}_i&=-\frac{1}{\tau}\phi_i  + k\cdot \hat{\theta}_{i,1}
f_i(t,\hat{\theta}_{i,2}), \ k,\tau\in\Real_{>0}, \
i\in\mathcal{I}
\end{split}
\end{equation}
where $\hat{\theta}_{i,1}$, $\hat{\theta}_{i,2}$ are the estimates
of $\theta_1$, $\theta_2$. The estimates $\hat{\theta}_{i,1}$,
$\hat{\theta}_{i,2}$ must track instantaneous changes of
${\theta}_1$, $\theta_2$. The information  required for such an
estimation should be kept at the minimal possible level. An
acceptable solution would be a simple mechanism capable of
tracking the perturbations from the measurements of the image
alone. The formal statement of this problem is provided below:

\begin{problem}[Invariance]\label{problem:invariance} For a given image $S_0$, template
$S_i$, and their spatiotemporal representations satisfying
(\ref{eq:lipshcitz_constraint}), (\ref{eq:sampling_functional}),
and (\ref{eq:temporal_coding}), find estimates
\begin{equation}\label{eq:algorithms_goal}
\hat{\theta}_{i,1}=\hat{\theta}_{i,1}(t,\tau,\kappa,\phi_0,\phi_i),
\ \
\hat{\theta}_{i,2}=\hat{\theta}_{i,2}(t,\tau,\kappa,\phi_0,\phi_i)
\end{equation}
as functions of time $t$, variables $\phi_0$, $\phi_i$ and
parameters $\tau$, $\kappa$ such that for all possible values of
parameters $\theta_1\in[\theta_{1,\min},\theta_{1,\max}]$,
$\theta_2\in[\theta_{2,\min},\theta_{2,\max}]$

1) solutions of system (\ref{eq:temporal_coding}) are bounded;

2) in case $f_0=f_i$ property (\ref{eq:invariance_goals}) is
ensured, and

3) the following holds for some
$\theta_1'\in[\theta_{1,\min},\theta_{1,\max}]$,
$\theta_2'\in[\theta_{2,\min},\theta_{2,\max}]$:
\begin{equation}\label{eq:invariance_goals_2}
\begin{split}
&
\limsup_{t\rightarrow\infty}|\hat{\theta}_{i,1}(t,\tau,\kappa,\phi_0(t),\phi_i(t))-\theta_{i,1}'|\leq
\varepsilon_{\theta,1}, \ \varepsilon_{\theta,1}\in\Real_+ \\
&
\limsup_{t\rightarrow\infty}|\hat{\theta}_{i,2}(t,\tau,\kappa,\phi_0(t),\phi_i(t))-\theta_{i,2}'|\leq
\varepsilon_{\theta,2}, \ \varepsilon_{\theta,2}\in\Real_+
\end{split}
\end{equation}
\end{problem}

Once the solution to Problem \ref{problem:invariance} is found,
the next step will be to ensure that similarities
(\ref{eq:invariance_goals}) are registered in the system.
Following the spirit of neural systems and in agreement with the
structure in Figure \ref{fig:chap:5:visual_system}, we propose
that detection of similarities is realized by a system of coupled
oscillators. In particular, we require that states of oscillators
$i$ and $0$ converge as soon as the signals $\phi_0(t)$,
$\phi_i(t)$ become sufficiently close.

In the present article we restrict ourselves to the class of
systems composed of linearly coupled Hindmarsh-Rose model neurons
\cite{ProcRSL:Hindmarsh_and_Rose:1984}. This choice is motivated
by the fact that these oscillators can reproduce a broad class of
behaviors observed in real neurons while being computationally
efficient \cite{IEEE_TNN:Izhikevich:2004}. A network of these
neural oscillators can be mathematically described  as follows:
\begin{equation}\label{HR_model_net}
\mathcal{S}_{D_i}: \
 \left\{
     \begin{array} {ll}
            \dot{x}_i &= -a{x}_i^3+b{x}_i^2+y_i-z_i+I+u_i + \phi_i(t),\\
            \dot{y}_i &= c-d{x}_i^2-y_i,\\
            \dot{z}_i &= \eps(s(x_i+x_0)-z_i),
     \end{array}
 \right.  \ i\in\mathcal{I}^+
\end{equation}
Variables $x_i$, $y_i$, $z_i$ correspond to  membrane potential,
and aggregated fast and slow adaptation currents, respectively.
Coupling $u_i$ in (\ref{HR_model_net}) is assumed to be linear and
symmetric:
\begin{equation}\label{eq:chap:5:image:coupling}
      \bfu=\left(
         \begin{array}{c}
        u_0\\
        u_1\\
        \vdots\\
        u_n
         \end{array}
        \right)=\Gamma \left(
                         \begin{array}{c}
                            x_0\\
                            x_1\\
                            \vdots\\
                            x_n
                          \end{array}
                       \right),
                                \ \Gamma=\gamma\left(
                                                \begin{array}{cccc}
                                                 - n & 1 & \cdots & 1\\
                                                  1 & -n & \cdots & 1\\
                                               \cdots & \cdots & \cdots & \cdots\\
                                                  1 & 1 & \cdots & -n\\
                                                \end{array}
                                         \right),
\end{equation}
and parameter $\gamma\in\Real_+$.  Our choice of the coupling
function in (\ref{eq:chap:5:image:coupling}) is motivated by the
following considerations. Fist, we wish to preserve the intrinsic
dynamics of the neural oscillators when they synchronize, e.g.
when $x_i=x_j$, $y_i=y_j$, $z_i=z_j$, $i,j\in\{0,\dots,n\}$. For
this reason it is desirable that the coupling vanishes when the
synchronous state is reached. Second, we seek for a system in
which synchronization between two arbitrary nodes, say the $i$-th
and the $j$-th nodes, is determined exclusively by the degree of
(mis)matches in $\phi_i(t)$, $\phi_j(t)$, independently of the
activity of other units in the system. Third, the coupling should
``pull'' the system trajectories towards the synchronous state.
Coupling function (\ref{eq:chap:5:image:coupling}) satisfies all
these requirements.

We set parameters of  equations (\ref{HR_model_net}) to the
following values:
\begin{equation}\label{eq:HR_model_param}
\begin{array}{cccc}
 a = 1, &  b = 3, &  c = 1, &  d = 5, \\
s = 4, &  x_0 = 1.6, & \eps = 0.001, &
\end{array}
\end{equation}
which correspond to the regime of chaotic bursting in each
uncoupled element in (\ref{HR_model_net}) \cite{PRL:Hansel:1992}.

The problem of detection of similarities in $\phi_0(t)$ and
$\phi_i(t)$ can now be stated as follows.

\begin{problem}[Detection]\label{problem:detection} Let system
(\ref{HR_model_net}), (\ref{eq:chap:5:image:coupling}) be given
and there exist $i\in\mathcal{I}$ such that condition
(\ref{eq:invariance_goals}) is satisfied. Determine the coupling
parameter $\gamma$ as a function of system (\ref{HR_model_net})
parameters such that

1) solutions of the system are bounded for all bounded $\phi_i$,
$i\in\mathcal{I}$;

2) states $(x_0(t),y_0(t),z_0(t))$ and $(x_i(t),y_i(t),z_i(t))$
asymptotically converge to a vicinity of the synchronization
manifold $x_0=x_i$, $y_0=y_i$, $z_0=z_i$. In particular,
\[
\begin{split}
& \limsup_{t\rightarrow\infty} |x_0(t)-x_i(t)|\leq
\delta(\varepsilon)\\
& \limsup_{t\rightarrow\infty}
|y_0(t)-y_i(t)|\leq \delta(\varepsilon)\\
& \limsup_{t\rightarrow\infty} |z_0(t)-z_i(t)|\leq
\delta(\varepsilon),
\end{split}
\]
where $\delta(\cdot)$ is a non-decreasing function vanishing at
zero.
\end{problem}

In the next section we present solutions to the problems of
invariance and detection. We start from considerations of what
would be the most adequate concept of analysis. Our considerations
will lead us to conclusion that for solving the problem of
invariance, using the concept of Milnor attractors is advantageous
over traditional concepts resting on the notion of Lyapunov
stability. This implies that the sets to which the estimates
$\hat{\theta}_{i,1}$, $\hat{\theta}_{i,2}$ converge should be
weakly attracting rather than Lyapunov stable. We present a simple
mechanism realizing this requirement for a wide class of models of
disturbances. With respect to the second problem, the problem of
detection, we provide sufficient conditions for asymptotic
synchrony in system (\ref{HR_model_net}).

\section{Main Results}\label{sec:main_results}

Consider a system of temporal integrators,
(\ref{eq:temporal_coding}), in which the template subsystem
(second equation in (\ref{eq:temporal_coding})) is designed to
mimic the temporal code of an image using adjustment mechanisms
(\ref{eq:algorithms_goal}). Ideally, the template subsystem should
have a single adjustment mechanism, which is structurally simple
and yet capable of handling a broad class of perturbations. In
addition it should require the least possible amount of a-priori
information about images and templates.

To search for a possible adaptation mechanism let us first explore
the available theoretical concepts which can be used in its
derivation. The problem of invariance, as stated in Problem
\ref{problem:invariance}, can generally be understood as a
specific optimization task. Particular solutions to such tasks as
well as choice of the appropriate mathematical tools depend
significantly on the following characteristics:  uniqueness of the
solutions, convexity with respect to parameters, and sensitivity
to the input data (images and templates). Let us consider wether
the invariant template matching problem meets these requirements.

\begin{figure}
\begin{center}
\includegraphics[width=400pt]{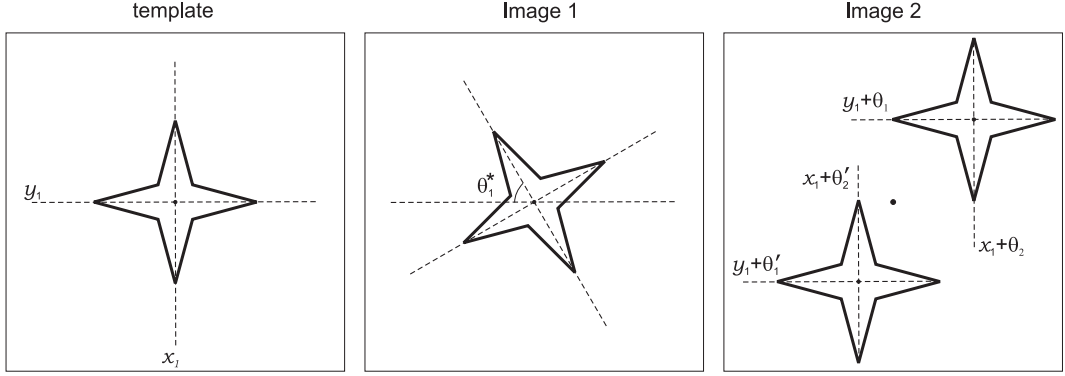}
\caption{Example of a template and images which lead to non-unique
solutions in the problem of invariant template matching. Image 1
is a rotated version of the template. Because the template has
rotational symmetry, the angles
$\theta_2=\theta_2^\ast\pm\frac{\pi}{2}n$, $n=0,1,\dots$ at which
the template and the image match to each other are not unique.
Image 2 contains two multiple instances of the template, which
also leads to non-uniqueness.}\label{fig:non-uniqueness_star}
\end{center}
\end{figure}

{\it Uniqueness.} Solutions to the problem of invariant template
matching are  generally {\it not unique}. The image may contain
multiple instances of the template. Even if there is only a single
unique object the template may fit it in multiple ways, for
instance because it has rotational symmetry. Both cases are
illustrated in Figure \ref{fig:non-uniqueness_star}.
A similar argument applies to translational invariance in the
images with multiple instances of the template (right picture in
Figure \ref{fig:non-uniqueness_star}).

{\it Non-linearity and non-convexity.} The problem of invariant
template matching is generally {\it nonlinear and nonconvex in
$\theta_1$, $\theta_2$}. The nonlinearity is already evident from
Table \ref{table:examples_nonlinear_perturbations}. To illustrate
the potential non-convexity consider, for instance, the following
function
\begin{equation}\label{eq:example_nonconvex}
\theta_1 f_i(t,\theta_2)=\theta_1
\int_{\Omega_{x,t}\times\Omega_{y,t}} e^{-|x-x_0|-|y-y_0|}
\left(\int_{\Omega_{x}\times\Omega_{y}} e^{-\frac{1}{\theta_2}
((x-\xi)^2+(y-\gamma)^2} S_i(\xi,\gamma)d\xi d\gamma\right) dx dy
\end{equation}
which is a composition of the Gaussian blur model (the forth row
in Table \ref{table:examples_nonlinear_perturbations}) with
spatial sampling and subsequent exponential weighting (the second
row in Table \ref{table:examples_sampling}). In the literature on
adaptation two versions of the convexity requirement are
available. The first version applies to the case where the
difference $\theta_1 f_i(t,\theta_2) - \hat{\theta}_{1,i}
f_i(t,\hat{\theta}_{i,2})$ is not accessible for explicit
measurement, and the variables $\phi_0(t)$, $\phi_i(t)$ should be
used instead. In this case the convexity condition will have the
following form  \cite{Avtoma_Telemech:Fradkov-79}:
\begin{equation}\label{eq:convexity_visual}
\begin{split}
e_i(\phi_0,\phi_i)&\left[(\theta_1-\hat{\theta}_{i,1})\frac{\pd}{\pd
\hat{\theta}_{i,1}}\hat{\theta}_{i,1}
f_i(t,\hat{\theta}_{i,2})+(\theta_2-\hat{\theta}_{i,2})\frac{\pd}{\pd
\hat{\theta}_{i,2}}\hat{\theta}_{i,1} f_i(t,\hat{\theta}_{i,2})\right]\geq\\
& e_i(\phi_0,\phi_i)\left[\theta_1 f_i(t,\theta_2) -
\hat{\theta}_{i,1} f_i(t,\hat{\theta}_{i,2}) \right]
\end{split}
\end{equation}
Term $e_i(\phi_0,\phi_i)$ in (\ref{eq:convexity_visual}) is
usually the difference $e_i(\phi_0,\phi_i)=\phi_0-\phi_i$ and has
the meaning of error. For the same pairs of points $\theta_1,
\theta_2$ and $\hat{\theta}_{i,1}$, $\hat{\theta}_{i,2}$ condition
(\ref{eq:convexity_visual}) may hold of fail depending on the sign
of $e_i(\phi_0(t),\phi_i(t))$ at the particular time instance $t$.
Hence it is not always satisfied, not even for convex
${\theta}_{i,1} f_i(t,{\theta}_{i,2})$.

The second version of the convexity requirement applies when the
difference $\theta_1 f_i(t,\theta_2) - \hat{\theta}_{i,1}
f_i(t,\hat{\theta}_{i,2})$ can be measured explicitly. In this
case the condition is formulated as  definiteness of the Hessian
of $\theta_1 f_i(t,\theta_2)$. It can easily be  verified,
however, that this depends, for instance,  on the values of
$S_i(\xi,\gamma)$ in (\ref{eq:example_nonconvex}). Hence both
versions of the convexity conditions generally fail in invariant
template matching.

{\it Critical dependence on stimulation.} An important feature of
of invariant template matching problem is that its solutions {\it
critically depend} on particular images and templates. Presence of
rotational symmetries in the templates affect the number of
solutions. Hence objects with different number of symmetries will
be characterized by sets of solutions with different cardinality.

We conclude that the problem of invariant template matching
generally assumes multiple alternative solutions, nonlinearity and
non-convexity with respect to parameters, and the structure of its
solutions depends critically on a-priori unknown stimulation. What
would be a suitable way to approach this class of problems in a
principled manner?

Traditionally, processes of matching and recognition are
associated with convergence of the system's state to an attracting
set. In our case the system's state is defined by vector $\bfx$:
\[
\bfx=(\phi_0,\phi_1,\dots,\phi_i,\dots,\hat{\theta}_{1,1},\hat{\theta}_{2,1},
\dots\hat{\theta}_{i,1},\hat{\theta}_{i,2},\dots)
\]
The attracting set, $\mathcal{A}$, is normally understood as a set
satisfying the following definition \cite{Guckenheimer:2002}:
\begin{defn}\label{defn:attracting_set} A set $\mathcal{A}$ is an attracting set iff it is

i) closed, invariant, and

ii) for some neighborhood $\mathcal{V}$ of $\mathcal{A}$ and for
all $\bfx_0\in\mathcal{V}$ the following conditions hold:
\begin{equation}\label{eq:attracting_set}
\bfx(t,\bfx_0)\in\mathcal{V} \ \forall \ t\geq 0;
\end{equation}
\begin{equation}\label{eq:attracting_limit}
\lim_{t\rightarrow\infty}\norms{\bfx(t,\bfx_0)}=0
\end{equation}
\end{defn}
Traditional techniques for proving attractivity employ the concept
of Lyapunov asymptotic stability\footnote{We recall that the set
$\mathcal{A}$ is (globally) Lyapunov asymptotically stable iff for
all $\varepsilon>0$ there exists $\delta(\bfx_0,\varepsilon)>0$
such that $\norms{\bfx_0}<\delta(\bfx_0,\varepsilon)$ $\Rightarrow
\ \norms{\bfx(t,\bfx_0)}\leq \varepsilon$ for all $t\geq 0$, and
$\lim_{t\rightarrow\infty}\norms{\bfx(t,\bfx_0)}=0$}. Although the
notion of set attractivity is wider, the method of Lyapunov
functions is constructive and, in addition, Lyapunov asymptotic
stability implies the desired attractiviy. For these reasons it is
highly practical, and the tandem of set attractivity in Definition
\ref{defn:attracting_set} and Lyapunov stability has been used
extensively in recognition systems, including Hopfield networks,
recurrent neural nets, etc.

The problem of {\it invariant} template matching, however,
challenges the universality of these concepts. First, because of
inherent non-uniqueness of the solutions, there are multiple
invariant sets in the system's state space. Hence, global Lyapunov
asymptotic stability cannot be ensured. Second, when each solution
is treated as a locally stable invariant set, it is essentially
important to know the domain of its attractivity. This domain,
however, depends on properties of function $\theta_1
f_0(t,\theta_2)$ in (\ref{eq:temporal_coding}) which vary with
stimulation. Third, no method exists for solving Problem
\ref{problem:invariance} for general nonlinearly parameterized
$\theta_1 f_0(t,\theta_2)$ that assures Lyapunov stability of the
system.

In order to solve the problem of invariant template matching we
therefore propose to replace the standard notion of attracting set
with a less restrictive concept. In particular we use the concept
of weak or Milnor attracting sets \cite{CommMathPhys:Milnor:1985}:
\begin{defn}\label{defn:Milnor_attracting_set}
A set $\mathcal{A}$ is {weakly attracting}, or Milnor attracting
set iff

i) it is closed, invariant and

ii) for some set $\mathcal{V}$ (not necessarily a neighborhood of
$\mathcal{A}$) with {\it strictly positive measure} and  for all
$\bfx_0\in\mathcal{V}$ limiting relation
(\ref{eq:attracting_limit}) holds
\end{defn}

\begin{figure}
\begin{center}
\includegraphics[width=400pt]{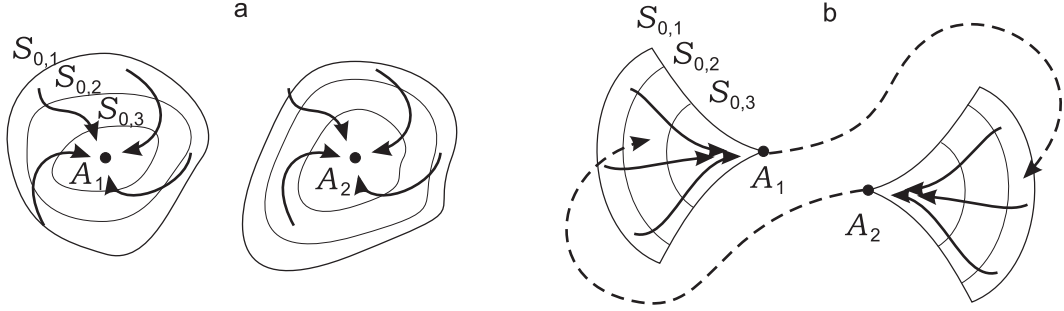}
\caption{Standard stable attractors, panel $a$, vs weak
attractors, panel $b$. Domains of stable attractors are
neighborhoods containing $\mathcal{A}_1$, $\mathcal{A}_2$.
Estimates of sizes of these domains depend on particular images
$S_{0,1}$, $S_{0,2}$, $S_{0,3}$. These estimates are depicted as
closed curves around $\mathcal{A}_1$, $\mathcal{A}_2$. Once the
state converges to either of the attractors it stays there unless,
probably, when the image changes. In contrast to this, domains of
attraction for Milnor attracting sets are not neighborhoods.
Hence, even a slightest perturbation in the image induces a finite
probability of escape from the attractor. Hence multiple
alternative representations of the image could eventually be
restored.}\label{fig:non-uniqueness_Lyapunov_vs_Milnor}
\end{center}
\end{figure}

The main difference between the notions of a weak attracting set,
Definition \ref{defn:Milnor_attracting_set}, and the standard one,
Definition \ref{defn:attracting_set}, is that the domain of
attraction is not required to be a neighborhood of $\mathcal{A}$.
On the one hand, this allows to us use mathematical tools beyond
the concept of Lyapunov stability in order to avoid problems with
nonlinear parametrization and critical dependance on stimulation.
On the other hand, it offers a natural mechanism for systems to
explore multiple image representations. This is illustrated in
Figure \ref{fig:non-uniqueness_Lyapunov_vs_Milnor}.

In the next paragraph we present technical details of how Problem
\ref{problem:invariance} could be solved within the framework of
Milnor attractors.

\subsection{Invariant template matching by Milnor attractors}

Consider system (\ref{eq:temporal_coding}):
\[
\begin{split}
\dot{\phi}_0&=-\frac{1}{\tau}\phi_0 + k\cdot\theta_1
f_0(t,\theta_2)\\
\dot{\phi}_i&=-\frac{1}{\tau}\phi_i  + k\cdot \hat{\theta}_{i,1}
f_i(t,\hat{\theta}_{i,2}), \ k,\tau\in\Real_{>0}, \
i\in\mathcal{I}
\end{split}
\]
and assume that the $i$-th  template is present in the image. This
implies that both the image and the template will have, at least
locally in space, sufficiently similar spatiotemporal
representations. Formally this can be stated as follows:
\begin{equation}\label{eq:Delta}
\exists \ \Delta\in\Real_{>0}: \  \ \left|\theta_1
f_0(t,\theta_2)-\theta_1 f_i(t,{\theta}_{2})\right|\leq \Delta, \
\ \forall \ \theta_1,\theta_2,  \ t\geq 0
\end{equation}
Hence without loss of generality we can replace equations
(\ref{eq:temporal_coding}) with the following
\begin{equation}\label{eq:invariance_temporal_coding}
\begin{split}
\dot{\phi}_0&=-\frac{1}{\tau}\phi_0 + k\cdot\theta_1
f_i (t,\theta_2) + \epsilon(t)\\
\dot{\phi}_i&=-\frac{1}{\tau}\phi_i  + k\cdot \hat{\theta}_{i,1}
f_i(t,\hat{\theta}_{i,2}), \ k,\tau\in\Real_{>0}, \
i\in\mathcal{I}
\end{split}
\end{equation}
where $\epsilon(t)\in L_\infty[0,\infty]$,
$\|\epsilon(t)\|_{\infty}\leq \Delta$ is a bounded disturbance.
Solving Problem \ref{problem:invariance}, therefore, amounts to
finding adjustment mechanisms (\ref{eq:algorithms_goal}) such that
trajectories $\phi_0(t)$, $\phi_i(t)$ in
(\ref{eq:invariance_temporal_coding}) converge and limiting
relations (\ref{eq:invariance_goals_2}) hold.

The main idea of our proposed solution to this problem can
informally be summarized as follows. First, we introduce an
auxiliary system
\begin{equation}\label{eq:extension_visual}
\dot{\lambdavec}=\bfg(\lambdavec,\phi_0,\phi_i,t),  \
\lambdavec\in\Real^\lambda, \
\bfg:\Real^\lambda\times\Real\times\Real\times\Real_{\geq
0}\rightarrow\Real^\lambda
\end{equation}
and define $\hat{\theta}_{i,1}$, $\hat{\theta}_{i,2}$ as functions
of  $\lambdavec$, $\phi_0$, and $\phi_i$:
\begin{equation}\label{eq:adjustment_visual}
\hat{\theta}_{i,1}=\hat{\theta}_{i,1}(\lambdavec,\tau,\kappa,\phi_0,\phi_i),
\
\hat{\theta}_{i,2}=\hat{\theta}_{i,2}(\lambdavec,\tau,\kappa,\phi_0,\phi_i).
\end{equation}
Second, we show that  for some $\varepsilon\in\Real_{>0}$, and
$\Omega_\lambda\subset\Real^\lambda$  the following set
\[
\Omega^\ast=\{\phi_0,\phi_i\in\Real, \lambdavec\in\Real^\lambda| \
|\phi_0(t)-\phi_i(t)|\leq \varepsilon, \
\lambdavec\in\Omega_\lambda\subset\Real^\lambda\}
\]
is forward-invariant in the extended system
(\ref{eq:invariance_temporal_coding}), (\ref{eq:extension_visual})
and (\ref{eq:adjustment_visual}).  Third, we restrict our
attention to systems which have a subset $\Omega$  in their state
space such that trajectories starting in $\Omega$ converge to
$\Omega^\ast$. Finally, we guarantee that the state will
eventually visit domain $\Omega$ thus ensuring that
(\ref{eq:invariance_goals}) holds.

We have found that choosing extension (\ref{eq:extension_visual})
in the class of simple third-order bilinear systems
\begin{equation}\label{eq:extension_visual_simple}
\left\{\begin{split}
\dot{\lambda}_1&=\frac{\gamma_1}{\tau}\cdot (\phi_0-\phi_i)\\
\dot{\lambda}_2&=\gamma_2\cdot \lambda_3\cdot \|\phi_0-\phi_i\|_{\varepsilon}, \ \ \gamma_1,\gamma_2\in\Real_{>0}  \\
\dot{\lambda}_3&=-\gamma_2 \cdot \lambda_2 \cdot
\|\phi_0-\phi_i\|_{\varepsilon}, \ \
\sqrt{\lambda_2^2(t_0)+\lambda_3^2(t_0)}=1
\end{split}\right.
\end{equation}
ensures solution to Problem \ref{problem:invariance}. Specific
technical details and conditions are provided in Theorem
\ref{theorem:visual_invariance}
\begin{thm}\label{theorem:visual_invariance} Let system
(\ref{eq:invariance_temporal_coding}),
(\ref{eq:extension_visual_simple}) be given, and  function
$f_i(t,\theta_2)$ be separated from zero and bounded. In other
words, there exist constants $D_3, D_4 \in\Real_{>0}$ such that
for all $t\geq 0, \ \theta_2\in[\theta_{2,\min}, \theta_{2,\max}]$
the following condition holds:
\begin{equation}\label{eq:nonlinearity_bounds}
D_3\leq f_i(t,\theta_2) \leq D_4
\end{equation}

\noindent Then there exist positive $\gamma_1$, $\gamma_2$, and
$\varepsilon$ (see Table \ref{table:adjustment_parameters} for the
particular values):
\begin{equation}\label{eq:parameters_qualitative}
\gamma_2\ll\gamma_1, \ \
\varepsilon>\tau\Delta\left(1+\frac{D_4}{D_3}\right)
\end{equation}
such that adaptation mechanisms
\begin{equation}\label{eq:invariance_adjustment}
\left\{\begin{split} \hat{\theta}_{i,1}&=e_i \gamma_1 +
\lambda_1\\
\hat{\theta}_{i,2}(t)&=\theta_{2,\min}+\left(\lambda_2(t)+1\right)\frac{\theta_{2,\max}-\theta_{2,\min}}{2}
\end{split}\right.
\end{equation}
deliver a solution to  Problem \ref{problem:invariance}. In
particular, for all $\theta_1\in [\theta_{1,\min},
\theta_{1,\max}]$, $\theta_2\in [\theta_{2,\min},
\theta_{2,\max}]$ the following properties are guaranteed:
\[
\limsup_{t\rightarrow\infty}|\phi_0(t)-\phi_i(t)|\leq \varepsilon;
\  \ \exists \ \theta_2'\in[\theta_{2,\min},\theta_{2,\max}]: \
\lim_{t\rightarrow\infty}\hat{\theta}_{i,2}(t)=\theta_2',
\]
where the value of $\varepsilon$, depending on the choice of
parameters $\gamma_2$, $\gamma_1$, can be made arbitrarily close
to $\tau\Delta\left(1+{D_4}/{D_3}\right)$.
\end{thm}
Proof of the theorem is provided in \ref{appendix:3}.

\begin{table}[t]
  \centering
  \caption{Parameters of the compensatory mechanisms (\ref{eq:invariance_adjustment})}\label{table:adjustment_parameters}
\vskip 5mm {\small
  \begin{tabular*}{\textwidth}{@{\extracolsep{\fill}}|c|c|}
    \hline
     &  \\
     Parameter &  Values \\
     & \\
    \hline
     & \\
     $\gamma_1$ & \begin{minipage}[l]{0.7\linewidth}
\[
     \frac{\gamma_1}{\gamma_2}=q, \ q\in\Real_{>0}
\]
                   \end{minipage} \\
 &   \\
  \hline
   & \\
   $\varepsilon$ &  \begin{minipage}[l]{0.8\linewidth}
\[
  \varepsilon>\tau \left(\Delta\left(1+\frac{D_4}{D_3}\right)+\frac{\gamma_2}{\gamma_1}\left[\frac{\theta_{1,\max} D
D_2 D_4}{(D_3)^2}M_1 \tau
\left(1+\frac{D_4}{D_3}\right)\frac{\theta_{2,\max}-\theta_{2,\min}}{2}\right]
\right)
\]
\[
M_1=\Delta + k\theta_{1,\max} D D_2
|\theta_{2,\max}-\theta_{2,\min}|
\]
                    \end{minipage} \\
 &  \\
   \hline
 &  \\
 $\gamma_2$ & \begin{minipage}[l]{0.7\linewidth}
\[
\gamma_2<\left(\frac{1}{4 \tau}\right)^2\left[k \theta_{1,\max} D
D_2
\left(1+\frac{D_4}{D_3}\right)\left(\frac{\theta_{2,\max}-\theta_{2,\min}}{2}\right)\right]^{-1}
\]
             \end{minipage} \\
    &   \\
 \hline
   \end{tabular*}
}
\end{table}

Let us comment on the conclusions and conditions of Theorem
\ref{theorem:visual_invariance}. First of all, the theorem shows
that each $i$-th subsystem ensuring invariance of spatiotemporal
image representation to the given modelled perturbations can be
composed of no more than four differential equations:
\begin{subequations}
\begin{eqnarray}
\mathrm{Temporal \ integration:} \ & &
\dot{\phi}_i=-\frac{1}{\tau}\phi_i  + k\cdot \hat{\theta}_{i,1}
f_i(t,\hat{\theta}_{i,2})\label{eq:invariance_system_adaptive_integration} \\
\mathrm{Fast \ adaptation \ dynamics:} \ & &  \dot{\lambda}_1=\frac{\gamma_1}{\tau}\cdot (\phi_0-\phi_i)\label{eq:invariance_system_adaptive_fast}\\
\mathrm{Slow \ adaptation \ dynamics:} \ & & \left\{
\begin{array}{ll}
                                   \dot{\lambda}_2&=\gamma_2\cdot \lambda_3\cdot \|\phi_0-\phi_i\|_{\varepsilon},   \\
                                    \dot{\lambda}_3&=-\gamma_2 \cdot \lambda_2
                                    \cdot \|\phi_0-\phi_i\|_{\varepsilon}
                                  \end{array}
                                   \right. \label{eq:invariance_system_adaptive_slow}
\end{eqnarray}
\end{subequations}

Notice that the time scales of temporal integration
(\ref{eq:invariance_system_adaptive_integration}),  adaptation to
linearly parameterized uncertainties,
(\ref{eq:invariance_system_adaptive_fast}),  and adaptation to
nonlinearly parameterized uncertainties,
(\ref{eq:invariance_system_adaptive_slow}), are different. Because
of this difference in the time scales, subsystem
(\ref{eq:invariance_system_adaptive_fast}) is referred to as {\it
slow adaptation dynamics} and subsystem
(\ref{eq:invariance_system_adaptive_slow})  as {\it fast
adaptation dynamics}. The difference between the time scales
determines the degree of invariance and precision in the resulting
system. For instance, as follows from Table
\ref{table:adjustment_parameters}, ratio $\gamma_2/\gamma_1$
affects the value of $\varepsilon$. This value defines the
acceptable level of  mismatches between an image and a template.
In other words, it regulates the sensitivity of the system. The
smaller the ratio $\gamma_2/\gamma_1$, the higher the sensitivity.
Ratio $\gamma_2/ (1/\tau)$ (see proof for details) affects the
conditions for convergence.

Slow adaptation dynamics,
(\ref{eq:invariance_system_adaptive_slow}), can be interpreted as
a {\it searching, or wandering} dynamics in the interval
$[\theta_{2,\min},\theta_{2,\max}]$.  Its functional purpose is to
explore the interval $[\theta_{2,\min},\theta_{2,\max}]$ for
possible values  of $\hat{\theta}_{i,2}$ when models of
perturbation are inherently nonlinear and no other choice except
of explorative search is available. Solutions of the searching
dynamics in (\ref{eq:invariance_system_adaptive_slow}) are
harmonic signals with time-varying frequency $\gamma_2
\|\phi_0(t)-\phi_i(t)\|_{\varepsilon}$. The larger the error, the
higher the frequency of oscillation. When $\gamma_2
\|\phi_0(t)-\phi_i(t)\|_{\varepsilon}$ is constant, for instance
equals to unit, equations
(\ref{eq:invariance_system_adaptive_slow}) reduce to
\begin{equation}\label{eq:searching_prototype}
\begin{split}
\dot{\lambda}_2&= \lambda_3\\
\dot{\lambda}_3&=- \lambda_2
\end{split}
\end{equation}
In general, every subsystem
\begin{equation}\label{eq:searching_general}
\begin{split}
\dot{\lambda}_2&= g_2(\lambda_2,\lambda_3,t),  \\
\dot{\lambda}_3&= g_3(\lambda_2,\lambda_3,t), \ \ \ g_2,g_3\in C^0
\end{split}
\end{equation}
generating dense trajectories $\lambda_2(t)$ in
$[\theta_{2,\min},\theta_{2,\max}]$ for some initial conditions
$\lambda_2(t_0)$, $\lambda_3(t_0)$ and, at the same time, ensuring
boundedness of $\lambda_2(t)$, $\lambda_3(t)$ for all
$t\in\Real_{\geq 0}$ could be a replacement for
(\ref{eq:searching_prototype}) in
(\ref{eq:invariance_system_adaptive_slow}) (see also
\cite{ArXive:Non-uniform:2006}). Conclusions of the theorem in
this case will remain the same except, probably, with respect to
the choice of the particular values of $\gamma_1$, $\gamma_2$,
$\varepsilon$ in Table \ref{table:adjustment_parameters}. Our
present choice of subsystem (\ref{eq:searching_prototype}) in
(\ref{eq:invariance_system_adaptive_slow}) as a prototype for the
searching trajectory was motivated primarily by its simplicity in
realization and linearity in state.

The fast adaptation dynamics,
(\ref{eq:invariance_system_adaptive_fast}), corresponds to
exponentially stable mechanisms. This can easily be verified by
differentiating the difference $\hat{\theta}_{i,1}(t)-\theta_1$
with respect to time (see also
(\ref{eq:theorem:invariance:error_dynamics_3}) in
\ref{appendix:3}). The function of the fast adaptation subsystem
is to track instantaneous changes in $\theta_1$ exponentially fast
in such a way that the difference $\hat{\theta}_{i,1}(t)-\theta_1$
is determined mostly by mismatches
$\hat{\theta}_{i,2}(t)-\theta_2$.

\begin{figure}
\begin{center}
\includegraphics[height=150pt]{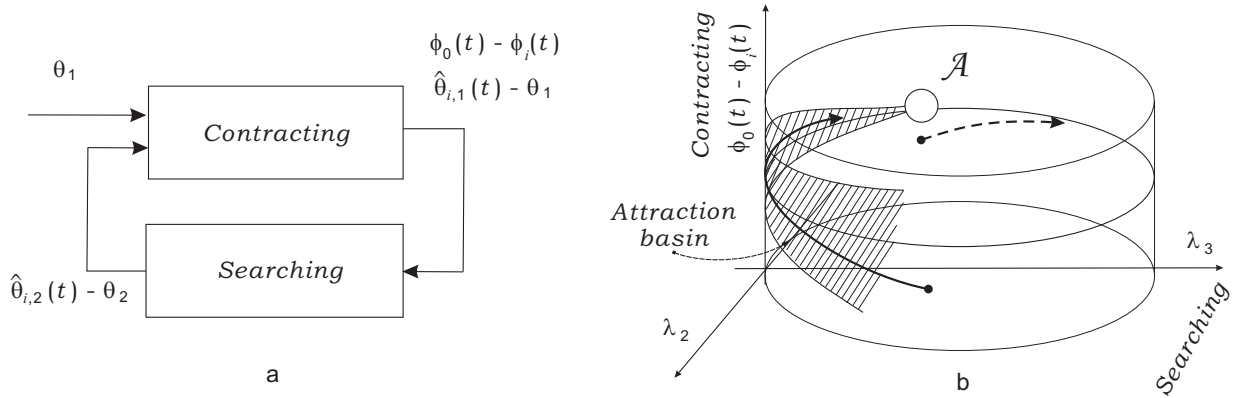}
\caption{The interplay between  temporal integration,
(\ref{eq:invariance_system_adaptive_integration}), and fast and
slow adaptation (\ref{eq:invariance_system_adaptive_fast}),
(\ref{eq:invariance_system_adaptive_slow}) in the proposed
solution to the problem of invariant template matching. {\bf Panel
a.} Contracting dynamics corresponds  to the processes of temporal
integration of a template and adaptation to linearly parameterized
uncertainties. Searching dynamics is due to the adaptation to
nonlinearly parameterized uncertainties. {\bf Panel b.} Diagram of
the phase portrait of system
(\ref{eq:invariance_system_adaptive_integration}),
(\ref{eq:invariance_system_adaptive_fast}),
(\ref{eq:invariance_system_adaptive_slow}). Interaction between
searching and contracting subsystems forms a weakly attracting
invariant set $\mathcal{A}$. Its basin of attraction is not
necessarily a neighborhood of $\mathcal{A}$. This means that some
trajectories starting in a small vicinity of $\mathcal{A}$ might
eventually leave its neighborhood (dashed trajectory), while
trajectories starting far away from $\mathcal{A}$ might enter such
neighborhoods and remain there (solid
line).}\label{fig:contracting_wandering}
\end{center}
\end{figure}

The problem of template matching is solved through the interplay
of searching dynamics $\hat{\theta}_{i,2}(t)-\theta_2$ (see Figure
\ref{fig:contracting_wandering}) and the contracting dynamics
expressed by $\phi_0(t)-\phi_i(t)$.
We use the results from \cite{ArXive:Non-uniform:2006} to prove
the emergence of weakly (Milnor) attracting sets in the system
state space.

In principle, linearity of the uncertainty models in $\theta_1$ is
not necessary to guarantee exponential stability of
$\hat{\theta}_{i,1}(t)-\theta_1$. As has been shown in
\cite{IEEE_TAC_2007}, exponential stability of
$\hat{\theta}_{i,1}(t)-\theta_1$ can be  ensured by the same
function $\hat{\theta}_{i,1}(t)$ as in
(\ref{eq:adjustment_visual}) if we replace $\theta_1
f_i(t,\theta_2)$ with
$\tilde{f}_i(t,\theta_1,\theta_2):\Real_{\geq
0}\times\Real\times\Real\rightarrow\Real$. Nonlinearities
$\tilde{f}_i(t,\theta_1,\theta_2)$, however, should be monotone in
$\theta_1$. In this case condition (\ref{eq:nonlinearity_bounds})
is to be replaced with the following
\begin{equation}\label{eq:nonlinearity_bounds_monotone}
D_3 \leq
\frac{\tilde{f}_i(t,\hat{\theta}_{i,1},\theta_2)-\tilde{f}_i(t,\theta_1,\theta_2)}{\hat{\theta}_{i,1}-\theta_1
}\leq  D_4, \ \forall \
\theta_2\in[\theta_{2,\min},\theta_{2,\max}]
\end{equation}
The general line of the proof will remain unaffected by this
extension.

The proposed compensatory mechanisms
(\ref{eq:invariance_temporal_coding}),
(\ref{eq:extension_visual_simple})
(\ref{eq:invariance_adjustment}) are nearly optimal in terms of
the dimension of the state of the whole system. Indeed, in order
to track uncertain and independent $\theta_1$, $\theta_2$ two
extra variables are to be introduced. This implies that the
minimal dimension of state of a system which solves Problem
\ref{problem:invariance} equals three. Our four-dimensional system
is therefore close to the optimal configuration. Furthermore, as
follows from the proof of the theorem, the dimension of the slow
subsystem could be reduced to one. Thus, in principle, a minimal
realization could be achieved. In this case, however, boundedness
of the state for every initial condition is no longer guaranteed.

Theorem \ref{theorem:visual_invariance} establishes conditions for
convergence of the trajectories of our prototype system
(\ref{eq:invariance_temporal_coding}),
(\ref{eq:extension_visual_simple})
(\ref{eq:invariance_adjustment}) to an invariant set in the system
state space. In particular, when matching condition
(\ref{eq:Delta}) is met, it assures that temporal representation
$\phi_i(t)$ of the template tracks temporal representation
$\phi_0(t)$ of the image. In the next subsection we discuss how
the similarity between these temporal representations can be
detected by a system of coupled spiking oscillators.

\subsection{Conditions for synchronization of coincidence
detectors}\label{chap:5:imag:synch}

Consider coincidence detectors (\ref{HR_model_net}),
(\ref{eq:chap:5:image:coupling}), (\ref{eq:HR_model_param})
modeled by a system of coupled Hindmarsh-Rose oscillators.  The
goal of this section is to provide a constructive solution to
Problem \ref{problem:detection}. First, we seek for conditions
ensuring global exponential stability of the synchronization
manifold of $\phi_0(t)=\phi_i(t)$ when $\phi_i(t)$ are identical
for each $i$. We do this by showing that solutions of the system
are globally bounded, and for each pair of indexes
$i,j\in\{0,\dots,n\}$ there exists a differentiable positive
definite function $V(x_i,y_i,z_i,x_j,y_j,z_j)$, ${\pd V}/{\pd
x_i}=-{\pd V}/{x_j}$ such that $V$ grows towards infinity with
distance from the synchronization manifold and for all bounded
continuous $\phi_i(t)=\phi_j(t)$ the following holds:
\begin{equation}\label{eq:HR_synch_pair}
\dot{V}\leq - \alpha V, \ \alpha\in\Real_{>0}.
\end{equation}
When $\phi_i(t)\ne \phi_j(t)$ equation (\ref{eq:HR_synch_pair})
implies that
\begin{equation}\label{eq:HR_synch_pair_2}
\dot{V}\leq - \alpha V + \frac{\pd V}{x_i}(\phi_i(t)-\phi_j(t)).
\end{equation}
Then using (\ref{eq:HR_synch_pair_2}) and comparison lemma
\cite{Khalil:2002} we show that convergence of $\phi_i(t)$ to
$\phi_j(t)$ at $t\rightarrow\infty$ implies convergence of
variables $x_i(t)$, $y_i(t)$, $z_i(t)$, $x_j(t)$, $y_j(t)$,
$z_j(t)$ to the synchronization manifold.
%
The formal statement of this result is provided in Theorem
\ref{theorem:coincidence_detectors}

\begin{thm}\label{theorem:coincidence_detectors} Let system
(\ref{HR_model_net}) be given, function  $\bfu$ be defined as in
 (\ref{eq:chap:5:image:coupling}) and functions $\phi_i(t)$, $i\in\{0,\dots,n\}$ be bounded. Then

1) solutions of the system are bounded for all $\gamma\in\Real_+$;

2) if, in addition, the following condition is satisfied
\begin{equation}\label{equ:global-upper-bound}
\gamma>\frac{1}{(n+1)\cdot a}\left(\frac{d^2}{2}+b^2\right),
\end{equation}
then for all $i,j\in\{0,\dots,n\}$ condition
\[
\limsup_{t\rightarrow\infty}|\phi_i(t)-\phi_j(t)|\leq \varepsilon
\]
implies that
\begin{eqnarray}\label{eq:chap:5:image:synch}
\begin{split}
\limsup_{t\rightarrow\infty}|x_i(t)-x_j(t)|&\leq
\delta(\varepsilon), \\
\limsup_{t\rightarrow\infty}|y_i(t)-y_j(t)|&\leq
\delta(\varepsilon), \\
\limsup_{t\rightarrow\infty}|z_i(t)-z_j(t)|&\leq
\delta(\varepsilon).
\end{split}
\end{eqnarray}
where $\delta:\Real_+\rightarrow\Real_+$ is a monotone and
vanishing at zero function.
\end{thm}

Theorem \ref{theorem:coincidence_detectors} specifies the
boundaries for stable synchrony in the system of coupled neural
oscillators (\ref{HR_model_net}) as a function of the coupling
strength, $\gamma$, and parameters $a$, $b$, and $d$ of a single
oscillator. The last three parameters represent properties of the
membrane and combined with $x_0$, $\varepsilon$, $s$ and $I$
completely characterize the dynamics of a single model neuron
\cite{ProcRSL:Hindmarsh_and_Rose:1984}, ranging from single
spiking to periodic or chaotic bursts.

The distinctive feature of Theorem
\ref{theorem:coincidence_detectors} is that it is suitable for
analysis of systems with external time-dependent perturbations
$\phi_i(t)$. This property is essential for the comparison task,
where the oscillators are fed with time-varying inputs and the
degree of their mutual synchrony is the measure of similarity
between the inputs.

While the theorem provides us with conditions for stable
synchrony, it allows us to estimate the domain of values of the
coupling parameter $\gamma$ corresponding to potential
intermittent,
itinerant\cite{Kaneko_and_Tsuda:2000,Chaos:Kaneko:2003} or
meta-stable regimes. In particular, as follows from Theorem
\ref{theorem:coincidence_detectors}, a necessary condition for
unstable synchronization in system (\ref{HR_model_net}) is
\begin{equation}\label{equ:necessary_unstable}
\gamma<\frac{1}{(n+1)\cdot a}\left(\frac{d^2}{2}+b^2\right).
\end{equation}
Notice that conditions (\ref{equ:necessary_unstable}),
(\ref{equ:global-upper-bound}) do not depend on the
``bifurcation'' parameter $I$ which usually determines the type of
bursting in the single oscillator. They also do not depend on the
differences in the time scales defined by parameter $\varepsilon$
between the fast $x$, $y$, and slow, $z$, variables. Hence these
conditions apply in a wide range of system behavior on the
synchronization manifold. This advantage also has  a downside,
because conditions (\ref{equ:necessary_unstable}),
(\ref{equ:global-upper-bound}) are too conservative. This,
however, seems to be a reasonable price for invariance of criteria
(\ref{equ:necessary_unstable}), (\ref{equ:global-upper-bound})
with respect to the full range of dynamical behavior of a
generally nonlinear system.

\section{Discussion}\label{sec:discussion}

We provided a principled solution to the issue of invariance in
the problem of template matching. The pattern recognition problem
can be solved using a network of nonlinear oscillators which
synchronize when mismatches in the temporal representations of
image and templates converge to zero. Although the solution to the
latter problem is not normative we tried to keep the number of
relevant parameters at minimum. In particular the dimension of the
state of a single adaptation compartment is three which is minimal
for generation of spikes ranging from periodic to  chaotic bursts.
Moreover, conditions (\ref{equ:necessary_unstable}),
(\ref{equ:global-upper-bound}) allow us to choose  coupling
strength $\gamma$ as single control parameter for regulating
stability/instability of the synchronous activity in the network.

In this section we provide further extensions of the basic results
of Theorems \ref{theorem:visual_invariance},
\ref{theorem:coincidence_detectors}, discuss possible links
between the normative part of our theory and known results in
vision, and provide simple illustrations how  particular systems
for invariant template matching can be constructed using these
results.

\subsection{Extension to the frequency-encoding schemes}

\begin{figure}[t!]
\centering
\includegraphics[height=230pt]{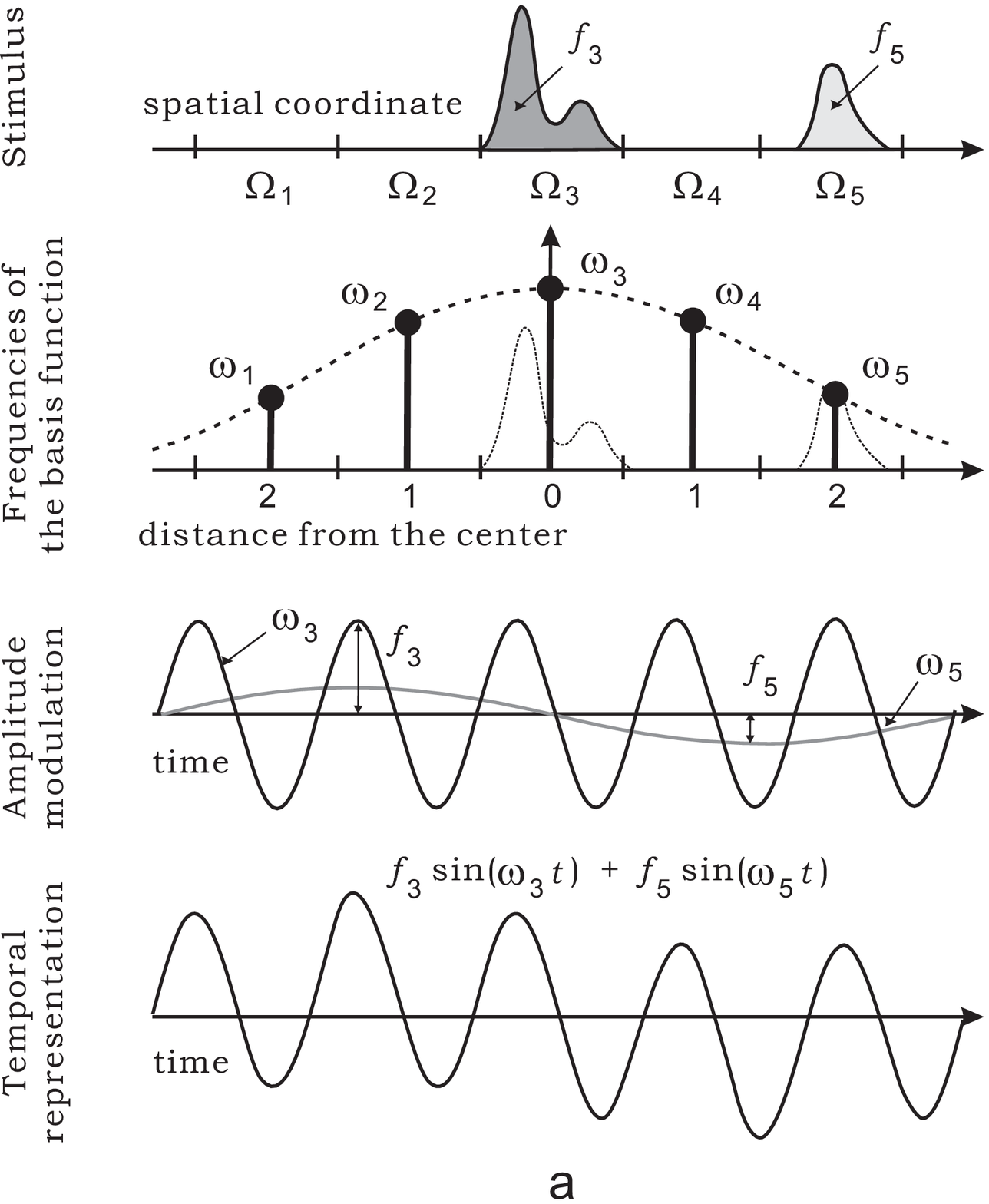}
\hspace{35pt}
\includegraphics[height=230pt]{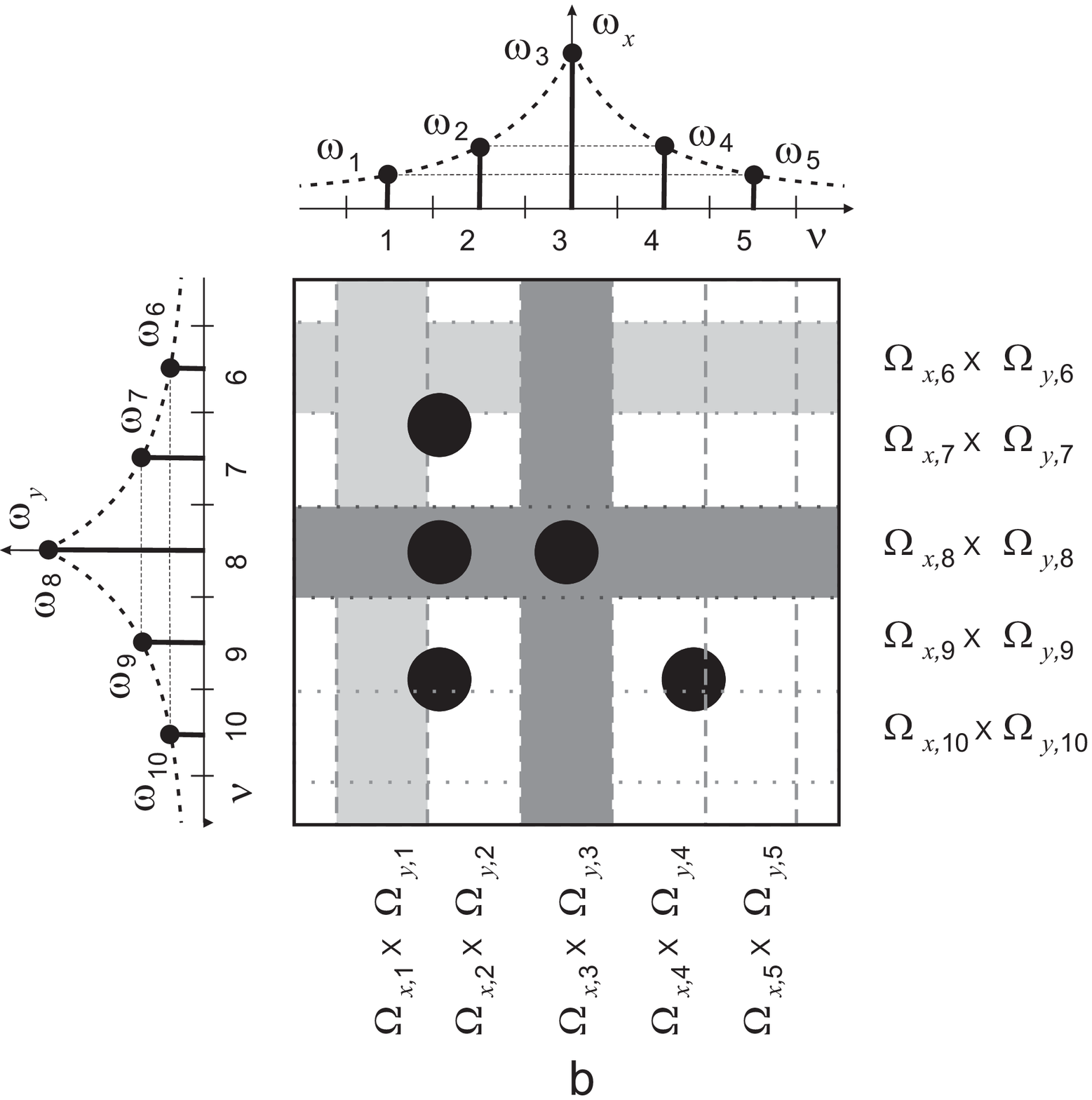}
\caption{{\bf Panel a.} Temporal representation of a spatially
distributed stimulus using frequency encoding. A stimulus (upper
row) $S(x)$ is spatially sampled by partitioning its domain into
the union of intervals $\Omega_i$. For each $\Omega_i$ an integral
$f_i=f(\mathcal{F}_i)=\int_{\Omega_i} S(x)d x$ is calculated and a
frequency $\omega_i$ is assigned. The resulting temporal
representation (lower row) is expressed as the sum of two
amplitude-modulated harmonic signals of frequencies, $\omega_3$,
$\omega_5$. {\bf Panel b.} Temporal representation of a
two-dimensional pattern. The pattern consists of black filled
circles. The image domain is partitioned into a collection of
horizontal and vertical strips. Dark domains correspond to higher
frequencies.}\label{fig:frequency_encoding}
\end{figure}

For the sake of notational simplicity we restricted our attention
to temporal representations (\ref{eq:sampling_def}),
(\ref{eq:sampling_functional_final_def}) of spatially sampled
images. These encoding schemes can  be interpreted as scanning of
an image over time. Yet, the results of Section
\ref{sec:main_results} apply to a broader class of encoding
schemes. One example is frequency-coding used in many neural
systems. Let us consider factorization (\ref{eq:sampling_def})
where in the notation $\mathcal{F}_t [S_0,\thetavec] (x,y)$ symbol
$t$ is replaced with $\nu$. In order to extend the initial
encoding scheme to the domain of frequency/spike rate encoding we
introduce an additional linear functional $f_{\omega}$ as follows:
\begin{equation}\label{eq:sampling_functional_freq_def}
f_{\omega}(t,\mathcal{F}_{\nu} [S_0,\thetavec])=\sum_{\nu}
h(\omega_{\nu}\cdot t)\cdot f(\mathcal{F}_{\nu} [S_0,\thetavec]),
\end{equation}
where $h:\Real\rightarrow\Real$ is a bounded periodic function,
and $\omega_\nu$ are distinct real numbers indexed by $\nu$.
Function $h(\omega_\nu \cdot t)$ in
(\ref{eq:sampling_functional_freq_def}) serves as a basis or
carrier function generating periodic impulses of various
frequencies $\omega_\nu$. Thus each $\nu$-th spatial sample of the
image is assigned a particular frequency, and the amplitude of the
oscillation is specified by $f(\mathcal{F}_{\nu}
[S_0,\thetavec])$. Temporal representation of a one-dimensional
stimulus according to encoding scheme
(\ref{eq:sampling_functional_freq_def}) is illustrated in Figure
\ref{fig:frequency_encoding}, panel a.

This encoding scheme is plausible to biological vision, when
frequencies $\omega_\nu$ are ordered according to relative
position of domains $\Omega_{x,\nu}$, $\Omega_{y,\nu}$ to the
center of the image. This corresponds, in particular, to the
receptive fields in cat retinal ganglion cells
\cite{JPhys:Cugell:1983}. Because the functional $f_{\omega}$ is
linear in $f(\mathcal{F}_{\nu} [S_0,\thetavec])$ and function
$h(\omega_{\nu}\cdot t)$ is bounded for all $t$, condition
(\ref{eq:sampling_functional}) will be satisfied for $f_{\omega}$.
Hence the conclusions of Theorem \ref{theorem:visual_invariance}
apply to these systems.

\subsection{Multiple representations of uncertainties}

Another property of system (\ref{eq:invariance_temporal_coding}),
(\ref{eq:extension_visual_simple}),
(\ref{eq:invariance_adjustment}), in addition to its ability to
accommodate relevant encoding schemes such as frequency/rate and
sequential/random scanning, is that each single value of
$\theta_2\in(\theta_{2,\min},\theta_{2,\max})$ induces at least
two distinct attracting sets in the extended space. Indeed
\[
\lambda_2^2(t)+\lambda_3^2(t)=\const=1
\]
along the trajectories of (\ref{eq:invariance_temporal_coding}),
(\ref{eq:extension_visual_simple}),
(\ref{eq:invariance_adjustment}) (see also the proof of Theorem
\ref{theorem:visual_invariance}). Hence for almost every value of
$\lambda_2$ (except when $\lambda_2=\pm 1$) in the definition of
$\hat{\theta}_2(t)$ in (\ref{eq:invariance_adjustment}) there will
always be two distinct values of $\lambda_3$:
\[
\lambda_{3,1}= \sqrt{1-\lambda_2^2}, \ \lambda_{3,2}=
-\sqrt{1-\lambda_2^2}.
\]
These give rise to distinct invariant sets in the system state
space for the single value of $\theta_2$. The presence of two
complementary encodings for the same figure is a plausible
assumption that has been used in the perceptual organization
literature to explain a range of phenomena, including perceptual
ambiguity, modal and amodal completion, etc. See
\cite{Psych_Res:vanLeeuwen:1990}, \cite{Psych_Bull:Hatfield:1985}
for a review. A consequence of the presence of multiple attractors
corresponding to the single value of perturbation is that the time
for convergence (the decision time) may change abruptly with small
variations of initial conditions. The latter property is well
documented in human subjects \cite{PsychRev:Gilden:2001}.
Furthermore, the presence of two attractors with different basins
for a single value of perturbation will lead to asymmetric
distributions of decision times, which is typically observed in
human and animal reaction time data \cite{TrNeuroSc:Smith:2004}.

\subsection{Multiple time scales for different modalities in vision}

An important property of the proposed solution to the problem of
invariance is that the time scales of adaptation to linearly and
nonlinearly parameterized uncertainties are substantially
different. This difference in time scales emerged naturally in the
course of our mathematical argument as a consequence of splitting
the system dynamics into a slow searching subsystem and a fast
asymptotically stable one. This allowed us to prove emergence of
unstable yet attracting invariant sets thus ensuring existence of
a solution to the problem of invariant template
matching.
In particular, Theorem \ref{theorem:visual_invariance} requires
that the time constants of adaptation to linearly parameterized
uncertainties (for instance, unknown intensity of the image) are
to be substantially smaller than the time constants of adaptation
to nonlinearly parameterized uncertainties (image blur, rotation,
scaling etc.). Furthermore, as follows from Table
\ref{table:adjustment_parameters}, the larger the difference in
the time scales the higher the possible precision and the smaller
the errors.

Even though the difference in time scales was motivated purely by
theoretical considerations, there is strong evidence that
biological systems adapt at different time scales to uncertainties
from different modalities. For example, the time scale of light
adaptation is within tens of milliseconds
\cite{VisionResearch:Wolfson:2002} while adaptation to
``higher-order'' modalities  like rotation and image blur extends
from hundreds of milliseconds to minutes
\cite{Nature:Neuroscience:Webster:2002}. In motor learning the
evidence of presence of slow and fast adaptation at the time scale
minutes is reported in \cite{PLOS:Biology:Smith:2006}. These
findings, therefore, motivate our belief that system
(\ref{eq:invariance_temporal_coding}),
(\ref{eq:extension_visual_simple}),
(\ref{eq:invariance_adjustment}) could serve as a simple, yet
qualitatively realistic, model for adaptation mechanisms in
vision, motor behavior, and decision making.

\subsection{Rotation-invariant matching and mental rotation experiments}

Let us illustrate how the results of Theorems
\ref{theorem:visual_invariance},
\ref{theorem:coincidence_detectors} can be applied to template
matching when an object is rotated over an unknown angle and its
brightness is uncertain a-priori. In order to obtain a temporal
representation of the image we use the frequency-encoding scheme
(\ref{eq:sampling_functional_freq_def}) as is illustrated in
Figure \ref{fig:frequency_encoding}, panel b. In particular we use
the following transformation
\begin{equation}\label{eq:sampling_functional_freq_rot}
\theta_1 f_i(t,\theta_2)=\theta_1 \sum_{\nu} h(\omega_{\nu}\cdot
t)\cdot f(\bar{\mathcal{F}}_{\nu} [S_i,\theta_2]),
\end{equation}
where $\theta_2$ is the rotation angle of image $S_i(x,y)$ around
its central point, $\theta_1$ is the image brightness, function
$h(\omega_{\nu}\cdot t)=\sin^2(\omega_{\nu}\cdot t)$, and
\[
f(\bar{\mathcal{F}}_{\nu}
[S_i,\theta_2])=\int_{\Omega_{x,\nu}\times\Omega_{y,\nu}}
\bar{\mathcal{F}}_{\nu} [S_i,\theta_2](\xi,\gamma) d\xi d\gamma.
\]
is simply an integral of the rotated image $S_i$ by the angle
$\theta_2$ over the strip $\Omega_{x,\nu}\times\Omega_{y,\nu}$.

According to (\ref{eq:invariance_temporal_coding}),
(\ref{eq:extension_visual_simple}),
(\ref{eq:invariance_adjustment}), (\ref{HR_model_net}) the
recognition system (see Figure \ref{fig:chap:5:visual_system} for
its general structure) can be described by the system of
differential equations provided in Table
\ref{table:rotation-invariant}. Details of their implementation,
specific values of the parameters, initial conditions, and the
source files of a working MATLAB Simulink model  can be found in
\cite{url:template_matching}.
\begin{table}
\small
\begin{tabular*}{\textwidth}{@{\extracolsep{\fill}}|c|c|c|}
\hline
    Function &  Image & Template\\
    \hline
 \begin{minipage}[c]{0.2\linewidth} \begin{center} Temporal\\ integration \end{center}  \end{minipage} & \begin{minipage}[c]{0.33\linewidth}
                            \[
                            \dot{\phi}_0=-\frac{1}{\tau}\phi_0 + k\cdot \theta_1 f(t,\theta_2)
                            \]
                            \vspace{1pt}
                            \end{minipage} &
                        \begin{minipage}[c]{0.33\linewidth}
                        \[
                          \dot{\phi}_i=-\frac{1}{\tau}\phi_i + k\cdot \hat{\theta}_{1,i} f(t,\hat{\theta}_{2,i})
                         \]
                          \vspace{1pt}
                         \end{minipage}\\
 \hline
  \begin{minipage}[c]{0.2\linewidth} \begin{center} Adaptation\\ to brightness \end{center} \end{minipage} &   No & \begin{minipage}[c]{0.33\linewidth}
                        \begin{eqnarray}
                        \hat{\theta}_{1,i}&=&(\phi_0-\phi_i)\gamma_1+\lambda_{i,1}\nonumber\\
                        \dot{\lambda}_{i,1}&=&\frac{\gamma_1}{\tau}(\phi_0-\phi_i)\nonumber
                        \end{eqnarray}
                        \vspace{1pt}
                      \end{minipage}\\
  \hline
  \begin{minipage}[c]{0.2\linewidth} \begin{center} Adaptation\\ to rotation \end{center} \end{minipage} &   No & \begin{minipage}[c]{0.33\linewidth}
                        \begin{eqnarray}
                        \hat{\theta}_{2,i}&=&(\lambda_{2,i}(t)+1)\pi\nonumber\\
                        \dot{\lambda}_{i,2}&=&\gamma_2 \lambda_{i,3} \|\phi_0-\phi_i\|_{\varepsilon}\nonumber\\
                        \dot{\lambda}_{i,3}&=&-\gamma_2 \lambda_{i,2}
                        \|\phi_0-\phi_i\|_{\varepsilon}\nonumber
                        \end{eqnarray}
                        \vspace{1pt}
                      \end{minipage}\\
  \hline
 \begin{minipage}[c]{0.2\linewidth} \begin{center} Detectors \\ of similarity \end{center} \end{minipage} & \begin{minipage}[c]{0.33\linewidth}
                         \[
                        \begin{array} {ll}
                         \dot{x}_0 &= -a{x}_0^3+b{x}_0^2+y_0-z_0 +I\\
                                   & +u_0 + \phi_0(t),\\
                         \dot{y}_0 &= c-d{x}_0^2-y_0,\\
                         \dot{z}_0 &= \eps(s(x_0+x_0)-z_0),
                         \end{array}
                         \] \vspace{1pt}
                        \end{minipage} & \begin{minipage}[c]{0.33\linewidth}
                                          \[
                                            \begin{array} {ll}
                                             \dot{x}_i &= -a{x}_i^3+b{x}_i^2+y_i-z_i+I\\
                                                       &+u_i + \phi_i(t),\\
                                             \dot{y}_i &= c-d{x}_i^2-y_i,\\
                                             \dot{z}_i &= \eps(s(x_i+x_i)-z_i),
                                             \end{array}
                                             \] \vspace{1pt}
                                           \end{minipage}\\
\hline
 \begin{minipage}[c]{0.2\linewidth}
 \vspace{10pt}
 \begin{center} Coupling \\ function
 \end{center} \vspace{1pt}
 \end{minipage} & $u_0=\gamma\left(-(n+1) x_0 + \sum_{j\neq 0} x_j\right)$ &  $u_i=\gamma\left(-(n+1) x_i + \sum_{j\neq i}
 x_j\right)$\\
\hline
\end{tabular*}
\caption{Equations of the system for rotation and
brightness-invariant template
matching.}\label{table:rotation-invariant}
\end{table}

\begin{figure}[t!]
\begin{center}
\begin{minipage}[c]{0.32\linewidth}
\centering

\includegraphics[width=0.5\textwidth]{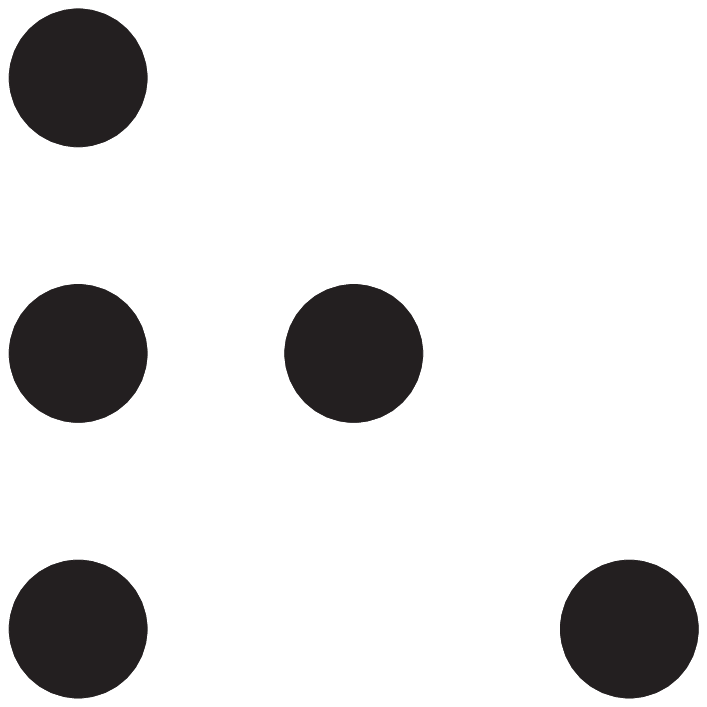}

\vspace{1mm}

\hspace{10pt}\includegraphics[width=0.9\textwidth]{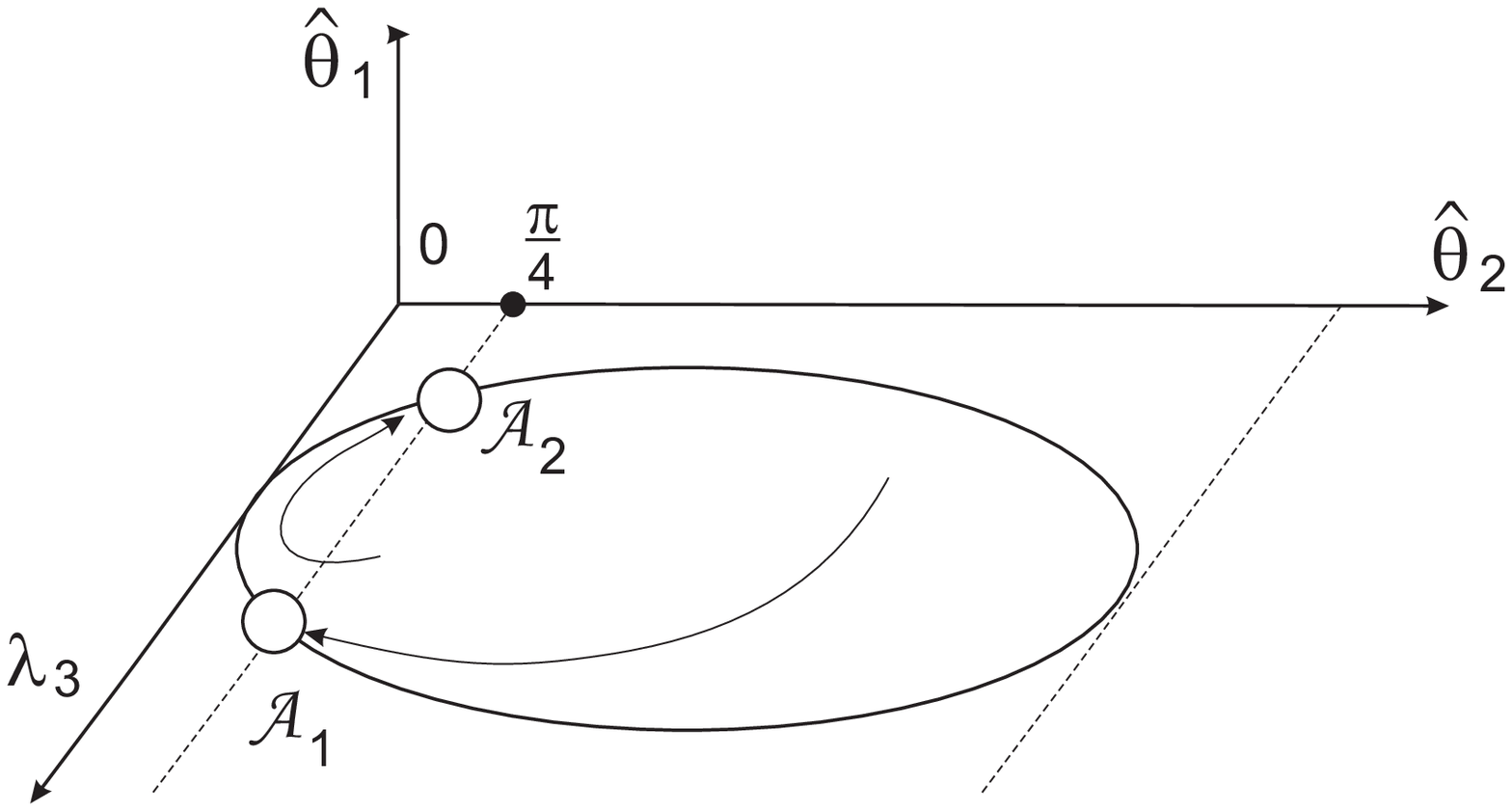}

\vspace{5mm}

\includegraphics[width=0.8\textwidth]{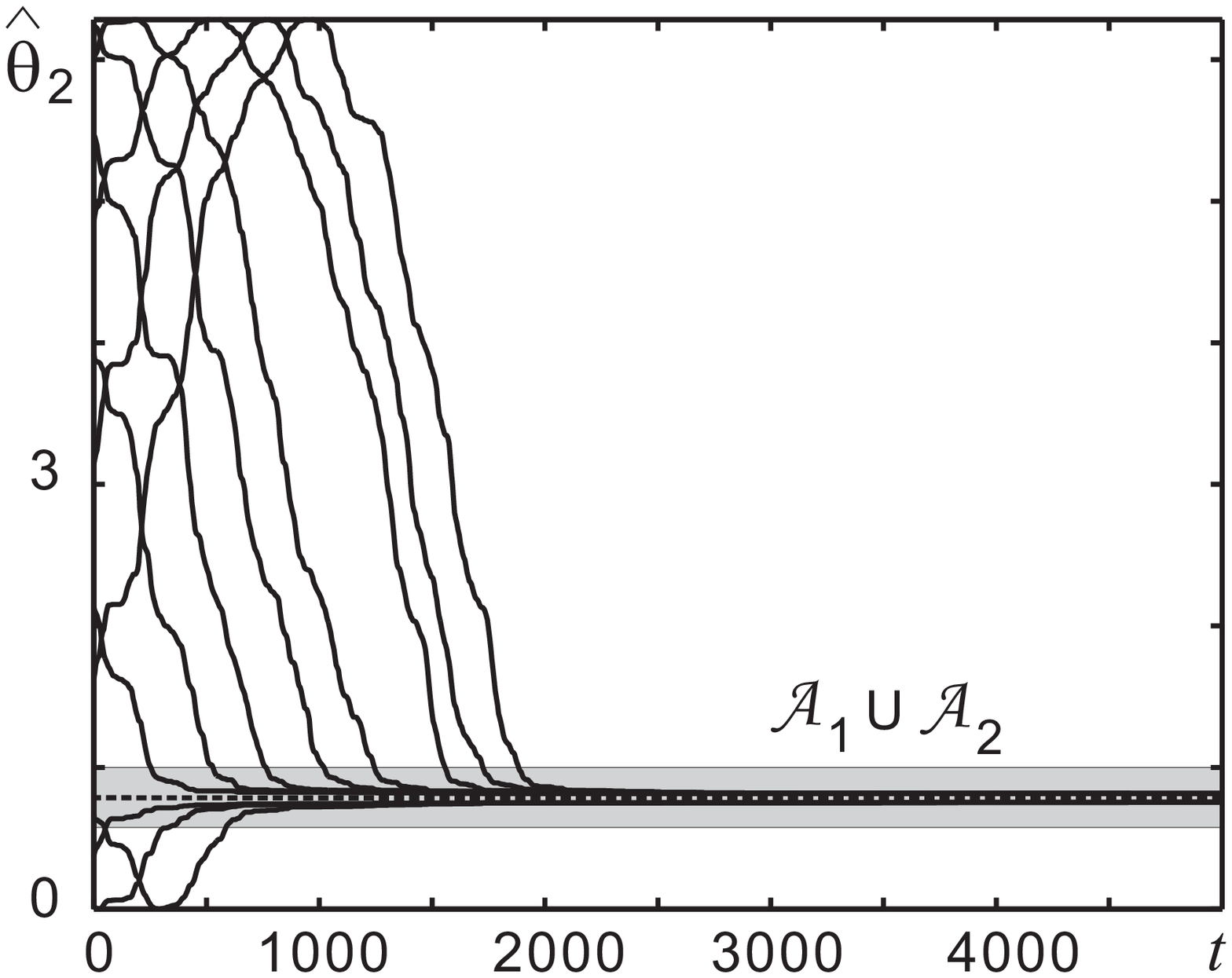}
\end{minipage}
\begin{minipage}[c]{0.32\linewidth}
\centering

\includegraphics[width=0.5\textwidth]{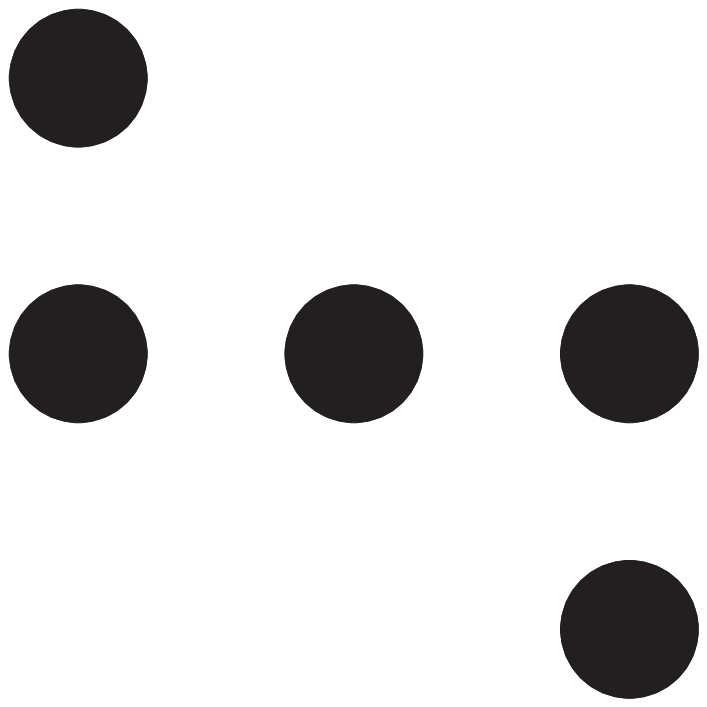}

\vspace{1mm}

\hspace{10pt}\includegraphics[width=0.9\textwidth]{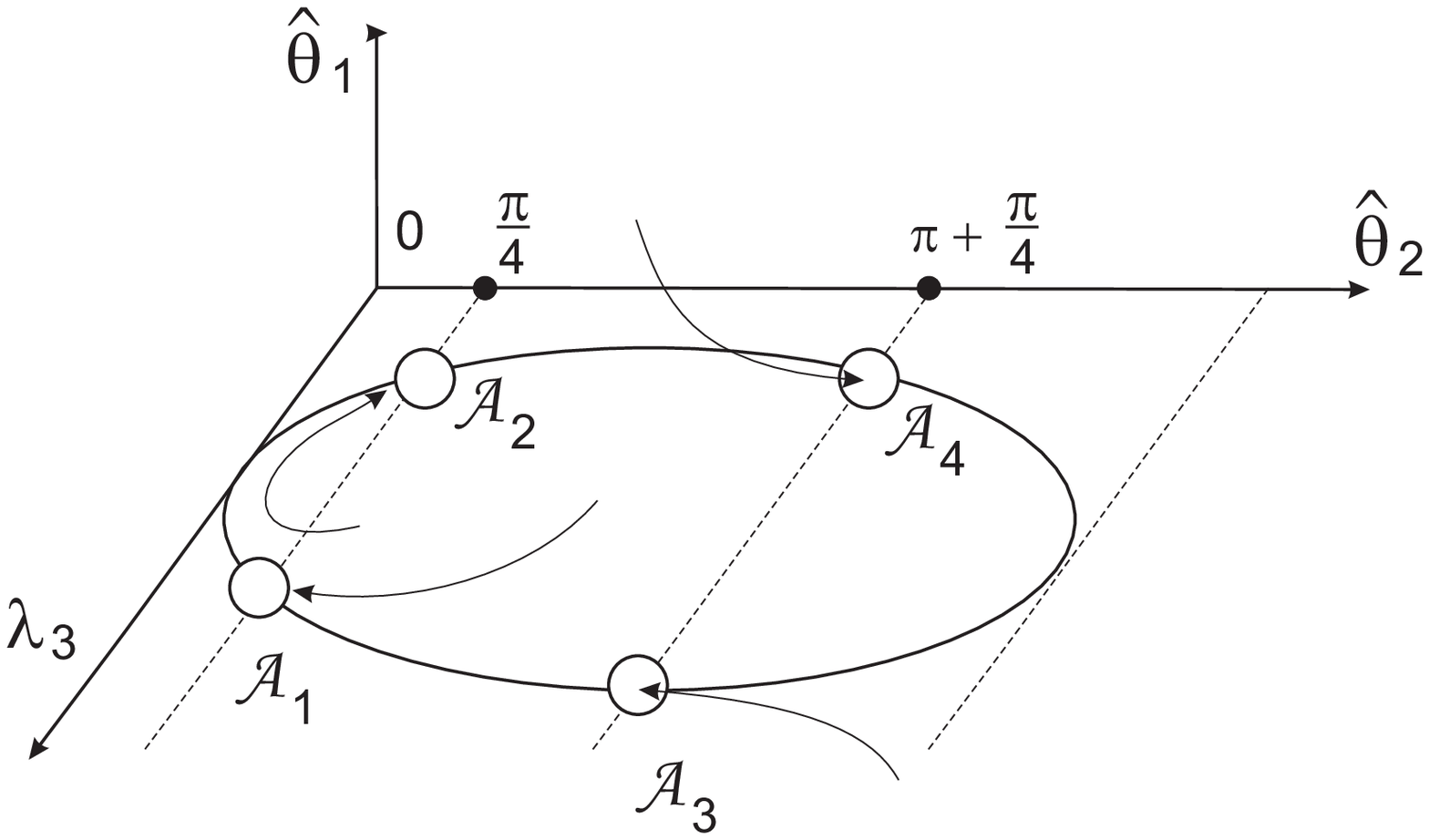}

\vspace{5mm}

\includegraphics[width=0.8\textwidth]{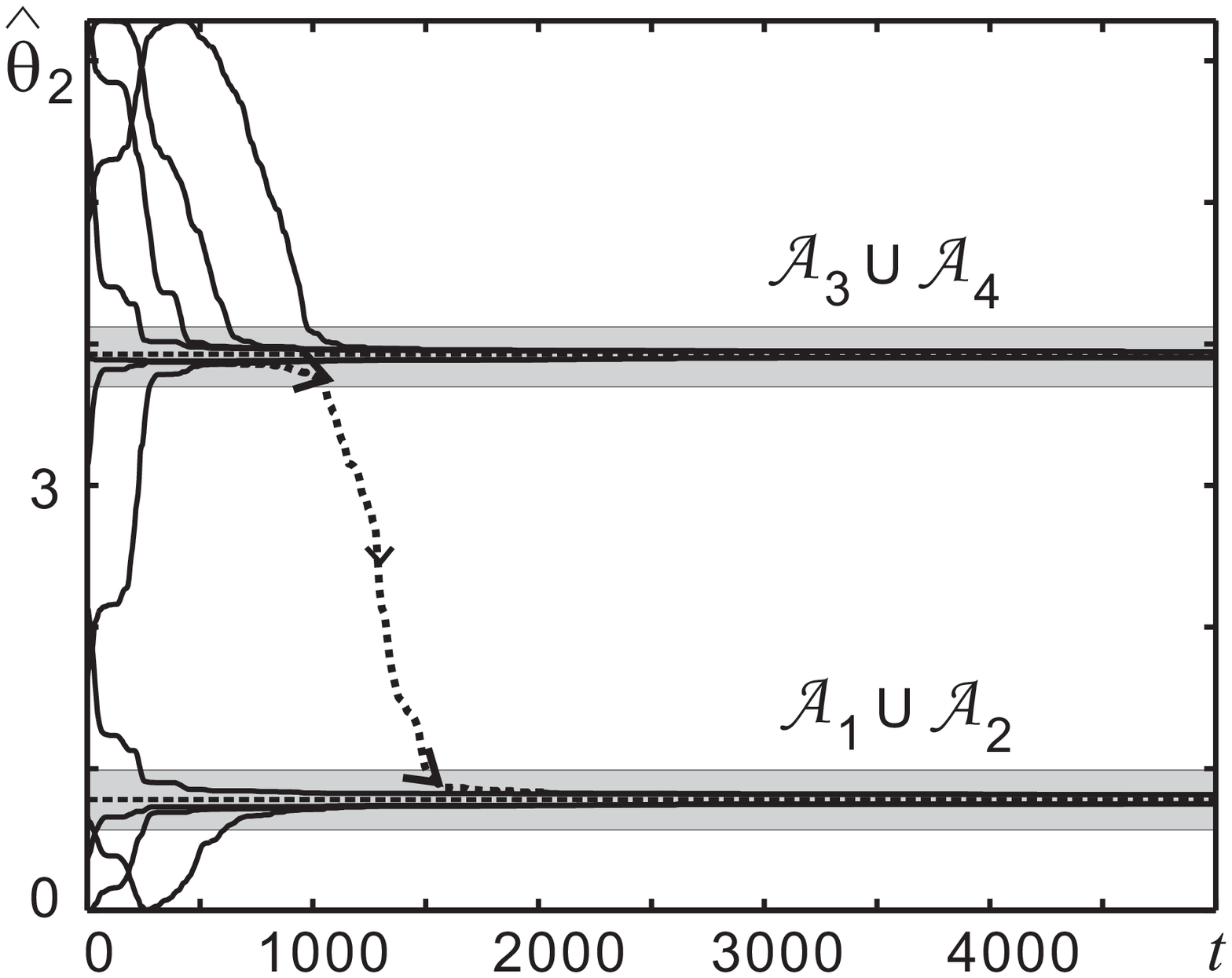}
\end{minipage}
\begin{minipage}[c]{0.32\linewidth}
\centering

\includegraphics[width=0.5\textwidth]{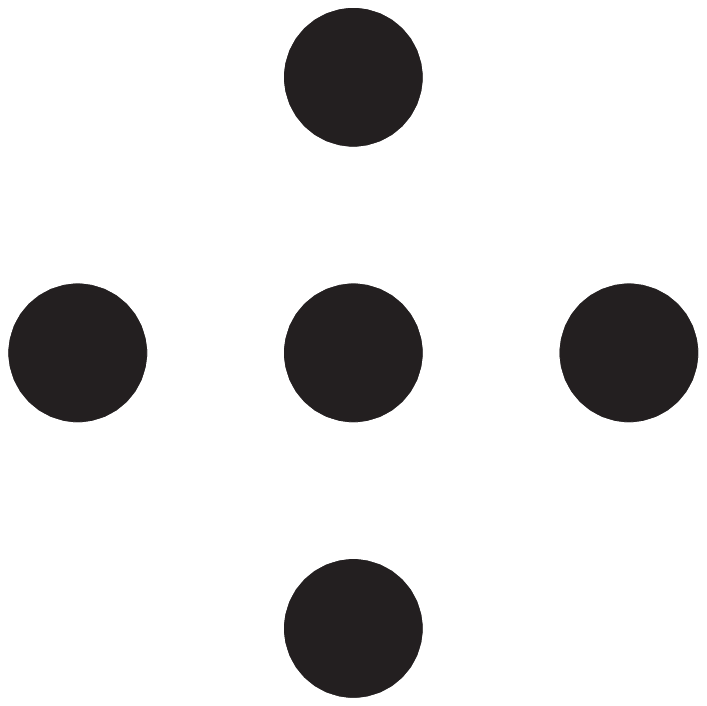}

\vspace{1mm}

\hspace{10pt}\includegraphics[width=0.9\textwidth]{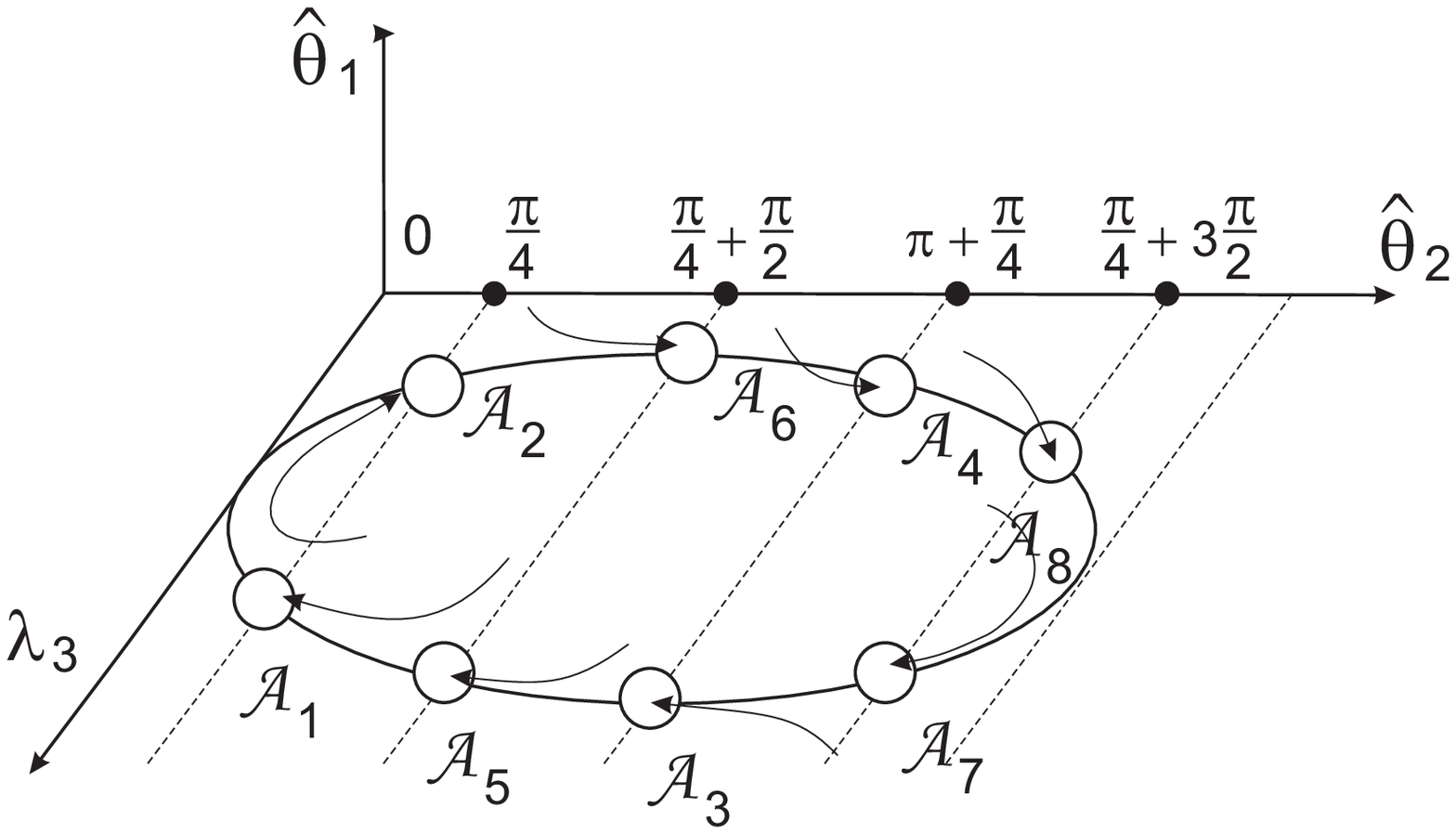}

\vspace{5mm}

\includegraphics[width=0.8\textwidth]{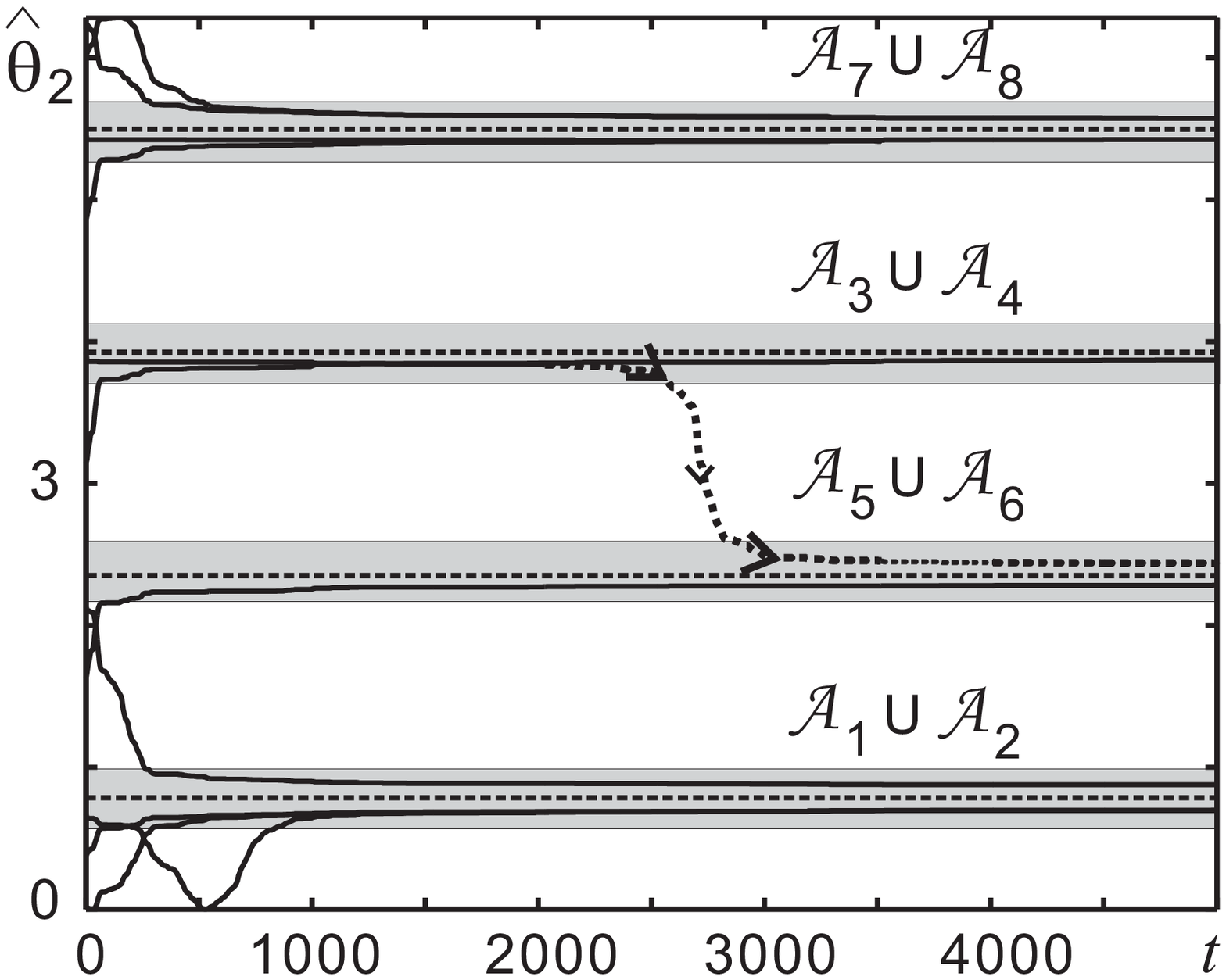}
\end{minipage}
\end{center}
\caption{Template matching of Garner patterns of various
complexity. The patterns (upper row) were rotated by $\pi/4$ and
had various intensity. Depending on the number of their rotational
symmetries they induced different number of invariant sets in the
system state space: two, four and eight respectively. The diagrams
of corresponding phase plots are provided in the middle row.
Estimates of the rotation angle as functions of time for different
initial conditions are depicted in the third row.}
\label{fig:rotation_task}
\end{figure}

We tested system performance for a variety of input images, in
particular the class of Garner patterns \cite{Garner_1962} (see
Figure \ref{fig:rotation_task}, the first row). These patterns are
widely used in the experiments with humans and therefore serve as
a convenient benchmark. Their distinctive property is that their
overall intensity does not vary from one pattern to another. At
the same time their complexity in terms of the number of rotation
and reflection symmetries can vary. In our example the first
pattern has the highest complexity. The second has mid level of
complexity (two symmetrical states) and the third has the lowest
degree of complexity (four symmetrical states).
The second row of Figure \ref{fig:rotation_task} illustrates the
system dynamics involved in invariant template matching for these
patterns. The diagrams represent phase plots of the successful
node $j$ (e.g. for the template subsystem in which the matching
occurs). The third row contains trajectories of the estimates of
the rotation angle $\hat{\theta}_{j,2}(t)$. Each object induces
various number of invariant sets in the template subsystems. The
number of these invariant sets is inversely proportional to
stimulus complexity.  Hence, the higher the complexity the more
time the system requires to converge to an attractor. Thus the
time needed for recognition increases monotonically with  the
stimulus complexity. This is consistent with empirical results
reported in many experimental studies, for instance
\cite{QJP:Lachmann:2005}.

An additional property of our system is that it is capable of
reporting multiple representations of the same object. This is
indicated by the dashed trajectories in Figure
\ref{fig:rotation_task}. Even though the system parameters are
chosen such that trajectories converge to an attractor, we can
still observe meta-stable behavior. This is because the attractors
in our system are of Milnor-type, which implies that trajectories
starting in the vicinity of one attractor may actually belong to
the basin of another attractor. Furthermore, it is even possible
to tune the system in such a way that it will always switch from
one representation to another. The latter property suggests that
our simple system in Table \ref{table:rotation-invariant} can
provide a simple model for visual perception, where spontaneous
switching and perceptual multi-stability  are commonly observed
\cite{SciAm:Attneave:1971}, \cite{TrendsCS:Leopold:1999}.

\subsection{Tracking disturbances in scanning microscopes}

\begin{figure}
\begin{center}
\includegraphics[width=420pt]{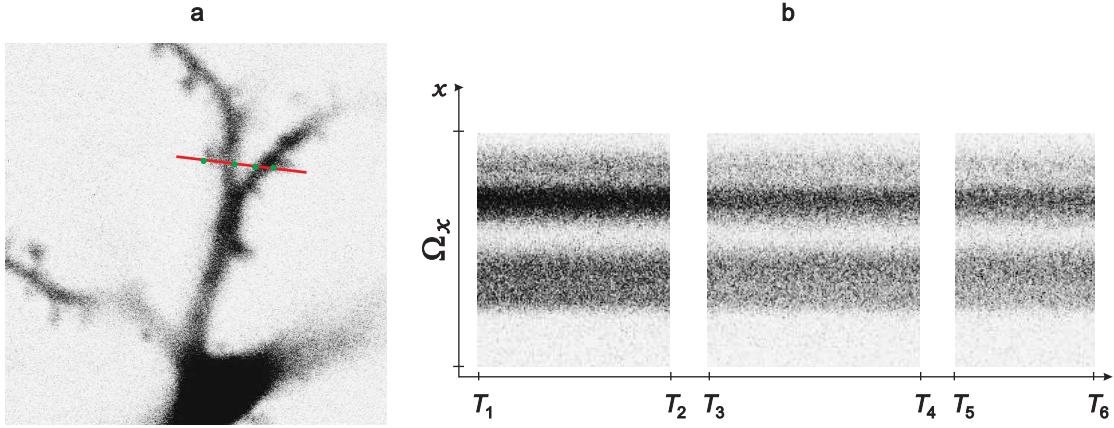}
\caption{ Typical images from the two-photon microscope. Panel
 $a$ shows a dendrite; the domain of scanning (red line) is
in the vicinity of two spines (small protrusions on the dendrite).
Size of the domain is $5.95$ micron, and speed of scanning,
 $v_s$, is $1$ pixel per $2$ micro seconds.  Panel $b$ shows results of scanning
 as a function of time in the beginning (interval $[T_1,T_2]$), in the middle of experiment (domain $[T_3,T_4]$), and in the end
 of the
 experiment (domain $[T_5,T_6]$).}\label{fig:chap:5:visual_system_test}
\end{center}
\end{figure}

We next consider the application of a the template-matching system
with weakly attracting sets to a problem of realistic complexity.
We applied our approach to the problem of tracking morphological
changes in dendritic spines based on measurements received from a
multiphoton scanning microscope {\it in vitro}. A distinctive
property of laser microscopy is that in order to "see" an object
one needs, first, to inject it with a photo-sensitive dye
(fluorophore). The particles of the fluorophore emit photons of
light under external stimulation, thus illuminating an object from
inside the tissue. Typical data from a two-photon microscope are
provided in  Figure \ref{fig:chap:5:visual_system_test}\footnote{
The images are provided by S. Grebenyuk, group of neuronal circuit
mechanisms, RIKEN BSI}.

We addressed the problem of  how to register fast dynamical
changes in spine geometry after application of chemical
stimulation. The measurements were performed on slices.
Measurements of this kind suffer from effects of photobleaching
and diffusion of the dye (see Figure
\ref{fig:chap:5:visual_system_test}), and dependance of the
scattering of the emitted light on the  {\it a-priori} unknown
position of the object in the slice.
On-line estimation and tracking the effects of photobleaching
(intensity) and changes of the object position (blur) in the slice
are therefore necessary.

The measured signal is already a temporal sequence, which fits
nicely to our approach. An inherent feature of scanning microscopy
is that the object is measured using a sequence of scans along
one-dimensional domains (see Figure
\ref{fig:chap:5:visual_system_test}, panel $a$). Hence the objects
in this case are  one-dimensional mappings, and the domain
$\Omega_x$ is an interval $\Omega_x=[x_{\min},x_{\max}]$. For the
particular images we set $x_{\min}=1$ and $x_{\max}=176$, which
corresponds to a scanning line of  $176$ pixels and $5.95$ micro
meters. In order to eliminate measurement noise we we consider the
averaged data in the scanning line over $n$ successive
subsequent trials. 

The measured image, $S_0$, was chosen to be the averaged data
along  the scanning line over $n$ successive subsequent trials.
The template, $S_1$, substituted the averaged measurements of the
object at the initial time $T_1$.

\begin{figure}
\begin{center}
\includegraphics[width=320pt]{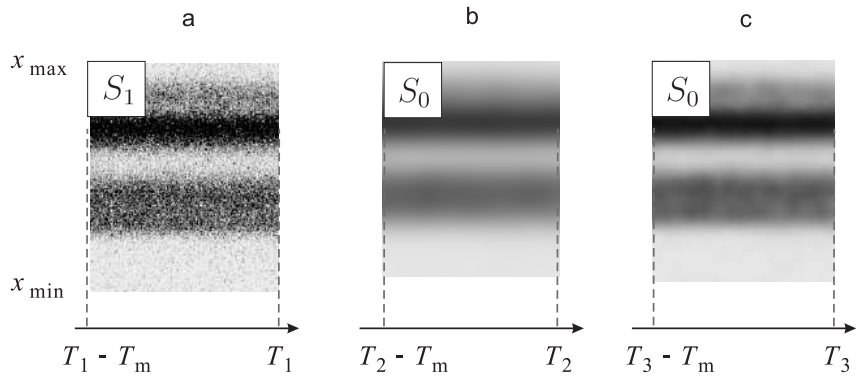}
\caption{Data which has been used to generate the template, $S_1$
(panel $a$), and perturbed measurements $S_0$ at time instants
$T_2$ and $T_3$ (panels $b$ and $c$
respectively).}\label{fig:chap:5:visual_system_example_stimuli}
\end{center}
\end{figure}

Samples of data used to generate $S_1$ are provided in Figure
\ref{fig:chap:5:visual_system_example_stimuli}, $a$. These
correspond to the intensity of the emitted radiation from the
object in the red part of the spectrum for the data shown in
Figure \ref{fig:chap:5:visual_system_test}, $b$, fragment 1.
Measured objects, $S_0$, are the averaged samples of data at the
time instants  $T_i\neq T_1$ (proportional to $T_s$). Focal
distortions were simulated using conventional filters from
Photoshop applied to $S_1$. These fragments are provided in Figure
\ref{fig:chap:5:visual_system_example_stimuli}, panels $b$ and
$c$.

Because the sources of perturbation are the effects of
photobleaching (affecting brightness) and deviations in the object
position in the slice (affecting scattering and leading to blurred
images) the following model of uncertainty was used:
\begin{equation}\label{eq:chap:5:image:perturbation:application}
\begin{split}
\theta_1 f_1(x,\theta_2,t)&=\theta_1 \int_{\Omega_x}
e^{-{{\theta}_{2}}(\xi-x(t))^2} S_1(\xi) d\xi
\end{split},
\end{equation}
where $x(t)$, the scanning trajectory in
(\ref{eq:chap:5:image:perturbation:application}), is defined as:
\[
x(t)=\left\{ \begin{array}{ll}
              x_{\min}+k_s\cdot t & t\leq x_{\max}-x_{\min}\\
              x(t-(x_{\max}-x_{\min})), & t> x_{\max}-x_{\min}
             \end{array}
\right., \ k_s=1.
\]

\begin{figure}
\begin{center}
\includegraphics[width=450pt]{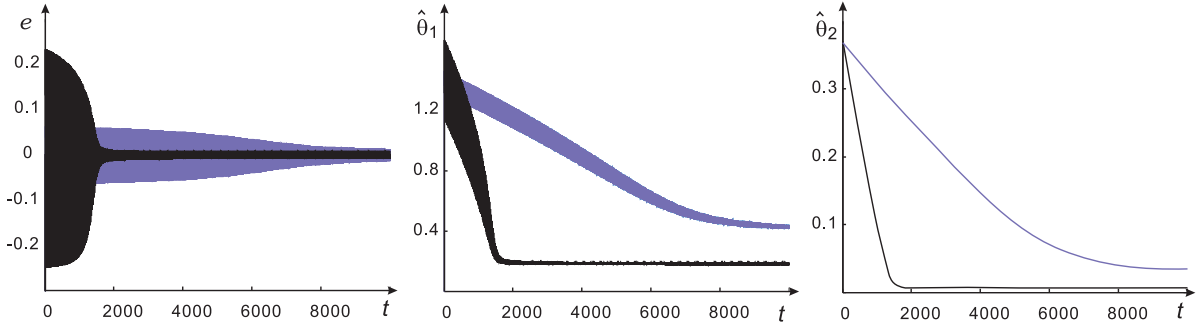}
\caption{Trajectories $e(t)$, $\hat{\theta}_1(t)$,
$\hat{\theta}_2(t)$ as functions of time. Black lines correspond
to  measurements in Fig.
\ref{fig:chap:5:visual_system_example_stimuli}, panel $b$. Blue
lines correspond to the data in Fig.
\ref{fig:chap:5:visual_system_example_stimuli}, panel
$c$.}\label{fig:chap:5:visual_system_example}
\end{center}
\end{figure}

\begin{figure}
\begin{center}
\includegraphics[width=370pt]{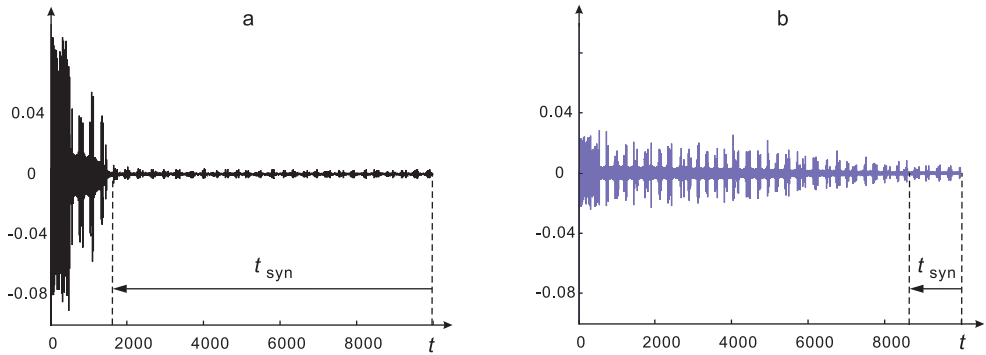}
\caption{ Plots of the synchronization errors $x_0(t)-x_1(t)$ as a
function of time. Panel $a$ corresponds to the data depicted in
Fig. \ref{fig:chap:5:visual_system_example_stimuli}, $b$. Panel
$b$ corresponds to the measurements shown in Fig.
\ref{fig:chap:5:visual_system_example_stimuli},
$c$.}\label{fig:chap:5:visual_system_example_detect}
\end{center}
\end{figure}

Figures \ref{fig:chap:5:visual_system_example},
\ref{fig:chap:5:visual_system_example_detect} show the performance
of our system (\ref{eq:invariance_temporal_coding}),
(\ref{eq:extension_visual_simple}),
(\ref{eq:invariance_adjustment}), (\ref{HR_model_net}) in tracking
focal/brightness perturbations for two measurements  $S_0$.
Figure \ref{fig:chap:5:visual_system_example}  shows the tracking
of unknown modelled perturbations in the images.
 Figure
\ref{fig:chap:5:visual_system_example_detect} shows the
synchronization errors of the detection subsystem. Symbol
$t_{\mathrm{syn}}$ denotes "synchronization time" spent in the
vicinity of the invariant synchronization manifold. As follows
from both figures, the system successfully tracks/reconstructs the
estimates of unknown perturbations applied to the object  (Fig.
\ref{fig:chap:5:visual_system_example}). Coincidence detectors
report synchrony only when the error between the profiles of the
template and object are is sufficiently small (Fig.
\ref{fig:chap:5:visual_system_example_detect}). Furthermore, the
physical time required for recognition on the standard PC was less
than 5 seconds.

\section{Conclusion}\label{sec:conclusion}

We provided a principled solution to the problem of invariant
template matching on the basis of temporal coding of spatial
information. We considered the problem at the levels of
mathematical analysis as well as implementation of specific
recognition systems. Our analysis showed that a solution to the
problem requires to abandon the traditional notion of attractor,
in the Lyapunov sense, for defining a system's target set. As a
substitute we proposed the concept of Milnor attracting sets. At
the level of implementation we provided systems in which such
attractors emerge as a result of external stimulation. These
systems are endowed with mathematical rigor in the form of
conditions sufficient for ensuring global convergence of
trajectories to their target invariant sets. The results provided
are normative in the sense that we require a minimal number of
additional variables and consider as simple structures as
possible.

Even though the proposed system stems from theoretical
considerations, it captures qualitatively a wide range of
phenomena observed in the literature on biological visual
perception. These include multiple time scales for different
modalities during adaptation \cite{VisionResearch:Wolfson:2002},
\cite{Nature:Neuroscience:Webster:2002},
\cite{PLOS:Biology:Smith:2006}, switching and perceptual
multi-stability \cite{SciAm:Attneave:1971},
\cite{TrendsCS:Leopold:1999}, perceptual ambiguity
\cite{Psych_Res:vanLeeuwen:1990}, \cite{Psych_Bull:Hatfield:1985},
empirical observations in mental rotation \cite{QJP:Lachmann:2005}
and decision time distributions \cite{PsychRev:Gilden:2001}. This
motivates our belief that present results may contribute to the
further understanding of visual perception in biological systems,
including humans. 

We demonstrated that the problem of invariant recognition can be
solved by a simple system of ordinary differential equations with
locally Lipschitz right-hand side. This result can be used as an
existence proof for solving the problem of adaptive recognition by
means of recurrent neural networks with fixed weights. Such
systems are being used in various computational tasks
\cite{Conf:IJCNN-02} without any guarantee of a solution. We
guarantee that solutions to realistic recognition problems (e.g.
insuring invariance to rotation, blur, scaling, translation etc.)
can be obtained with networks approximating our prototype system
sufficiently well.

\newcounter{appendixc}
\setcounter{appendixc}{1}
\renewcommand\thesection{Appendix \arabic{appendixc}}

\section{Optimality of sampled representations}\label{appendix:1}

Consider an image $S(x,y)$ and its quantized version $S_q$
obtained from $S(x,y)$ by dividing  domain
$\Omega_x\times\Omega_y$ into the union of finite number of
subsets $\Omega_{x,j}\times\Omega_{y,i}$, $\Omega_x=\cup_j^{N_x}
\Omega_{x,j}$, $\Omega_y=\cup_{i}^{N_y}\Omega_{y,i}$. To each
subset $\Omega_{x,j}\times\Omega_{y,i}$ a value is assigned, which
can be thought of as the median value of $S(x,y)$ over
$\Omega_{x,j}\times\Omega_{y,i}$. We represent $S_q$ as a function
of indices $i,j$: $S_q(i,j)$ and assume that the value of
$S_q(i,j)$ is quantized by a set of $N_s$ levels.

Consider a system of sensors which are capable of measuring image
$S(x,y)$ instantaneously over the given $k$-union of subsets
$\Omega_{x,j}\times\Omega_{y,i}$. The system's cost can be
naturally defined in terms of its  total number of sensors. In
order to measure the entire image at once the system must have at
least $N_x N_y/k$ sensors\footnote{For simplicity we assume that
$N_x N_y$ can be expressed as multiples of $k$.}, so in the
optimal case its cost $C$ should equal $C(k)=N_x N_y/k$.

We estimate the amount of information contained in this sampled
representation of the image. The image is represented by an $N_x
N_y/k$-tuple of elements. Each element is assigned a value, say
$\sigma_i$, from a set of $N_s$ levels with the given probability
$p(\sigma_i)$. Hence the entropy of the representation is
\[
H(k)=-\sum_{i}p(\sigma_i)\log \left(\frac{k}{N_x
N_y}p(\sigma_i)\right)=\log\left(\frac{N_x N_y}{k}\right)-\sum_i
p(\sigma_i)\log p(\sigma_i)
\]
The entropy characterizes the informational content of a
representation, and function $1/H(k)$ its ambiguity.

Overall losses, $Q(k)$, therefore can be defined as a weighted sum
of costs, $C(k)$, and ambiguity, $1/H(k)$:
\[
Q(k)=\lambda_1 C(k) + \lambda_2 1/H(k), \
\lambda_1,\lambda_2\in\Real_{>0}, \ k\in[1,N_x N_y]
\]
Function $Q(k)$ is unimodal and increasing towards the boundaries
of $k$: $k=1$, $k=N_x N_y$. This implies that the minimum of
$Q(k)$ is achieved for some $k=k^\ast\in(1,N_x N_y)$. In other
words, a representation is optimal only when it is sampled, e.g.
induced by a finite, yet neither complete nor elementary,
partition of the domain $\Omega_x\times\Omega_y$.

%


\setcounter{appendixc}{2}

\section{Proofs of the theorems}\label{appendix:3}

{\it Proof of Theorem \ref{theorem:visual_invariance}.} We prove
the theorem in three steps. First, we show that the solution of
the extended system (\ref{eq:invariance_temporal_coding}),
(\ref{eq:extension_visual_simple}),
(\ref{eq:invariance_adjustment}) is bounded. Second, we prove that
there are constants $\rho$, $b$, $\varepsilon$ and time instant
$t'>0$ such that the following holds for system solutions:
\begin{equation}\label{eq:theorem:invariance:transformation:1}
\|\phi_0(t)-\phi_i(t)\|_{\varepsilon}\leq e^{-\rho
(t-t_0)}\|\phi_0(t_0)-\phi_i(t_0)\|_{\varepsilon}+b
\|\theta_2-\hat{\theta}_{i,2}(\tau)\|_{\infty,[t_0,t]} \ \ \forall
\ t\geq t_0>t'
\end{equation}
Third, using this representation we invoke results from (our
paper) and demonstrate that conclusions of the theorem follow.

{\it 1. Boundedness.} To prove boundedness of solutions of the
extended system in forward time let us first consider the
difference $e_i(t)=\phi_0(t)-\phi_i(t)$. According to
(\ref{eq:invariance_temporal_coding}), dynamics of $e_i(t)$ will
be defined as
\begin{equation}\label{eq:theorem:invariance:error_dynamics_1}
\begin{split}
\dot{e}_i&=-\frac{1}{\tau}e_i + k\left(\theta_1 f_i (t,\theta_2)-
\hat{\theta}_{i,1} f_i(t,\hat{\theta}_{i,2})\right) + \epsilon(t)
\end{split}
\end{equation}
Noticing that
\[
\theta_1 f_i (t,\theta_2)- \hat{\theta}_{i,1}
f_i(t,\hat{\theta}_{i,2})= [\theta_1 f_i (t,\theta_2)-\theta_1
f_i(t,\hat{\theta}_{i,2})]+[\theta_1 f_i(t,\hat{\theta}_{i,2})
-\hat{\theta}_{i,1} f_i(t,\hat{\theta}_{i,2})]
\]
and denoting $\delta_{1}=\hat{\theta}_{i,1}- \theta_1$,
$\delta_2(t,\theta_1,\theta_2,\hat{\theta}_{i,2})=\theta_1 f_i
(t,\theta_2)-\theta_1 f_i(t,\hat{\theta}_{i,2})$ we can rewrite
(\ref{eq:theorem:invariance:error_dynamics_1}) as follows
\begin{equation}\label{eq:theorem:invariance:error_dynamics_2}
\begin{split}
\dot{e}_i&=-\frac{1}{\tau}e_i -  \delta_1 [k
f_i(t,\hat{\theta}_{i,2})] +
\delta_2(t,\theta_1,\theta_2,\hat{\theta}_{i,2}) k + \epsilon(t)
\end{split}
\end{equation}

Let us now write equations for $\hat{\theta}_{i,1}-\theta_1$ in
(\ref{eq:adjustment_visual}) in differential form. To do so we
differentiate variable $\hat{\theta}_{i,1}-\theta_1=\delta_1$ with
respect to time, taking into account equations
(\ref{eq:theorem:invariance:error_dynamics_1}),
(\ref{eq:theorem:invariance:error_dynamics_2}):
\begin{equation}\label{eq:theorem:invariance:error_dynamics_3}
\dot{\delta}_1=-\gamma_1\left(\delta_1 [k
f_i(t,\hat{\theta}_{i,2})] -
\delta_2(t,\theta_1,\theta_2,\hat{\theta}_{i,2}) k - \epsilon(t)
\right)
\end{equation}

Variable $\epsilon(t)$ in
(\ref{eq:theorem:invariance:error_dynamics_3}) is bounded
according to (\ref{eq:Delta}). Let us show that
$\delta_2(t,\theta_1,\theta_2,\hat{\theta}_{i,2})$ is also
bounded. First of all notice that the following positive definite
function
\[
V_\lambda=0.5\left(\lambda_2^2+\lambda_3^2\right)
\]
is not growing with time:
\[
\dot{V}=\lambda_2 \gamma_2 \lambda_3
\|\phi_0(t)-\phi_i(t)\|_{\varepsilon}- \lambda_3  \gamma_2
\lambda_2 \|\phi_0(t)-\phi_i(t)\|_{\varepsilon}=0
\]
Furthermore
\begin{equation}\label{eq:theorem:invariance:error_dynamics_4}
\begin{split}
\lambda_2(t)&=r\cdot \sin\left(\gamma_2\int_{t_0}^t \|\phi_0(\tau)-\phi_i(\tau)\|_{\varepsilon}d\tau+\varphi_0\right)\\
\lambda_3(t)&= r\cdot \cos\left(\gamma_2\int_{t_0}^t
\|\phi_0(\tau)-\phi_i(\tau)\|_{\varepsilon}d\tau+\varphi_0\right),
\ r,\varphi_0\in\Real
\end{split}
\end{equation}
Choosing initial conditions $\lambda_2^2(t_0)+\lambda_3^2(t_0)=1$
ensures that $r=1$. Hence, according to equation
(\ref{eq:invariance_adjustment}), variable $\hat{\theta}_{i,2}$
belongs to the interval $[\theta_{2,\min},\theta_{2,\max}]$.

Consider variable
$\delta_2(t,\theta_1,\theta_2,\hat{\theta}_{i,2})$:
\begin{equation}\label{eq:theorem:invariance:error_dynamics_8}
\delta_2(t,\theta_1,\theta_2,\hat{\theta}_{i,2})=\theta_1 f_i
(t,\theta_2)-\theta_1
f_i(t,\hat{\theta}_{i,2}(t))=\theta_1\left(f_i
(t,\theta_2)-f_i(t,\hat{\theta}_{i,2}(t)\right)
\end{equation}
Taking into account notational agreement
(\ref{eq:sampling_functional_final_def}), and properties
(\ref{eq:lipshcitz_constraint}), (\ref{eq:sampling_functional}),
we  conclude that the following estimate holds
\begin{equation}\label{eq:theorem:invariance:error_dynamics_9}
|\delta_2(t,\theta_1,\theta_2,\hat{\theta}_{i,2})|\leq \theta_1
|f_i (t,\theta_2)-f_i(t,\hat{\theta}_{i,2}(t)|\leq \theta_{1,\max}
D D_2 |\theta_2-\hat{\theta}_{i,2}(t)|
\end{equation}
Given that
$\hat{\theta}_{i,2}(t)\in[\theta_{2,\min},\theta_{2,\max}]$  and
using (\ref{eq:theorem:invariance:error_dynamics_9}) we can
provide the following estimate for
$\delta_2(t,\theta_1,\theta_2,\hat{\theta}_{i,2})$:
\begin{equation}\label{eq:theorem:invariance:error_dynamics_13}
|\delta_2(t,\theta_1,\theta_2,\hat{\theta}_{i,2})|\leq
\theta_{1,\max} D D_2 |\theta_{2,\max}-\theta_{2,\min}|
\end{equation}

Let us consider equality
(\ref{eq:theorem:invariance:error_dynamics_3}). According to
condition 1) of the theorem, term
\[
\alpha(t)=k f_i(t,\hat{\theta}_{i,2}(t))
\]
is nonnegative and bounded from below:
\begin{equation}\label{eq:theorem:invariance:error_dynamics_5}
\alpha(t)=k f_i(t,\hat{\theta}_{i,2}(t))\geq k D_3, \ \forall \
t\geq 0
\end{equation}

Taking into account equations
(\ref{eq:theorem:invariance:error_dynamics_3}),
(\ref{eq:theorem:invariance:error_dynamics_5}) we can estimate
$|\delta_1(t)|$ as follows:
\begin{equation}\label{eq:theorem:invariance:error_dynamics_6}
|\delta_1(t)|\leq e^{-\gamma_1
\int_{t_0}^{t}\alpha(\tau)d\tau}|\delta_1(t_0)|+ \gamma_1
e^{-\gamma_1 \int_{t_0}^{t}\alpha(\tau)d\tau}\int_{t_0}^{t}
e^{\gamma_1
\int_{t_0}^{\tau}\alpha(\tau_1)d\tau_1}|\epsilon(\tau)+\delta_2(\tau)k|d\tau
\end{equation}
According to (\ref{eq:Delta}),
(\ref{eq:theorem:invariance:error_dynamics_13}) we have that for
all $t\geq t_0 \geq 0$
\begin{equation}\label{eq:theorem:invariance:error_dynamics_14}
|\epsilon(t)+\delta_2(t)k|\leq
\|\epsilon(\tau)+k\delta_2(\tau)\|_{\infty,[t_0,t]}\leq \Delta +
k\theta_{1,\max} D D_2 |\theta_{2,\max}-\theta_{2,\min}|=M_1
\end{equation}
Furthermore
\begin{equation}\label{eq:theorem:invariance:error_dynamics_15}
\int_{t_0}^{t} e^{\gamma_1
\int_{t_0}^{\tau}\alpha(\tau_1)d\tau_1}d\tau=\frac{1}{\gamma_1}\left(\frac{1}{\alpha(t)}e^{\gamma_1
\int_{t_0}^{t}\alpha(\tau)d\tau}-\frac{1}{\alpha(t_0)}\right) \leq
\frac{1}{\gamma_1 D_3 k} e^{\gamma_1
\int_{t_0}^{t}\alpha(\tau)d\tau}
\end{equation}
Taking into account
(\ref{eq:theorem:invariance:error_dynamics_6}),
(\ref{eq:theorem:invariance:error_dynamics_14}), and
(\ref{eq:theorem:invariance:error_dynamics_15}) we can obtain the
following estimate:
\begin{equation}\label{eq:theorem:invariance:error_dynamics_7}
\begin{split}
|\delta_1(t)|\leq & e^{-\gamma_1 k D_4
(t-t_0)}|\delta_1(t_0)|+\frac{M_1}{D_3 k}
\end{split}
\end{equation}
Inequality (\ref{eq:theorem:invariance:error_dynamics_7}) proves
that $\delta_1(t)$ is bounded.

In order to complete this step of the proof it is sufficient to
show that $e_i(t)$ is bounded. This would automatically imply
boundedness of $\phi_i(t)$, thus confirming boundedness of state
of the extended system. To show boundedness of $e_i(t)$ let us
write the closed-form solution of
(\ref{eq:theorem:invariance:error_dynamics_1}):
\begin{equation}\label{eq:theorem:invariance:error_dynamics_0}
e_i(t)=e^{-\frac{(t-t_0)}{\tau}}e_i(t_0)+e^{-\frac{t}{\tau}}\int_{t_0}^t
e^{\frac{\tau_1}{\tau}}\left(\delta_1(\tau_1)\alpha(\tau_1)+k
\delta_2(\tau_1)+\epsilon(\tau_1)\right)d\tau_1
\end{equation}
Using (\ref{eq:theorem:invariance:error_dynamics_14}) and
(\ref{eq:theorem:invariance:error_dynamics_7}) we can derive that
\begin{equation}\label{eq:theorem:invariance:error_dynamics_17}
|e_i(t)|\leq e^{-\frac{(t-t_0)}{\tau}}|e_i(t_0)|+ M_1 \tau \left(1
+  \frac{D_4}{D_3}\right) + \epsilon_1(t),
\end{equation}
where $\epsilon_1(t)$ is an exponentially decaying term:
\begin{equation}\label{eq:theorem:invariance:error_dynamics_18}
|\epsilon_1(t)|\leq e^{-\gamma_1 k D_3 (t-t_0)}\left(\frac{
1-e^{-\left(\frac{1}{\tau}-\gamma_1 k D_3\right)(t-t_0)}
}{\frac{1}{\tau}-\gamma_1 k
D_3}\right)|\theta_1-\hat{\theta}_{i,1}(t_0)|.
\end{equation}
As follows from (\ref{eq:theorem:invariance:error_dynamics_4}),
(\ref{eq:theorem:invariance:error_dynamics_7}),
(\ref{eq:theorem:invariance:error_dynamics_17}),
(\ref{eq:theorem:invariance:error_dynamics_18}), variables
 $e_i(t)$, $\hat{\theta}_{i,1}(t)$, $\hat{\theta}_{i,2}(t)$ are
bounded. Hence state of the extended system is bounded in forward
time.

{\it 2. Transformation.} Let us now show that there exists a time
instance $t'$ and constants $\rho,c\in\Real_{>0}$ such that the
dynamics of $e_i(t)=\phi_0(t)-\phi_i(t)$ satisfies inequality
(\ref{eq:theorem:invariance:transformation:1}). In order to do so
we first show that term
\[
\delta_1(t) k f_i(t,\hat{\theta}_{i,2}(t))
\]
in (\ref{eq:theorem:invariance:error_dynamics_2}) can be estimated
as
\begin{equation}\label{eq:theorem:invariance:transformation:2}
|\delta_1(t) k f_i(t,\hat{\theta}_{i,2}(t))|\leq M_2
|\theta_2-\hat{\theta}_{i,2}(t)|+ \Delta_2 + \epsilon_2(t)
\end{equation}
where $M_2$, $\Delta_2$ are positive constants and $\epsilon_2(t)$
is a function of time which converges to zero asymptotically with
time.

According to (\ref{eq:theorem:invariance:error_dynamics_6}) the
following holds
\[
|\delta_1(t)|\leq e^{-\gamma_1
\int_{t_0}^{t}\alpha(\tau)d\tau}|\delta_1(t_0)|+ \gamma_1
e^{-\gamma_1 \int_{t_0}^{t}\alpha(\tau)d\tau}\int_{t_0}^{t}
e^{\gamma_1
\int_{t_0}^{\tau}\alpha(\tau_1)d\tau_1}|\epsilon(\tau)+\delta_2(\tau)k|d\tau
\]
Taking into account (\ref{eq:Delta}),
(\ref{eq:theorem:invariance:error_dynamics_5}) we can conclude
that
\begin{equation}\label{eq:theorem:invariance:transformation:3}
|\delta_1(t)|\leq e^{-\gamma_1 k D_3 (t-t_0)}|\delta_1(t_0)|+
\frac{\Delta}{k D_3}+ \gamma_1 e^{-\gamma_1
\int_{t_0}^{t}\alpha(\tau)d\tau}\int_{t_0}^{t} e^{\gamma_1
\int_{t_0}^{\tau}\alpha(\tau_1)d\tau_1}|\delta_2(\tau)k|d\tau
\end{equation}
Substituting  (\ref{eq:theorem:invariance:error_dynamics_9}) into
(\ref{eq:theorem:invariance:transformation:3}) results in
\begin{equation}\label{eq:theorem:invariance:transformation:4}
\begin{split}
|\delta_1(t)|\leq& e^{-\gamma_1 k D_3 (t-t_0)}|\delta_1(t_0)|+
\frac{\Delta}{k D_3}+ \\
& \gamma_1 e^{-\gamma_1
\int_{t_0}^{t}\alpha(\tau)d\tau}\int_{t_0}^{t} e^{\gamma_1
\int_{t_0}^{\tau}\alpha(\tau_1)d\tau_1}|\theta_2-\hat{\theta}_{i,2}(\tau)|
d\tau \cdot \left(k \theta_{1,\max} D D_2\right)
\end{split}
\end{equation}

Consider the following term in
(\ref{eq:theorem:invariance:transformation:4}):
\begin{equation}\label{eq:theorem:invariance:error_dynamics_11}
\int_{t_0}^{t} e^{\gamma_1
\int_{t_0}^{\tau}\alpha(\tau_1)d\tau_1}|\theta_2-\hat{\theta}_{i,2}(\tau)|
d\tau
\end{equation}
Integration of (\ref{eq:theorem:invariance:error_dynamics_11}) by
parts yields
\begin{equation}\label{eq:theorem:invariance:error_dynamics_12}
\begin{split}
&\int_{t_0}^{t} e^{\gamma_1
\int_{t_0}^{\tau}\alpha(\tau_1)d\tau_1}|\theta_2-\hat{\theta}_{i,2}(\tau)|
d\tau=\frac{1}{\gamma_1} \left(\frac{1}{\alpha(t)} e^{\gamma_1
\int_{t_0}^{t}\alpha(\tau)d\tau}|\theta_2-\hat{\theta}_{i,2}(t)|-\frac{|\theta_2-\hat{\theta}_{i,2}(t_0)|}{\alpha(t_0)}\right)\\
&-\frac{1}{\gamma_1} \int_{t_0}^{t}\frac{1}{\alpha(\tau)}
e^{\gamma_1
\int_{t_0}^{\tau}\alpha(\tau_1)d\tau_1}\left(\frac{d|\theta_2-\hat{\theta}_{i,2}(\tau)|}{d\tau}\right)
d\tau\leq\\
& \frac{1}{\gamma_1 k D_3} e^{\gamma_1
\int_{t_0}^{t}\alpha(\tau)d\tau}|\theta_2-\hat{\theta}_{i,2}(t)|+
\frac{1}{\gamma_1 k D_3} \int_{t_0}^t e^{\gamma_1
\int_{t_0}^{\tau}\alpha(\tau_1)d\tau_1}\left|\frac{d|\theta_2-\hat{\theta}_{i,2}(\tau)|}{d\tau}\right|
d\tau
\end{split}
\end{equation}
Given that
\[
\hat{\theta}_{2,i}=\theta_{2,\min}+\frac{\theta_{2,\max}-\theta_{2,\min}}{2}(\lambda_2(t)+1),
\]
we can estimate the derivative
$d|\theta_2-\hat{\theta}_{2,i}(t)|/dt$ as follows:
\begin{equation}\label{eq:theorem:invariance:transformation:5}
\frac{d|\theta_2-\hat{\theta}_{i,2}(t)|}{dt}\leq
\frac{\theta_{2,\max}-\theta_{2,\min}}{2}\cdot\gamma_2\cdot
|\phi_0(t)-\phi_i(t)|
\end{equation}
Notice that the value of $|\phi_0(t)-\phi_i(t)|=e_i(t)$ in
(\ref{eq:theorem:invariance:transformation:5}) can be estimated
according to (\ref{eq:theorem:invariance:error_dynamics_17}) as
\[
|\phi_0(t)-\phi_i(t)|\leq M_1 \tau \left(1 +
\frac{D_4}{D_3}\right) + \mu_1(t),
\]
where $\mu_1(t)\sim\epsilon_1(t)+e_i(t)e^{-\frac{(t-t_0)}{\tau}}$
is an asymptotically decaying term.

Hence, taking into account
(\ref{eq:theorem:invariance:error_dynamics_15}),
(\ref{eq:theorem:invariance:error_dynamics_17}),
(\ref{eq:theorem:invariance:transformation:4}),
(\ref{eq:theorem:invariance:error_dynamics_12}), and
(\ref{eq:theorem:invariance:transformation:5}) we may conclude
that the following inequality holds
\[
|\delta_1(t)|\leq \frac{\theta_{1,\max} D D_2}{D_3
}|\theta_2-\hat{\theta}_{i,2}(t)|+\frac{\Delta}{D_3
k}+\frac{\gamma_2}{\gamma_1}\frac{\theta_{1,\max} D D_2}{D_3^2 k}
\frac{\theta_{2,\max}-\theta_{2,\min}}{2} M_1 \tau
\left(1+\frac{D_4}{D_3}\right) + \mu(t)
\]
where $\mu(t)$ is asymptotically vanishing term. Therefore
(\ref{eq:theorem:invariance:transformation:2}) holds with the
following values of $M_2$ and $\Delta_2$:
\begin{equation}\label{eq:theorem:invariance:transformation:6}
\begin{split}
M_2&=\frac{k \theta_{1,\max} D D_2 D_4}{D_3}\\
\Delta_2&= \frac{\gamma_2}{\gamma_1}\left[\frac{\theta_{1,\max} D
D_2 D_4}{(D_3)^2}M_1 \tau
\left(1+\frac{D_4}{D_3}\right)\frac{\theta_{2,\max}-\theta_{2,\min}}{2}\right]
+ \frac{\Delta D_4}{D_3}
\end{split}
\end{equation}

To finalize this step of the proof consider variable $e_i(t)$ for
$t\in[t_1,\infty]$, $t_1\geq t_0$. According to
(\ref{eq:theorem:invariance:error_dynamics_0}),
(\ref{eq:theorem:invariance:error_dynamics_9}) we have that
\begin{equation}\label{eq:theorem:invariance:transformation:7}
\begin{split}
|e_i(t)|&\leq e^{-\frac{t-t_1}{\tau}}|e_i(t_1)|+ \tau M_2 \|\theta_2-\hat{\theta}_{i,2}(t)\|_{\infty,[t_1,t]}+\\
&\tau \Delta_2 \left(1-e^{-\frac{(t-t_1)}{\tau}}\right) + \tau
\|\epsilon_2(t)\|_{\infty,[t_1,\infty]}
\left(1-e^{-\frac{(t-t_1)}{\tau}}\right) +\\
& \tau k \theta_{1,\max} D D_2
\|\theta_2-\hat{\theta}_{i,2}(t)\|_{\infty,[t_1,t]} +  \tau \Delta
\left(1-e^{-\frac{(t-t_1)}{\tau}}\right)
\end{split}
\end{equation}
Regrouping terms in (\ref{eq:theorem:invariance:transformation:7})
yields:
\[
\begin{split}
|e_i(t)|-\tau\left(\Delta_2 + \Delta +
\|\epsilon_2(t)\|_{\infty,[t_1,\infty]}\right)&\leq
e^{-\frac{t-t_1}{\tau}} \left(|e_i(t_1)| - \tau \left(\Delta_2 +
\Delta +
\|\epsilon_2(t)\|_{\infty,[t_1,\infty]}\right)  \right)  \\
& + \tau (M_2 + k \theta_{1,\max} D D_2)
\|\theta_2-\hat{\theta}_{i,2}(t)\|_{\infty,[t_1,t]}
\end{split}
\]
Denoting
\begin{equation}\label{eq:theorem:invariance:transformation:8}
\Delta'=\tau\left(\Delta_2 + \Delta +
\|\epsilon_2(t)\|_{\infty,[t_1,\infty]}\right)
\end{equation}
we can obtain
\begin{equation}\label{eq:theorem:invariance:transformation:9}
\begin{split}
|e_i(t)|-\Delta'& \leq e^{-\frac{t-t_1}{\tau}} \left(|e_i(t_1)| -
\Delta'  \right) + \tau (M_2 + k \theta_{1,\max} D D_2)
\|\theta_2-\hat{\theta}_{i,2}(t)\|_{\infty,[t_1,t]}\\
& \leq e^{-\frac{t-t_1}{\tau}} \|e_i(t_1)\|_{\Delta'} + \tau (M_2
+ k \theta_{1,\max} D D_2)
\|\theta_2-\hat{\theta}_{i,2}(t)\|_{\infty,[t_1,t]}
\end{split}
\end{equation}
Given that
\[
\|e_i(t)\|_{\Delta'}=\left\{
                    \begin{array}{ll}
                    |e_i(t)|-\Delta', & |e_i(t)|> \Delta'\\
                    0, & |e_i(t)|\leq \Delta'
                    \end{array}
                    \right.
\]
and taking into account inequality
(\ref{eq:theorem:invariance:transformation:9}), we can conclude
that
\begin{equation}\label{eq:theorem:invariance:transformation:10}
\|e_i(t)\|_{\Delta'} \leq e^{-\frac{t-t_1}{\tau}}
\|e_i(t_1)\|_{\Delta'} + \tau (M_2 + k \theta_{1,\max} D D_2)
\|\theta_2-\hat{\theta}_{i,2}(t)\|_{\infty,[t_1,t]}
\end{equation}
Because equations (\ref{eq:theorem:invariance:transformation:7})
-- (\ref{eq:theorem:invariance:transformation:10}) hold for any
$t_1\in(t_0,\infty]$ and that
\[
\limsup_{{t_1}\rightarrow\infty}
\|\epsilon_2(t)\|_{\infty,[t_1,\infty]}=0
\]
for every
\[
\varepsilon> \tau (\Delta+\Delta_2)
\]
there exists a time instant $t'\geq t_0$ such that the following
inequality is satisfied
\begin{equation}\label{eq:theorem:invariance:transformation:11}
\|e_i(t)\|_{\varepsilon} \leq e^{-\frac{t-t_1}{\tau}}
\|e_i(t_1)\|_{\varepsilon} + \tau (M_2 + k \theta_{1,\max} D D_2)
\|\theta_2-\hat{\theta}_{i,2}(t)\|_{\infty,[t_1,t]}
\end{equation}
for all $t\geq t_1\geq t'$.  This proves
(\ref{eq:theorem:invariance:transformation:1}) for
$\rho=\frac{1}{\tau}$, $b=\tau (M_2 + k \theta_{1,\max} D D_2)$.
Hence the second step of the proof is completed.

{\it 3. Convergence.} In order to prove convergence we employ the
following result from \cite{ArXive:Non-uniform:2006}:
\begin{lem}[Corollary 3 in \cite{ArXive:Non-uniform:2006}]\label{cor:small_gain_like}
Consider the following interconnection of two systems:
\begin{equation}\label{eq:interconnection}
\begin{split}
\mathcal{S}_a: \ &\norms{\bfx(t)}\leq \norms{\bfx(t_0)}\cdot
\beta_t(t-t_0)+c\cdot\|h(\tau)\|_{\infty,[t_0,t]}, \ \bfx:\Real_{\geq 0}\rightarrow \Real^n \\
\mathcal{S}_w: \  &\int_{t_0}^{t} \underline{\gamma}
\norms{\bfx(\tau)} d\tau \leq h(t_0)-h(t)\leq \int_{t_0}^{t}
\bar{\gamma} \norms{\bfx(\tau)} d\tau, \ \forall \ t\geq t_0, \
t_0\in\Real_+
\end{split}
\end{equation}
where the systems $\mathcal{S}_a$, $\mathcal{S}_w$ are
forward-complete\footnote{We say that a system is forward-complete
iff its state is defined in forward time for all admissible
inputs. For the system $\mathcal{S}_a$ the inputs are functions
$h(t)$ from $L_{\infty}[t_0,t]$. For the system $\mathcal{S}_w$
the inputs are locally-bounded in $t$ functions $\bfx(t)$.},
function $\beta_t: \Real_{\geq 0}\rightarrow \Real_{\geq 0}$ is
strictly monotone and decreases to zero as $t\rightarrow \infty$.
Let us suppose that the following condition is satisfied
\begin{equation}\label{eq:small_gain_like_condition}
\bar{\gamma} \cdot c \cdot \mathcal{G}<1,
\end{equation}
where
\[
\mathcal{G}=
\beta_t^{-1}\left(\frac{d}{\kappa}\right)\frac{k}{k-1}\left(\beta_t(0)\left(1+\frac{\kappa}{1-d}\right)+1\right)
\]
for some $d\in(0,1)$, $\kappa\in(1,\infty)$.

Then there exists a set $\Omega_\gamma$ of initial conditions
corresponding to trajectories $\bfx(t)$, $h(t)$ such that
\[
\limsup_{t\rightarrow\infty} \norms{\bfx(t)}\leq c\cdot h(t_0); \
 \  h(t) \in [0,h(t_0)] \ \forall \ t\geq t_0
\]
In particular, $\Omega_\gamma$ contains the following domain
\[
\begin{split}
\norms{\bfx(t_0)}\leq \frac{1}{\beta_t(0)}
\left[\frac{1}{\bar{\gamma}}\left(\beta_t^{-1}\left(\frac{d}{\kappa}\right)\right)^{-1}\frac{k-1}{k}-
c
\left(\beta_t(0)\left(1+\frac{\kappa}{1-d}\right)+1\right)\right]h(t_0).
\end{split}
\]
\end{lem}

In order to apply Lemma \ref{cor:small_gain_like} we need to
further transform equations
(\ref{eq:theorem:invariance:error_dynamics_4}),
(\ref{eq:theorem:invariance:transformation:11}) and
\begin{equation}\label{eq:theorem:invariance:convergence:1}
\hat{\theta}_{i,2}(t)=\theta_{2,\min}+\frac{\theta_{2,\max}-\theta_{2,\min}}{2}(\lambda_2(t)+1)
\end{equation}
into the form of equation (\ref{eq:interconnection}). First, we
notice that for every
$\theta_2\in[\theta_{2,\min},\theta_{2,\max}]$ there always exists
a real number $\lambda^\ast\in[-1,1]$ such that
\[
\theta_2=\theta_{2,\min}+\frac{\theta_{2,\max}-\theta_{2,\min}}{2}(\lambda_2^\ast+1)
\]
Hence, denoting
\[
c= \tau (M_2 + k \theta_{1,\max} D D_2)
\frac{\theta_{2,\max}-\theta_{2,\min}}{2}
\]
and using (\ref{eq:theorem:invariance:transformation:11}) we
ascertain that the following holds for solutions of system
(\ref{eq:invariance_temporal_coding}),
(\ref{eq:extension_visual_simple}),
(\ref{eq:invariance_adjustment}):
\begin{equation}\label{eq:theorem:invariance:convergence:2}
\|e_i(t)\|_{\varepsilon} \leq e^{-\frac{t-t_1}{\tau}}
\|e_i(t_1)\|_{\varepsilon} + c
\|\lambda_2^\ast-\lambda_2(t)\|_{\infty,[t_1,t]}
\end{equation}
for $\varepsilon>\tau (\Delta+\Delta_2)$, and $t\geq t_1\geq t'$.

Consider the difference $\lambda_2^\ast-\lambda_2(t)$. According
to (\ref{eq:theorem:invariance:error_dynamics_4}) we have
\begin{equation}\label{eq:theorem:invariance:convergence:3}
|\lambda_2^\ast -\lambda_2(t)|\leq |\sigma^\ast - \int_{t_1}^t
\gamma_2 \|e_i(\tau)\|_\varepsilon - \varphi_0|, \
\lambda_2^\ast=\sin(\sigma^\ast)
\end{equation}
Denoting
\begin{equation}\label{eq:theorem:invariance:convergence:3_1}
h(t)= \sigma^\ast - \int_{t_1}^t \gamma_2
\|e_i(\tau)\|_\varepsilon - \varphi_0
\end{equation}
and taking into account
(\ref{eq:theorem:invariance:convergence:2}), we therefore obtain
the following equations
\begin{equation}\label{eq:theorem:invariance:convergence:4}
\begin{split}
\|e_i(t)\|_{\varepsilon}& \leq e^{-\frac{t-t_1}{\tau}}
\|e_i(t_1)\|_{\varepsilon} + c \|h(t)\|_{\infty,[t_1,t]}\\
h(t_1)-h(t)&=\int_{t_1}^t \gamma_2 \|e_i(\tau)\|_\varepsilon d\tau
\end{split}
\end{equation}
Equations (\ref{eq:theorem:invariance:convergence:4}) are a
particular case of equations (\ref{eq:interconnection}) to which
Lemma \ref{cor:small_gain_like} applies. In system
(\ref{eq:theorem:invariance:convergence:4}), however, function
$\beta_t(t)$ is defined as $\beta_t(t)=e^{-\frac{t}{\tau}}$. Hence
\[
\beta_t^{-1}(t)=-\tau \ln(t)
\]
Therefore, according to Lemma \ref{cor:small_gain_like},
satisfying inequality
\begin{equation}\label{eq:theorem:invariance:convergence:5}
\gamma_2 \cdot c \cdot \tau
\ln\left(\frac{\kappa}{d}\right)\frac{k}{k-1}\left(\left(1+\frac{\kappa}{1-d}\right)+1\right)
<1
\end{equation}
for some $\kappa\in(1,\infty)$, $d\in(0,1)$ ensures existence of
initial conditions $e_i(t_1)$, $h(t_1)$ such that $h(t)$ is
bounded. Given that
\[
\min_{\kappa\in (1,\infty), ,
d\in(0,1)}\ln\left(\frac{\kappa}{d}\right)\frac{k}{k-1}\left(\left(1+\frac{\kappa}{1-d}\right)+1\right)\approx
15.6886<16
\]
we can rewrite condition
(\ref{eq:theorem:invariance:convergence:5}) in a more
conservative, yet simpler form:
\[
\gamma_2 \cdot c \cdot \tau < \frac{1}{16}
\]
Taking into account notations
(\ref{eq:theorem:invariance:transformation:6}),
(\ref{eq:theorem:invariance:convergence:3_1}) we can rewrite this
inequality as follows:
\[
\gamma_2 < \left(\frac{1}{4 \tau}\right)^2\left[k \theta_{1,\max}
D D_2
\left(1+\frac{D_4}{D_3}\right)\left(\frac{\theta_{2,\max}-\theta_{2,\min}}{2}\right)\right]^{-1}
\]

Notice that because the function $\sin(\cdot)$ is periodic, the
value of $\sigma^\ast$ in
(\ref{eq:theorem:invariance:convergence:3}) and, subsequently the
value of $h(t_1)$, can be chosen arbitrarily large. Hence for any
finite $e_i(t_1)$ and $\varphi_0$ there will always exist
$\sigma^\ast$ and $h(t_1)$ such that variable $h(t)$ is bounded.

Taking into account that $h(t)$ is monotone and bounded, we can
conclude that according to the Bolzano-Weierstrass theorem
function $h(t)$ has a limit in $[0, h(t_1)]$:
\[
\exists h^\ast\in [0, h(t_1)]: \
\lim_{t\rightarrow\infty}h(t)=h^\ast.
\]
This in turn implies that
\[
\lim_{t\rightarrow\infty} \int_{t_1}^t
\gamma_2\|e_i(\tau)\|_\varepsilon
d\tau=\sigma^\ast-\varphi_0-h^\ast < \infty
\]
Therefore
\[
\exists \ \theta_2'\in [\theta_{2,\min},\theta_{2,\max}]: \
\lim_{t\rightarrow\infty}\hat{\theta}_{i,2}(t)=\theta_{2,\min}+\frac{\theta_{2,\max}-\theta_{2,\min}}{2}(\sin
(\sigma^\ast-\varphi_0-h^\ast)+1)=\theta_2'
\]
Moreover, because $\|e_i(t)\|_\varepsilon$ is uniformly continuous
in $t$, convergence of $\|e_i(t)\|_\varepsilon$ to zero as
$t\rightarrow\infty$ follows immediately from Barbalat's lemma.
{\it The theorem is proven.}

{\it Proof of Theorem \ref{theorem:coincidence_detectors}.} The
proof consists of three major steps. First, we show that single
Hindmarsh-Rose oscillator is a semi-passive system with radially
unbounded storage function \cite{IntJBC:Pogromsky-98}. In other
words, system:
\begin{equation}\label{eq:HRmodel}
\begin{split}
\dot{x}&=-a x^3 + b x^2 + y -z + I +u\\
\dot{y}&=c - d x^2 - y\\
\dot{z}&=\varepsilon(s(x+x_0)-z), \ \ a,b,c,d,\varepsilon,s>0
\end{split}
\end{equation}
obeys the following inequality
\begin{equation}\label{eq:HR_Semi-Passivity}
V(x(t),y(t),z(t))-V(x(0),y(0),z(0))\leq \int_0^t  x(\tau) u (\tau)
- H(x(\tau),y(\tau),z(\tau))d\tau.
\end{equation}
where function $H(\cdot)$ is non-negative outside a ball in
$\Real^3$, and function $V$ is positive definite and radially
unbounded. Second, similar to  \cite{IntJBC:Pogromsky-98}, we show
that semi-passivity of (\ref{eq:HRmodel}) implies that solutions
of the coupled system (\ref{HR_model_net}) are bounded. Third, for
an arbitrary pair $(i,j)$ of the oscillators we present a
nonnegative function such that properties
(\ref{eq:HR_synch_pair}), (\ref{eq:HR_synch_pair_2}) hold for
sufficiently large values of $\gamma$. Then we use the comparison
lemma \cite{Khalil:2002} to complete the proof.

{\it 1) Semi-passivity of the Hindmash-Rose oscillator.} Let us
consider the following class of functions $V$:
\[
V(x,y,z)=\frac{1}{2}\left(c_1 x^2+c_2 y^2+ c_3 z^2\right)
\]
Then showing existence of a function $V$ from the above class
which, in addition satisfies inequality
\begin{equation}\label{eq:Lyapunov_Candidate}
\dot{V}\leq x u - H(x,y,z),
\end{equation}
where $H$ is non-negative outside some ball in $\Real^3$, would
imply semi-passivity of (\ref{eq:HRmodel}).

Consider the time-derivative of $V$:
\begin{multline}\label{eq:HRmodel:dV}
    \dot{V}(x,y,z) = -c_1ax^4-c_2dx^2y
        -c_2y^2+c_1xy\\
        -c_3\eps z^2+(c_3\eps s-c_1)xz+
        c_1bx^3+c_1Ix+c_2cy+c_3\eps sx_0z + c_1 xu.
\end{multline}
Let us rewrite (\ref{eq:HRmodel:dV}) such that the cross terms
$xy$, $xz$ and $x^2y$ are expressed in terms of the powers of
$x,y,z$ and their sums. In order to do this we employ the
following three equalities:
\begin{equation}\label{eq:equality1}
    -c_2y^2+c_1xy=-c_2\Var{2}y^2-c_2(1-\Var{2})\Big(y-\frac{c_1}{2c_2(1-\Var{2})}x\Big)^2+\frac{{c_1}^2}{4c_2(1-\Var{2})}x^2
\end{equation}
\begin{multline}
    -c_3\eps z^2+(c_3\eps s-c_1)xz=-c_3\eps \Var{3}z^2-c_3\eps(1-\Var{3})\Big(z-\frac{c_3\eps s-c_1}{2c_3\eps(1-\Var{3})}x\Big)^2+
    \frac{(c_3\eps s-c_1)^2}{4c_3\eps(1-\Var{3})}x^2
\end{multline}
\begin{multline}\label{eq:equality3}
    -c_1ax^4-c_2dx^2y=-c_1a\Var{1}x^4-c_1a(1-\Var{1})\Big(x^2+\frac{c_2d}{2c_1a(1-\Var{1})}y\Big)^2+
    \frac{(c_2d)^2}{4c_1a(1-\Var{1})}y^2
\end{multline}
In what follows we will assume that constants $\Var{1}$, $\Var{2}$
and $\Var{3}$ in (\ref{eq:equality1})--(\ref{eq:equality3}) are
chosen arbitrarily in the interval $(0,1)$: $0<\Var{i}<1,\;
i=1,2,3$.

Taking equalities (\ref{eq:equality1})--(\ref{eq:equality3}) into
account, we can rewrite the time derivative  of $V$ (equation
(\ref{eq:HRmodel:dV})) in the following form:
\begin{multline}\label{eq:HRmodel:dV_1}
    \dot{V}(x,y,z) = -c_1a(1-\Var{1})\Big(x^2+\frac{c_2d}{2c_1a(1-\Var{1})}y\Big)^2
        -c_2(1-\Var{2})\Big(y-\frac{c_1}{2c_2(1-\Var{2})}x\Big)^2+\\
        -c_3\eps(1-\Var{3})\Big(z-\frac{c_3\eps s-c_1}{2c_3\eps(1-\Var{3})}x\Big)^2
        -c_2\Big(\Var{2}-\frac{c_2d^2}{4c_1a(1-\Var{1})}\Big)y^2+c_2cy+\\
        -c_3\eps\Var{3}z^2+c_3\eps sx_0z
        -c_1a\Var{1}x^4+c_1bx^3+
        \Big(\frac{{c_1}^2}{4c_2(1-\Var{2})}+\frac{(c_3\eps
        s-c_1)^2}{4c_3\eps(1-\Var{3})}\Big)x^2+c_1Ix + c_1 xu
\end{multline}

Our goal is to express the right-hand side of
(\ref{eq:HRmodel:dV_1}) in the following form:
\begin{equation}\label{eq:equality_PreSemiPassivity}
\dot{V}\leq c_1 xu + \left(M - H_0(x,y,z)\right),
\end{equation}
where $H_0(x,y,z)$ is  a radially unbounded nonnegative function
outside a ball in $\Real^3$, and $M$ is a constant. For this
reason we select constants $\lambda_2, c_2$ in
(\ref{eq:Lyapunov_Candidate}) as follows:
\begin{equation}\label{eq:HRmodel:dV_C2}
\Var{2}-\frac{c_2d^2}{4c_1a(1-\Var{1})}>0, \ \
 \mathrm{or} \ \
\frac{c_2}{c_1}<\frac{4a\Var{2}(1-\Var{1})}{d^2}.
\end{equation}

Noticing that
\begin{multline}
    -c_2\Big(\Var{2}-\frac{c_2d^2}{4c_1a(1-\Var{1})}\Big)y^2+c_2cy=\\
        -c_2\big(\Var{2}-\frac{c_2d^2}{4c_1a(1-\Var{1})}\big)\Big(y-\frac{2cc_1a(1-\Var{1})}{4\Var{2}c_1a(1-\Var{1})-c_2d^2}\Big)^2+
        \frac{c_1c_2c^2a(1-\Var{1})}{4\Var{2}c_1a(1-\Var{1})-c_2d^2}
\end{multline}
\begin{equation}
    -c_3\eps\Var{3}z^2+c_3\eps sx_0z=
    -c_3\eps\Var{3}\Big(z-\frac{sx_0}{2\Var{3}}\Big)^2+\frac{c_3\eps s^2{x_0}^2}{4\Var{3}}
\end{equation}
proves representation (\ref{eq:equality_PreSemiPassivity}) for any
fixed $x=\mathrm{const}$. In order to show that
(\ref{eq:equality_PreSemiPassivity}) holds with respect to the
complete set of variables, e.g. $(x,y,z)$ we use the following
sequence of equalities:
\begin{multline}\label{eq:HRmodel:dV_in_x}
    -c_1a\Var{1}x^4+c_1bx^3+\Big(\frac{{c_1}^2}{4c_2(1-\Var{2})}+\frac{(c_3\eps s-c_1)^2}{4c_3\eps(1-\Var{3})}\Big)x^2+c_1Ix = (\mathrm{see \ \ notations \ \ below})\\
    -a_0x^4+a_1x^3+a_2x^2+a_3x+a_4= \\
    -b_0x^4-\big(x-b_1\big)^4+b_2x^2+b_3x+b_4=\\
    -b_0x^4-\big(x-b_1\big)^4+\big(b_2+d_0\big)x^2-d_0\big(x-d_1\big)^2+d_2=\\
    -b_0\big(x^2-e_0\big)^2-\big(x-b_1\big)^4-d_0\big(x-d_1\big)^2+e_1
\end{multline}
with
\begin{equation}\label{eq:HRmodel:dV_in_x_notation}
\begin{split}
    a_0&=c_1 a \Var{1}, \ a_1=c_1 b, \  a_2=\frac{{c_1}^2}{4c_2(1-\Var{2})}+\frac{(c_3 \eps s-c_1)^2}{4c_3\eps(1-\Var{3})}\\
    a_3&=c_1 I, \  a_4=0  \  b_0= a_0-1, \  b_1= \tfrac{1}{4}a_1, \
    b_2=a_2+\tfrac{3}{8}{a_1}^2, \  b_3=a_3-\tfrac{1}{16}{a_1}^3,
    \  b_4=a_4+\tfrac{1}{256}{a_1}^4 \\
    d_0&= 1, \  d_1= \frac{b_3}{2d_0}, \  d_2= b_4+{d_1}^2 d_0, \
    e_0=\frac{b_2+d_0}{2b_0}, \ e_1= d_2+b_0{e_0}^2
\end{split}
\end{equation}

Notice that we want the value of $b_0$ in
(\ref{eq:HRmodel:dV_in_x}), (\ref{eq:HRmodel:dV_in_x_notation}) be
positive. Hence the value of
\[
a_0= c_1 a \Var{1}
\]
should be greater than $1$. This can be ensured by choosing the
value of $c_1$ in (\ref{eq:Lyapunov_Candidate}) to be sufficiently
large. As a result of this choice, taking restrictions
(\ref{eq:HRmodel:dV_C2}) into account, we conclude that the value
of $c_2$ in (\ref{eq:Lyapunov_Candidate}) must be sufficiently
small, e.g. satisfy the following inequality:
\[
{c_2}<c_1\frac{4a\Var{2}(1-\Var{1})}{d^2}.
\]
The value for $d_0$ can be chosen arbitrarily, here $d_0=1$.

Time-derivative $\dot{V}$ can now be written as follows
\begin{multline}\label{eq:HRmodel:dV_2}
    \dot{V}(x,y,z) = -c_1a(1-\Var{1})\Big(x^2+\frac{c_2d}{2c_1a(1-\Var{1})}y\Big)^2\\
        -c_3\eps(1-\Var{3})\Big(z-\frac{c_3\eps s-c_1}{2c_3\eps(1-\Var{3})}x\Big)^2\\
        -c_2(1-\Var{2})\Big(y-\frac{c_1}{2c_2(1-\Var{2})}x\Big)^2\\
        -c_3\eps\Var{3}\Big(z-\frac{sx_0}{2\Var{3}}\Big)^2 +\frac{c_3\eps s^2{x_0}^2}{4\Var{3}}+\\
-c_2\big(\Var{2}-\frac{c_2d^2}{4c_1a(1-\Var{1})}\big)\Big(y-\frac{2cc_1a(1-\Var{1})}{4\Var{2}c_1a(1-\Var{1})-c_2d^2}\Big)^2+\frac{c_1c_2c^2a(1-\Var{1})}{4\Var{2}c_1a(1-\Var{1})-c_2d^2}+\\
        -b_0\big(x^2-e_0\big)^2-\big(x-b_1\big)^4-d_0\big(x-d_1\big)^2+e_1 + c_1 xu
\end{multline}

It is straightforward to see that expression
(\ref{eq:HRmodel:dV_2}) is of the form
(\ref{eq:equality_PreSemiPassivity}), where
\[
\begin{split}
H_0(x,y,z)=& \:
c_1a(1-\Var{1})\Big(x^2+\frac{c_2d}{2c_1a(1-\Var{1})}y\Big)^2+
c_3\eps(1-\Var{3})\Big(z-\frac{c_3\eps
s-c_1}{2c_3\eps(1-\Var{3})}x\Big)^2\\
&+c_2(1-\Var{2})\Big(y-\frac{c_1}{2c_2(1-\Var{2})}x\Big)^2+c_3\eps\Var{3}\Big(z-\frac{sx_0}{2\Var{3}}\Big)^2\\
& +
c_2\big(\Var{2}-\frac{c_2d^2}{4c_1a(1-\Var{1})}\big)\Big(y-\frac{2cc_1a(1-\Var{1})}{4\Var{2}c_1a(1-\Var{1})-c_2d^2}\Big)^2\\
& + b_0\big(x^2-e_0\big)^2+\big(x-b_1\big)^4+d_0\big(x-d_1\big)^2
\end{split}
\]
\[
M=\frac{c_3\eps s^2{x_0}^2}{4\Var{3}} +
\frac{c_1c_2c^2a(1-\Var{1})}{4\Var{2}c_1a(1-\Var{1})-c_2d^2} + e_1
\]

Let us denote
\[
H_1(x,y,z)=H_0-M
\]
and rewrite (\ref{eq:equality_PreSemiPassivity}) as
\[
\dot{V}\leq c_1 x u - H_1(x,y,z)
\]
Function $H_1(x,y,z)$ is radially unbounded. Furthermore, it is
non-negative outside a ball in $\Real^3$. Hence choosing
\[
V^\ast(x,y,z)=\frac{1}{c_1} V(x,y,z)
\]
we assure existence of (radially unbounded) positive definite
$V^\ast(x,y,z)$ such that
\begin{equation}\label{eq:HR_Semi-Passivity_proof}
\dot{V}^\ast\leq x u - \frac{H_1(x,y,z)}{c_1},
\end{equation}
where $H_1(x,y,z)/c_1$ is radially unbounded and non-negative
outside a ball in $\Real^3$. Thus, according to
(\ref{eq:Lyapunov_Candidate}), semi-passivity of the
Hindmarsh-Rose system follows.

{\it 2) Boundedness of the solutions.} We aim to prove that
boundedness of  $\phi_i(t)$, $i\in\{0,\dots,n\}$ implies
boundedness of the state of the coupled system. Without loss of
generality we assume that
\[
\|\phi_i(\tau)\|_{\infty,[0,\infty]}\leq D_\phi
\]

Let us denote
\[
V_i=V^\ast(x_i,y_i,z_i), \ \
H_{1,i}=\frac{1}{c_1}H_1(x_i,y_i,z_i).
\]
Consider the following function
\begin{equation}\label{eq:HR_boundedness:1}
V_{\Sigma}(\bfx,\bfy,\bfz)=\rho\left(\sum_{i=0}^n
V_i(x_i,y_i,z_i), C\right).
\end{equation}
where $\bfx=\col(x_0,\dots,x_n)$, $\bfy=\col(y_0,\dots,y_n)$,
$\bfz=\col(z_0,\dots,z_n)$ and
\[
\rho(s,C)=\left\{\begin{array}{cl}
                  s-C,& \ s\geq C\\
                  0,& s < C
                  \end{array}
             \right.
\]
Function $V_{\Sigma}$ is nonnegative for any $C\in\Real$ and,
furthermore, is radially unbounded. Hence, its boundedness for
some $C\in\Real$ implies boundedness of $x_i$, $y_i$, $z_i$,
$i\in\{0,\dots,n\}$.

Let us pick $C\in\Real$ such that interior of the domain
\[
\Omega_{C}=\{\bfx,\bfy,\bfz\in\Real \ \left|\right. \ \sum_{i=0}^n
V_i(x_i,y_i,z_i)\leq C\}
\]
contains the domain
\[
\sum_{i=0}^n H_{1,i}(x_i,y_i,z_i) - \kappa x_i^2 < M_i, \
M_i\in\Real_{>0}, \ \kappa\in\Real_{>0}
\]
where $M_i$ is an arbitrarily large and $\kappa$ is an arbitrary
small positive constant. In other words the following implication
holds:
\begin{equation}\label{eq:HR_coupled_interior}
\sum_{i=0}^n V_i(x_i,y_i,z_i)\geq C \Rightarrow \sum_{i=0}^n
H_{1,i}(x_i,y_i,z_i)-\kappa x_i^2\geq M_i
\end{equation}
Such $C$ always exists because $H_{1,i}(x_i,y_i,z_i)-\kappa x_i^2$
can be expressed as a sum of a nonnegative quadratic form in
$x_i,y_i,z_i$ and non-negative functions of the higher order plus
a constant, and $V_i(x_i,y_i,z_i)$ is a positive-definite
quadratic form.

Consider time-derivative of function $V_{\Sigma}(\bfx,\bfy,\bfz)$.
According to (\ref{eq:HR_boundedness:1}),
(\ref{eq:HR_Semi-Passivity_proof}) it is zero for all
$\bfx,\bfy,\bfz\in\Omega_C$, and satisfies the following
inequality  otherwise:
\[
\dot{V}_{\Sigma}\leq \sum x_i u_i - \sum_{i=0}^n
H_{1,i}(x_i,y_i,z_i)= \gamma \bfx^{T} \Gamma \bfx + \sum_{i=0}^n
x_i \phi_i(t) - \sum_{i=0}^n H_{1,i}(x_i,y_i,z_i)
\]
Using Gershgorin's circle theorem, we can conclude that
\[
\dot{V}_{\Sigma}\leq \sum_{i=0}^n x_i \phi_i(t) - \sum_{i=0}^n
H_{1,i}(x_i,y_i,z_i)
\]
Rewriting
\[
x_i \phi_i(t)=-\kappa \left(x_i-\frac{\phi_i(t)}{2\kappa}\right)^2
+ \kappa x_i^2 + \frac{1}{4\kappa}\phi_i^2(t), \ \kappa>0
\]
leads to the following inequality
\[
\dot{V}_{\Sigma}\leq \kappa \sum_{i=0}^n x_i^2  - \sum_{i=0}^n
\left(H_{1,i}(x_i,y_i,z_i)-\frac{D_\phi^2}{4\kappa}\right)=-\sum_{i=0}^n
\left(H_{1,i}(x_i,y_i,z_i)-\frac{D_\phi^2}{4\kappa}-\kappa
x_i^2\right)
\]
Hence, choosing the value of $C$ such that $M_i\geq
{D_\phi^2}/{4\kappa}$ in (\ref{eq:HR_coupled_interior}) we can
ensure that
\[
\dot{V}_{\Sigma}\leq 0
\]
This implies that $V_{\Sigma}(\bfx(t),\bfy(t),\bfz(t))$ is not
growing with time. Hence trajectories $x_i(t)$, $y_i(t)$, $z_i(t)$
in the coupled system are bounded.

{\it 3) Convergence to a vicinity of the synchronization
manifold.} Consider the $i$-th and $j$-th oscillators in
(\ref{HR_model_net}), $i,j\in\{0,\dots,n\}$, $i\ne j$. Let us
introduce the following function
\begin{eqnarray}\label{proof:prop1:0}
V=0.5 \left(C_x (x_i-x_{j})^{2}+C_y (y_i-y_{j})^{2}+C_z
(z_i-z_{j})^{2}\right),
\end{eqnarray}
where $C_x$, $C_y >0$ are to be defined and $C_z=C_x/( s\eps)$.

Its time-derivative can be expressed as follows:
\begin{eqnarray}\label{proof:prop1:1}
\dot{V}&=&- C_x (x_i-x_{j})^2\left(\frac{a x_i^{2}}{2}+\frac{a
x_{j}^{2}}{2}+\frac{a(x_{i}+x_{j})^2}{2}-b(x_i+x_{j}) + \gamma
(n+1) \right) \\ & & + C_x (y_{i}-y_{j})(x_i-x_{j})- C_y
d(x_{i}-x_{j})(x_{i}+x_{j})(y_{i}-y_{j}) \nonumber\\   & &-
C_y(y_{i}-y_{j})^2-C_z\eps(z_{i}-z_{j})^2+C_x(x_i-x_j)(\phi_i-\phi_j)\nonumber
\end{eqnarray}
Consider the following term in (\ref{proof:prop1:1}):
\[
C_x (y_{i}-y_{j})(x_i-x_{j})- C_y
d(x_{i}-x_{j})(x_{i}+x_{j})(y_{i}-y_{j}) - C_y(y_{i}-y_{j})^2.
\]
It can be written as follows:
\begin{eqnarray}\label{eq:HR_synch_proof_1}
& & \frac{C_x^2}{4
C_y\Delta_1}(x_i-x_{j})^{2}-\left(\left(\frac{C_x^2}{4
C_y\Delta_1}\right)^{0.5}(x_{i}-x_{j})-\left(\Delta_1
C_y\right)^{0.5}(y_i-y_{j})\right)^2 + \nonumber\\ & & +\frac{C_y
d^2}{4\Delta_2}(x_i-x_{j})^{2}(x_i+x_{j})^2 -
C_{y}\left(\left(\frac{
d^2}{4\Delta_2}\right)^{0.5}(x_i^2-x_{j}^2)+\Delta_2^{0.5}(y_{i}-y_{j})
\right)^2\nonumber\\
&& - (1-\Delta_1-\Delta_2) (y_i-y_j)^2,
\end{eqnarray}
where $\Delta_1,\Delta_2\in\Real_{>0}$ and
$\Delta_1+\Delta_2\in(0,1)$. Taking (\ref{eq:HR_synch_proof_1})
into account we rewrite (\ref{proof:prop1:1}) as:
\begin{eqnarray}\label{proof:prop1:2}
\dot{V}& \leq & -C_x
(x_i-x_{j})^2\left(\frac{ax_i^{2}}{2}+\frac{ax_{j}^{2}}{2}+\frac{a(x_{i}+x_{j})^2}{2}
- \frac{C_y d^2}{C_x 4 \Delta_2}(x_i+x_{j})^{2}
\right. \\
& & \left. - b(x_i+x_{j}) + \gamma (n+1) -\frac{C_x}{4
C_y\Delta_1}\right)-C_z\eps(z_{i}-z_{i+1})^2 \nonumber\\
& &- C_y(1-\Delta_1
-\Delta_2)(y_i-y_j)^2+C_x(x_i-x_j)(\phi_i-\phi_j)\nonumber
\end{eqnarray}
Let
\[
\frac{C_y }{C_x}=\frac{2a \Delta_2}{d^{2}}.
\]
Then
\begin{eqnarray}\label{eq:HR_synch_proof_2}
\dot{V} &\leq&  -C_x
(x_i-x_{j})^2\left(\frac{a}{2}\left(x_i-\frac{b}{a}\right)^2+\frac{a}{2}\left(x_{j}-\frac{b}{a}\right)^2
+ \gamma (n+1) -\frac{d^2}{8 a \Delta_1\Delta_2}-\frac{b^2}{a}
\right)\nonumber\\
& &  - (1-\Delta_1-\Delta_2) C_y (y_i-y_j)^2
-C_z\eps(z_{i}-z_{i+1})^2 + C_x(x_i-x_j)(\phi_i-\phi_j).
\end{eqnarray}
Hence, choosing
\[
\gamma > \frac{1}{(n+1) a}\left({\frac{d^2}{8 \Delta_1
\Delta_2}+{b^2}}\right).
\]
we can ensure that the first term in (\ref{eq:HR_synch_proof_2})
is non-positive. The minimal value of $\gamma$ ensuring this
property can be calculated by minimizing the value
\[
\frac{1}{8 \Delta_1 \Delta_2}
\]
for all $\Delta_1$, $\Delta_2\in\Real_{>0}$:
$\Delta_1+\Delta_2<1$. This can be done by letting
$\Delta_2=r-\Delta_1$, $r\in(0,1)$ and differentiating the term
${1}/(8 \Delta_1 (r-\Delta_1))$ with respect to $\Delta_1$. This
leads to the following solution: $\Delta_1=r/2$, $\Delta_2=r/2$.
Taking this into account we rewrite (\ref{eq:HR_synch_proof_2}) as
follows
\begin{eqnarray}\label{eq:HR_synch_proof_3}
\dot{V} &\leq&  -C_x
(x_i-x_{j})^2\left(\frac{a}{2}\left(x_i-\frac{b}{a}\right)^2+\frac{a}{2}\left(x_{j}-\frac{b}{a}\right)^2
+ \gamma (n+1) -\frac{d^2}{2 a r}-\frac{b^2}{a}
\right)\nonumber\\
& &  - (1-r) C_y (y_i-y_j)^2 -C_z\eps(z_{i}-z_{i+1})^2 +
C_x(x_i-x_j)(\phi_i-\phi_j).
\end{eqnarray}

Let
\[
\gamma = \frac{1}{(n+1) a}\left({\frac{d^2}{2}+{b^2}}\right) +
\varepsilon_1, \ \varepsilon_1\in\Real_{>0}.
\]
Alternatively, we can rewrite this as
\[
\gamma = \frac{1}{(n+1) a}\left({\frac{d^2}{2r}+{b^2}}\right) +
\varepsilon_2, \ r\in(0,1), \ \varepsilon_2\in\Real_{>0}
\]
Hence, according to (\ref{eq:HR_synch_proof_3}) the following
inequality holds:
\[
\dot{V} \leq  -C_x \varepsilon_2 (x_i-x_{j})^2  - (1-r) C_y
(y_i-y_j)^2 -C_z\eps(z_{i}-z_{i+1})^2 +
C_x(x_i-x_j)(\phi_i-\phi_j).
\]
Then denoting $\alpha=2\min\{\varepsilon_2,\varepsilon,(1-r)\}$ we
obtain
\begin{equation}\label{eq:HR_synch_proof_4}
\dot{V}\leq - \alpha V + C_x(x_i-x_j)(\phi_i-\phi_j)
\end{equation}

Consider the following differential equation
\begin{equation}\label{eq:HR_synch_proof_5}
\dot{\upsilon}=-\alpha \upsilon + C_x (x_i-x_j)(\phi_i-\phi_j)
\end{equation}
Its solution can be estimated as follows
\[
|\upsilon(t)|\leq e^{-\alpha (t-t_0)} |\upsilon(t_0)|+e^{-\alpha
t}\int_{t_0}^t e^{\alpha \tau} C_x
(x_i(\tau)-x_j(\tau))(\phi_i(\tau)-\phi_j(\tau))d\tau
\]
for all $t\geq t_0$. Given that $x_i(t)$, $x_j(t)$ are bounded
there exists a constant $B$ such that
\[
|\upsilon(t)|\leq e^{-\alpha (t-t_0)} |\upsilon(t_0)|+\frac{C_x
B}{\alpha} \|\phi_i(\tau)-\phi_j(\tau)\|_{\infty,[t_0,t]}
\]
Then applying comparison lemma  (see, for example
\cite{Khalil:2002}, page 102) we can conclude that
\[
V(t)\leq  e^{-\alpha (t-t_0)}V(t_0) + \frac{C_x B}{\alpha}
\|\phi_i(\tau)-\phi_j(\tau)\|_{\infty,[t_0,t]}.
\]
Hence, conclusion 2) of the theorem follows. {\it The theorem is
proven.}

\bibliographystyle{plain}
\bibliography{Adaptive_visual_system}

\end{document}